\documentclass[a4paper,fleqn, review]{cas-dc}
\usepackage{amsmath}
\usepackage[switch]{lineno}
\usepackage{graphicx}
\usepackage{subfig} 
\usepackage{tabu}  
\usepackage{hyperref}
\usepackage{multirow}
\usepackage{dsfont}
\usepackage{stfloats}
\usepackage{cite}
\usepackage{booktabs}
\usepackage{algorithm}
\usepackage{algorithmic}
\usepackage{amssymb}
\usepackage{makecell}
\usepackage{caption}
\usepackage{natbib}

\def\tsc#1{\csdef{#1}{\textsc{\lowercase{#1}}\xspace}}
\tsc{WGM}
\tsc{QE}
\tsc{EP}
\tsc{PMS}
\tsc{BEC}
\tsc{DE}

\begin{document}
\let\WriteBookmarks\relax
\def\floatpagepagefraction{1}
\def\textpagefraction{.001}
\shorttitle{}

\shortauthors{Hongruixuan Chen et~al.}

\title [mode = title]{Exchange means change: an unsupervised single-temporal change detection framework based on intra- and inter-image patch exchange}                      



%
\author[1,2]{Hongruixuan Chen}[orcid=0000-0003-0100-4786]

\author[1,2]{Jian Song}





\author[3]{Chen Wu}

\author[4]{Bo Du}

\author[1,2]{Naoto Yokoya}
\cormark[1]

\affiliation[1]{organization={Graduate School of Frontier Sciences, The University of Tokyo},
    city={Chiba},
    postcode={277-8561}, 
    country={Japan}}



\affiliation[2]{organization={RIKEN Center for Advanced Intelligence Project (AIP), RIKEN},
    city={Tokyo},
    postcode={103-0027}, 
    country={Japan}}

\affiliation[3]{organization={State Key Laboratory of Information Engineering in Surveying, Mapping, and Remote Sensing, Wuhan University},
    city={Wuhan},
    postcode={430079}, 
    country={PR China}}
    
\affiliation[4]{organization={School of Computer, Wuhan University},
    city={Wuhan},
    postcode={430079}, 
    country={PR China}}


\cortext[cor1]{Corresponding author}


\nonumnote{Manuscript submitted on May 23, 2023.}

\begin{abstract}
Change detection is a critical task in studying the dynamics of ecosystems and human activities using multi-temporal remote sensing images. While deep learning has shown promising results in change detection tasks, it requires a large number of labeled and paired multi-temporal images to achieve high performance. Pairing and annotating large-scale multi-temporal remote sensing images is both expensive and time-consuming. To make deep learning-based change detection techniques more practical and cost-effective, we propose an unsupervised single-temporal change detection framework based on intra- and inter-image patch exchange (I3PE). The I3PE framework allows for training deep change detectors on unpaired and unlabeled single-temporal remote sensing images that are readily available in real-world applications. The I3PE framework comprises four steps: 1) intra-image patch exchange method is based on an object-based image analysis (OBIA) method and adaptive clustering algorithm, which generates pseudo-bi-temporal image pairs and corresponding change labels from single-temporal images by exchanging patches within the image; 2) inter-image patch exchange method can generate more types of land-cover changes by exchanging patches between images; 3) a simulation pipeline consisting of several image enhancement methods is proposed to simulate the radiometric difference between pre- and post-event images caused by different imaging conditions in real situations; 4) self-supervised learning based on pseudo-labels is applied to further improve the performance of the change detectors in both unsupervised and semi-supervised cases. Extensive experiments on two large-scale datasets covering Hongkong, Shanghai, Hangzhou, and Chengdu, China, demonstrate that I3PE outperforms representative unsupervised approaches and achieves F1 value improvements of 10.65$\%$ and 6.99$\%$ to the state-of-the-art method. Moreover, I3PE can improve the performance of the change detector by 4.37$\%$ and 2.61$\%$ on F1 values in the case of semi-supervised settings. Additional experiments on a dataset covering a study area with 144 $km^{2}$ in Wuhan, China, confirm the effectiveness of I3PE for practical land-cover change analysis tasks.
\end{abstract}


\maketitle

\section{Introduction}\label{sec:1}
\par Ecosystems and human activities on the Earth's surface are constantly changing. Obtaining accurate information on surface changes in real-time is essential to understanding and studying human activities, the natural environment, and their interactions \citep{Coppin2004}. Remote sensing technology is a powerful tool that allows for large-scale, long-term, periodic observations of the Earth's surface, making it a vital tool for studying changes in the Earth's ecosystem and human society. As such, detecting land-cover changes from multi-temporal remote sensing images acquired by sensors mounted on spaceborne and airborne remote sensing platforms has become a topic of great interest in the field of remote sensing \citep{Tewkesbury2015,Zhu2017b}.

\par As one of the earliest and most widely used technologies in the field of remote sensing, there have been numerous approaches and paradigms developed for change detection. Before the advent of deep learning techniques, traditional change detection methods could be roughly classified into four types: image algebra methods, image transformation methods, post-classification comparison methods, and other advanced methods. Image algebra methods measure the change intensity by directly comparing spectral bands of bi-temporal images. The most classic method in this category is change vector analysis (CVA) \citep{Sharma2007, Bovolo2007a,DU2020278}. Image transformation methods aim to extract features that are beneficial for change detection by transforming the raw image features into a new feature space. Representative methods include multivariate alteration detection (MAD) \citep{Nielsen1997}, principal component analysis (PCA) \citep{Deng2008,Celik2009}, slow feature analysis (SFA) \citep{Wu2014}, Fourier transform \citep{chen2023fourier}, and so on. Post-classification comparison methods first execute classification algorithms to obtain classification maps and then compare the classification maps to generate change maps \citep{Xian2009}. Other advanced methods mainly include the utilization of machine learning models such as support vector machine \citep{Bovolo2008}, conditional random field \citep{Hoberg2015}, Markov random field \citep{Kasetkasem2002}, and the object-based image analysis (OBIA) methods for change detection \citep{Hussain2013,Gil-Yepes2016}.

\par The emergence of deep learning techniques in recent years has brought about new paradigms and solutions to change detection, resulting in improved efficiency and accuracy in analyzing multi-temporal remote sensing imagery \citep{Shi2020Change}. These deep learning-based methods can be categorized into unsupervised and supervised types, depending on whether prior annotated information is provided to the change detector. For unsupervised methods based on deep learning, the primary research direction is to develop or utilize deep learning models to extract spatial-spectral features from multi-temporal remote sensing images and subsequently employ models or operations to calculate change intensity from these features. In \citep{Zhang2016c}, the deep belief network (DBN) was used to extract features from bi-temporal images for change detection. Likewise, autoencoder and its variants were also widely utilized to extract features by reconstructing the input multi-temporal images for unsupervised change detection \citep{Zhang2016,Liu2018,Bergamasco2022Unsupervised}. \cite{Saha2019} proposed a deep CVA (DCVA) framework for unsupervised binary and multiclass change detection, which utilizes a pre-trained deep convolutional neural network to extract features from bi-temporal images and then performs binarization operation and the CVA algorithm to detect land-cover changes. \cite{Liu2020Bipartite} proposed a bipartite differential neural network to make the detection results robust to co-registration errors. In \citep{Liu2022}, a bipartite convolutional neural network combined with a Gibbs probabilistic model was proposed for change detection on heterogeneous data. In \citep{Wu2022Unsupervised}, an unsupervised feature extraction model based on kernel PCA, called KPCA convolution, was developed for extracting spatial-spectral features from remote sensing images. Based on this model, a deep network architecture was further proposed for unsupervised change detection. Recently, graph convolutional networks (GCNs) \citep{kipf2016semi} have also been introduced to the change detection task for capturing nonlocal dependencies in the spatial and temporal order of multi-temporal remote sensing images \citep{Tang2022,chen2022unsupervised}. Although unsupervised approaches do not require labeled data for training change detectors, the features extracted may not be suitable for change detection, as the feature extraction process of the model is unconstrained. Furthermore, the absence of annotated data makes applying more powerful deep architectures challenging. Consequently, practical applications of these unsupervised models are often restricted to analyzing land-cover changes in small study areas.

\par In contrast to unsupervised change detection methods, supervised change detection methods require annotated data to train change detectors. These methods achieved higher accuracy due to the availability of prior information on land-cover change and the potential of applying more advanced deep architectures as change detectors. The dominant approaches are based on convolutional neural networks (CNNs) among the existing supervised methods. \cite{Zhan2017} designed a deep siamese convolutional network based on contrastive learning for change detection in optical aerial images. \cite{CayeDaudt2018} first introduced the fully convolutional network (FCN) with encoder-decoder architecture to the change detection task and presented three FCN architectures. After this, various more advanced network architectures were introduced and studied. An improved UNet++ was developed in \citep{Peng2019End} inspired by the UNet++ architecture proposed for medical images \citep{zhou2018unet}. \cite{HOU2021103} designed a dynamic-scale triple network to learn multi-scale land-cover change information. \cite{Zheng2022} proposed a deep multi-task encoder-transformer-decoder architecture for semantic change detection. \cite{CAO2023full} designed a full-level fused cross-task transfer learning architecture for building change detection. Attention and self-attention mechanisms were introduced to capture the most important channels and spatial areas for change detection \citep{Zhang2020, guo2021deep, CHEN2022101}. Some work attempts to combine CNNs with other deep architectures. In \citep{Mou2019,Chen2019a}, CNNs and RNNs were combined to detect land-cover change information better. In \citep{Wu2021multiscale}, GCNs were introduced to help CNNs model nonlocal relationships in multi-temporal images. The potential of combining OBIA methods and CNN architecture in change detection and damage assessment tasks was also studied \citep{ZHENG2021Building, Liu2021}. More recently, with the advances in computer vision, vision transformer architecture \citep{dosovitskiy2020image} has been introduced for change detection. This architecture has achieved better results than CNNs in some benchmark datasets and practical applications \citep{Chen2022Remote, Bandara2022Transformer,chen2022dual}. 

\par Behind the promising results of these supervised methods are many paired multi-temporal images and high-quality labeled data. In other words, in order to train a change detector that performs well and can be applied in practice, we need numerous pairwise annotated multi-temporal remote sensing images. Different from so-called single-temporal tasks such as land-cover/land-use classification and building footprint extraction tasks, obtaining a large-scale and high-quality training set for change detection is often more time-consuming and expensive \citep{TIAN2022164}. For each training sample, we need both paired pre- and post-event remote sensing images. Additional radiometric correction and geometric co-registration operations are required to preprocess the paired images. Moreover, since two images are involved, and many types of land-cover change combinations exist, labeling changed objects in large-scale scenes is also very labor-intensive. These points greatly restrict the application of supervised change detection models in real-world applications. Compared with paired and labeled multi-temporal remote sensing images, unpaired and unlabeled single-temporal images can be obtained more easily and at a lower cost. Every day we can obtain numerous unpaired remote sensing images from different satellite sensors. Therefore, we ask whether we could train a change detector with good performance from unlabeled and unpaired single-temporal images. Some of the previous studies have attempted to address one of these points. On the one hand, pre-detection methods are able to train supervised models in an unsupervised manner \citep{Gong2017Superpixel,Luppino2022Deep}. These methods first adopt unsupervised change detection methods to obtain pre-detection results as pseudo-labels. The pseudo-labels are then used to train deep change detectors. However, these methods still require paired multi-temporal images. Moreover, the pre-detection methods require additional change detection algorithms to be run on each image pair, which is very time-consuming in large-scale scenes. On the other hand, \cite{zheng2021change} tried to train change detectors using unpaired remote sensing imagery. However, although the limitation of paired images is lifted, the proposed framework requires high-quality land-cover/land-use semantic labels of remote sensing images, which is also very expensive in practice. 

\par In this paper, we lift these two restrictions on the inputs of change detection for training supervised learning models, namely paired and labeled multi-temporal images, and present an unsupervised single-temporal change detection framework. The whole framework is based on a very simple yet effective idea: exchanging image patches to generate land-cover changes. Specifically, we propose an intra-image patch exchange method and an inter-image patch exchange method based on an adaptive clustering algorithm and the OBIA method. They can generate pseudo-bi-temporal image pairs and corresponding change labels from unpaired and unlabeled single-temporal images. Then, we propose a simulation method for different imaging conditions to fit practical scenarios where radiation differences exist between pre- and post-event images due to varying imaging conditions. Afterward, we can train the change detector directly on the generated pseudo-bi-temporal remote sensing image samples as in supervised learning methods. Additionally, we introduce a pseudo-label-based self-supervised learning method to further enhance the performance of change detectors in unsupervised and semi-supervised scenarios. 

\par The remainder of this paper is organized as follows. Section \ref{sec:2} briefly describes two large benchmark datasets and research areas. Section \ref{sec:3} elaborates on the proposed framework. Experimental results and discussion are presented in Section \ref{sec:4}. In Section \ref{sec:5}, we present the limitations of the current framework and discuss future research in light of these limitations. Finally, we draw conclusions in Section \ref{sec:6}.

\section{Data description}\label{sec:2}
\subsection{Large-scale benchmark datasets}
\begin{figure}[!t]
  \centering
  \subfloat[]{
    \includegraphics[width=1.55in]{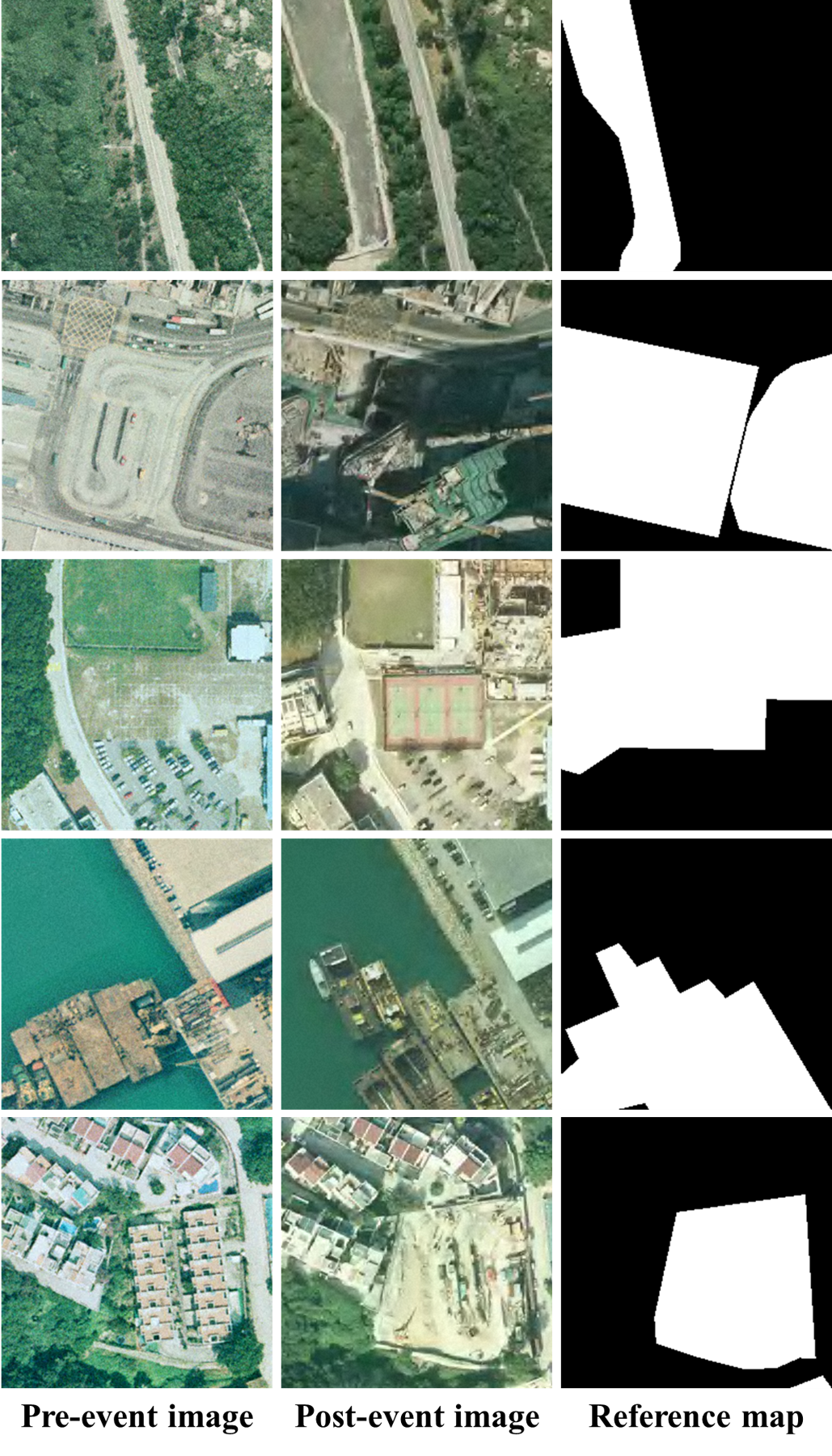}
  \label{fig_second_case}}
  \hfil
  \subfloat[]{
    \includegraphics[width=1.55in]{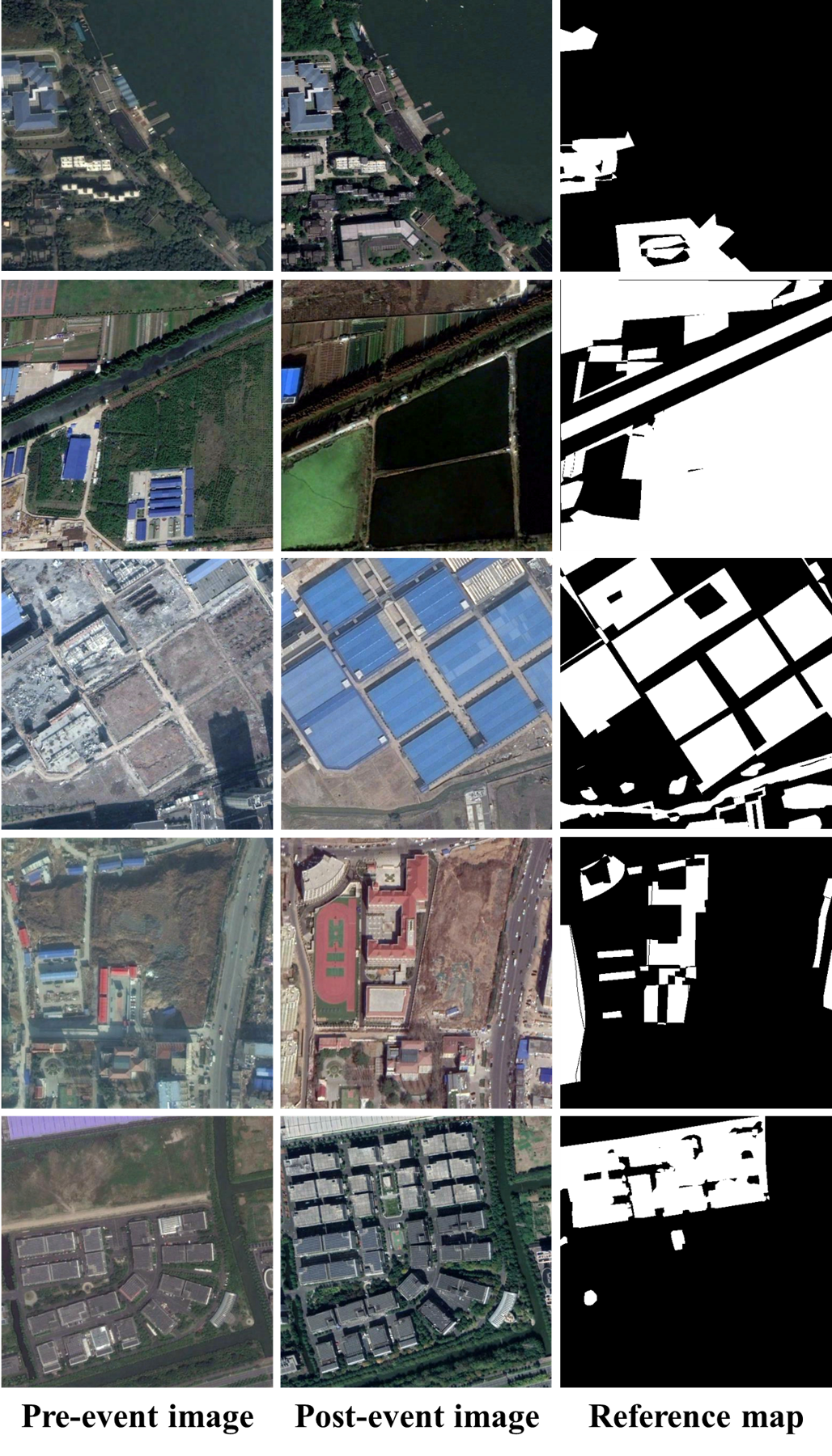}
  \label{fig_first_case}}
 
  \caption{Examples of bi-temporal image pairs and corresponding change reference maps from (a) SYSU dataset and (b) SECOND dataset.}
  \label{SECOND_SYSU_dataset}
\end{figure}
\par Most of the existing research on unsupervised change detection has only been validated on a few pairs of multi-temporal remote sensing images. In order to fully validate the performance of our proposed method under various scenarios and change events and provide a common benchmark for the remote sensing community, we utilize two publicly available large-scale land-cover change detection datasets: the SYSU dataset \citep{Shi2022Deeply} and the SECOND dataset \citep{Yang2022Asymmetric}.  

\par The SYSU dataset\footnote{https://github.com/liumency/SYSU-CD} comprises 20,000 pairs of bi-temporal aerial images with a spatial resolution of 0.5 m/pixel, captured between 2007 and 2014 in Hong Kong, China, a populous cosmopolitan city with a total land area of 1106.66 $km^{2}$ and a total population of approximately 7.2 million as of the end of 2014. This dataset presents the changes in urban built-up and port areas in response to the significant increase in construction and maintenance of port, sea, marine, and coastal projects in major shipping hubs during this period. The dataset contains six primary types of land-cover changes, which include new urban construction, suburban expansion, pre-construction groundwork, vegetation changes, road sprawl, and marine construction. These image pairs and corresponding change labels were split into three sets: a training set, a validation set, and a test set, comprising 12,000, 4,000, and 4,000 pairs, respectively. Figure \ref{SECOND_SYSU_dataset}-(a) shows some examples from the SYSU dataset. 

\par The SECOND dataset\footnote{https://captain-whu.github.io/SCD/} is another large-scale benchmark dataset with 4,662 pairs of bi-temporal images collected from various remote sensing platforms. The dataset mainly covers important cities in China, including Shanghai, Hangzhou, and Chengdu. It focuses on six land-cover categories, namely buildings, playgrounds, water, non-vegetated land surface, trees, and low vegetation, which are often involved in natural and human-induced changes. These categories produce 29 common land-cover change categories that adequately reflect the true distribution of land-cover categories when change events occur. Compared to the SYSU dataset, the SECOND dataset covers more research sites and has a much richer and more complex set of land-cover changes. The 4,662 pairs of bi-temporal images and corresponding change labels were initially split into a training set and a test set, comprising 2,968 and 1,694 pairs, respectively. Figure \ref{SECOND_SYSU_dataset}-(b) shows some instances from the SECOND dataset. 

\par Table \ref{dataset_info} summarizes the basic information of the two large-scale benchmark datasets. 

\begin{table*}[width=2.0\linewidth,cols=5,pos=t]
\caption{Information of the two large-scale change detection datasets used in our paper.}\label{dataset_info} 
\begin{tabular*}{\tblwidth}{@{} LLLLL @{} }
\toprule
Dataset	& Study site & Number of image pairs & Image size & Number of change types\\
\midrule
SYSU & Hong Kong, China & 20,000 (12,000/4,000/4,000) & 256 $\times$ 256 & 6 \\
SECOND & Shanghai, Hangzhou and Chengdu, China & 4,664 (2,968/1,694) & 512 $\times$ 512 & 29\\
\bottomrule
\end{tabular*}
\end{table*}

\subsection{Dataset for a local study area}
\begin{figure*}[!t]
  \centering
  \subfloat[]{
    \includegraphics[width=1.95in]{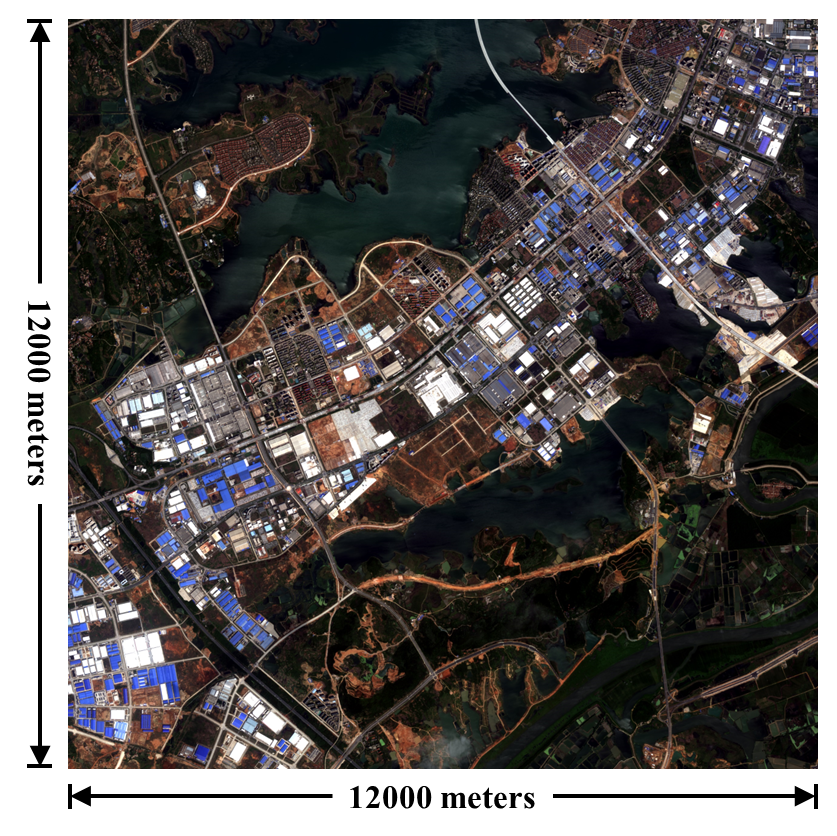}
  \label{fig_second_case}}
  \hfil
  \subfloat[]{
    \includegraphics[width=1.95in]{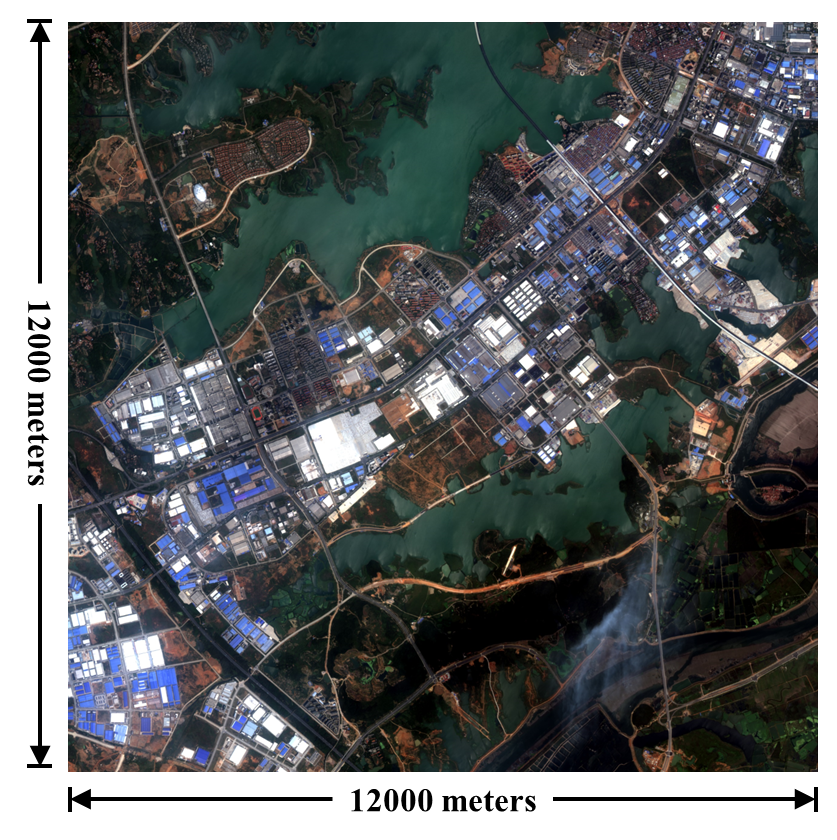}
  \label{fig_first_case}}
  \hfil
  \subfloat[]{
    \includegraphics[width=1.95in]{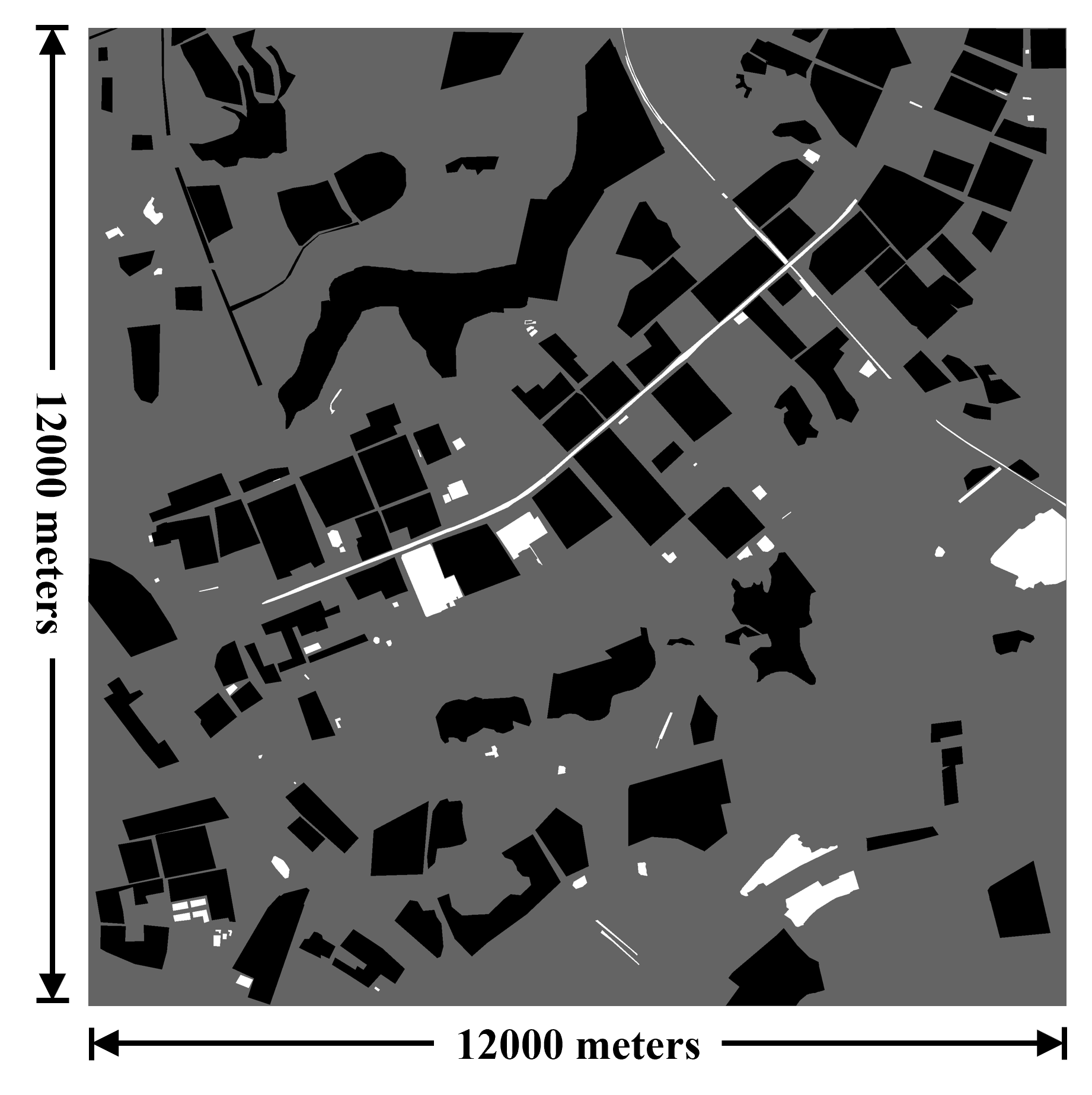}
  \label{fig_first_case}}
   \hfil
  \includegraphics[width=0.65in]{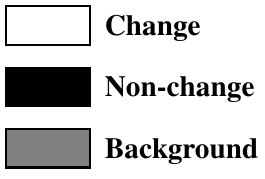}
  \caption{Wuhan dataset. (a) Pre-event image. (b) Post-event image. (c) Change reference map, where white indicates changed areas, black indicates unchanged areas, and gray is the background.}
  \label{WH_dataset}
\end{figure*}
\par In addition to verifying the effectiveness of our proposed method on two large-scale change detection datasets, we evaluate its applicability on a real-world dataset, namely the Wuhan dataset, to demonstrate its potential for land-cover change analysis at specific research sites. As shown in Figure \ref{WH_dataset}, the Wuhan dataset comprises pre-event and post-event images captured by the GF-2 satellite with an image size of 3,000$\times$3,000 and a spatial resolution of 4m/pixel on 2016/04/11 and 2016/09/01, respectively. The dataset covers 144 $km^{2}$ of developed and newly developing regions in Wuhan, China, the most populous city in Central China, with a population of over 11 million. The dataset has been processed by systematic radiometric correction and geometric co-registration with ground control points. In the reference map, white represents the changed area with 180,652 pixels, black represents the unchanged areas with 2,270,341 pixels, and the remaining gray areas are undefined and not involved in the accuracy assessment. Owing to the rapid development of Wuhan city, the study area experienced obvious land-cover changes caused by urban construction. The main change events between pre-event and post-event images are the construction of factories and railways, groundwork before building over, vegetation change, and water blooms.

\section{Methodology}\label{sec:3}
\begin{figure*}[!t]
  \centering
  \includegraphics[width=6.75in]{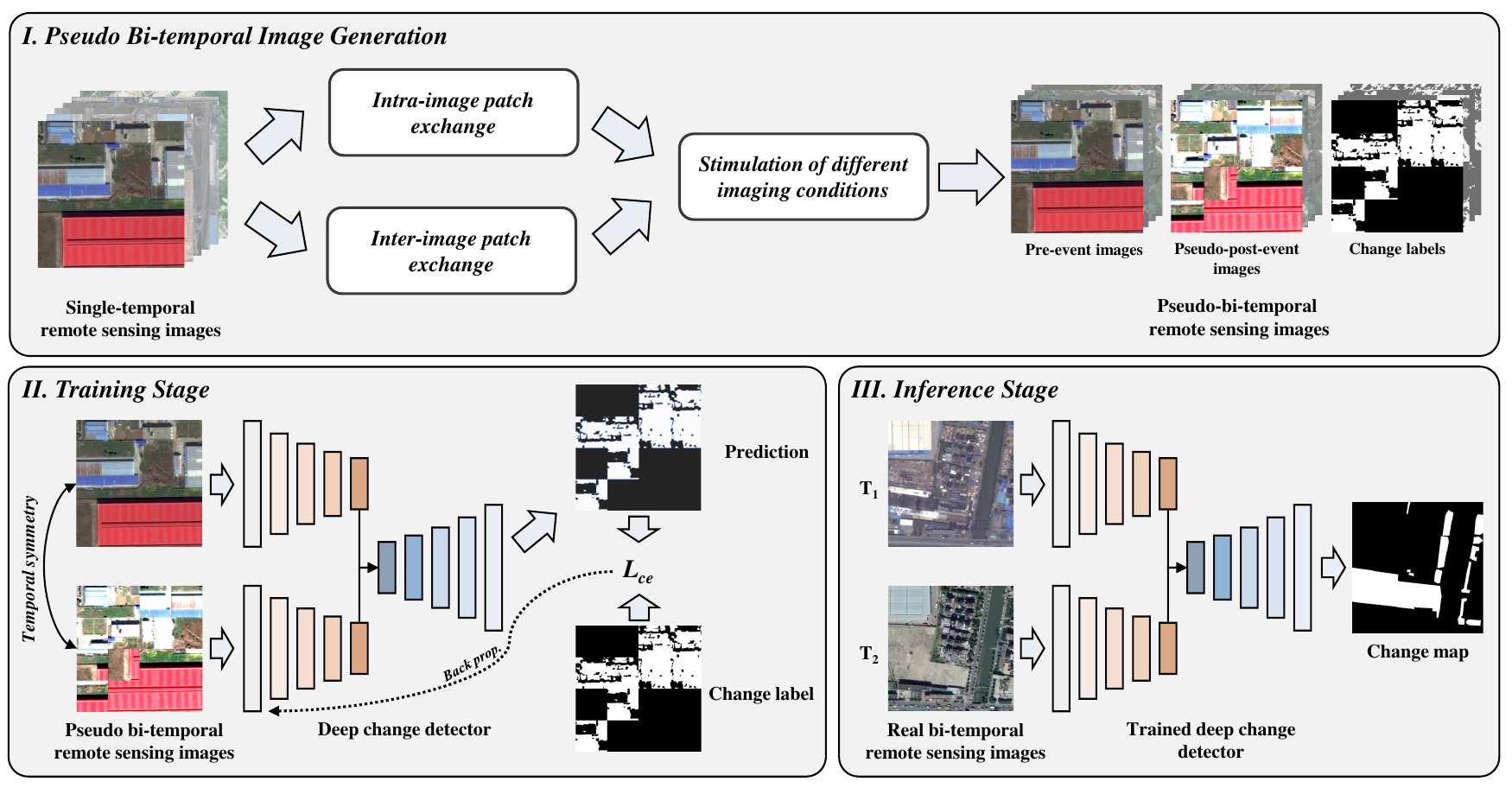}
  \caption{The overview of the proposed unsupervised single-temporal change detection framework based on intra- and inter-image patch exchange (I3PE).}
  \label{I3EP_framework}
\end{figure*}

\par The proposed unsupervised single-temporal change detection framework based on intra- and inter-image patch exchange is shown in Figure \ref{I3EP_framework}. Firstly, pseudo-bi-temporal remote sensing image pairs and associated change labels are generated from unlabelled and unpaired remote sensing images based on intra-image patch exchange and inter-image patch exchange methods. Then, a simulation algorithm is designed based on commonly used image enhancement methods to simulate radiometric differences caused by different imaging conditions. Subsequently, we train a deep change detector using the generated samples. In addition, we further employ a self-supervised learning approach based on pseudo-label training for improving detection performance in unsupervised and semi-supervised scenarios. Finally, the trained deep change detector is applied to detect land-cover changes from real bi-temporal remote sensing images during the inference stage. 

\subsection{Generating changes by exchanging image patches}
\par As we mentioned in Section \ref{sec:1}, we want to alleviate the constraints of the supervised deep learning-based change detection techniques on the input data and train a deep change detector from easily available unlabeled and unpaired images. The key to achieving this goal is to find a way to obtain (pseudo)-bi-temporal images and the corresponding change labels, which is necessary for training a deep change detector, from unlabelled and unpaired images. This work presents a simple but effective idea, i.e., generating pseudo-bi-temporal images and land-cover changes by exchanging image patches.

\begin{figure}[!t]
  \centering
  \includegraphics[width=3.3in]{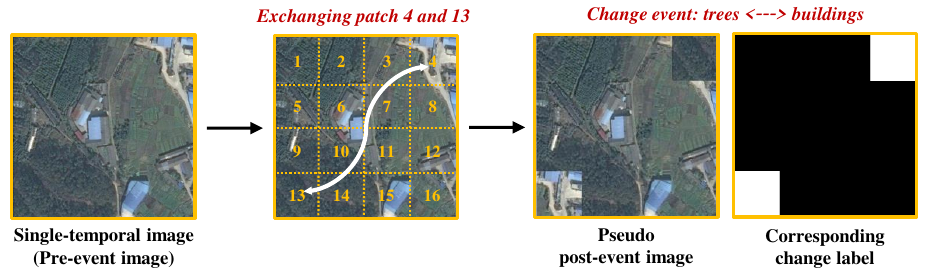}
  \caption{Illustration of generating pseudo-bi-temporal images and land-cover changes from an unlabelled image by simply exchanging patches within the image.}
  \label{motivation}
\end{figure}
\par Since a remote sensing image usually contains different kinds of land-cover objects, we can artificially generate land-cover changes by exchanging the image patches where different land-cover objects are located. For example, buildings, farmland, and trees are major land-cover features in the single-temporal image in Figure \ref{motivation}. After we exchange image patches numbered \textbf{4} and \textbf{13}, we can get a pseudo-post-event image. The change event happening in this artificially constructed image pair is the transformation of trees into buildings and buildings into trees. 

\par However, two main problems exist with using the above process directly to generate training samples. Firstly, we do not know exchanging which image patches can yield land-cover changes. Secondly, the two exchanged image patches do not necessarily contain totally different land-cover objects. Therefore, there would be much noise in the labels obtained by directly treating all pixels within the areas where the exchanged image patches are located as changes. We propose an intra-image patch exchange method by introducing an OBIA method and adaptive clustering algorithm to address the above problems, thereby effectively and efficiently yielding bi-temporal remote sensing images with relatively accurate change labels for training deep change detectors. 

\subsubsection{Intra-image patch exchange method}
\begin{figure*}[!t]
  \centering
  \includegraphics[width=6.75in]{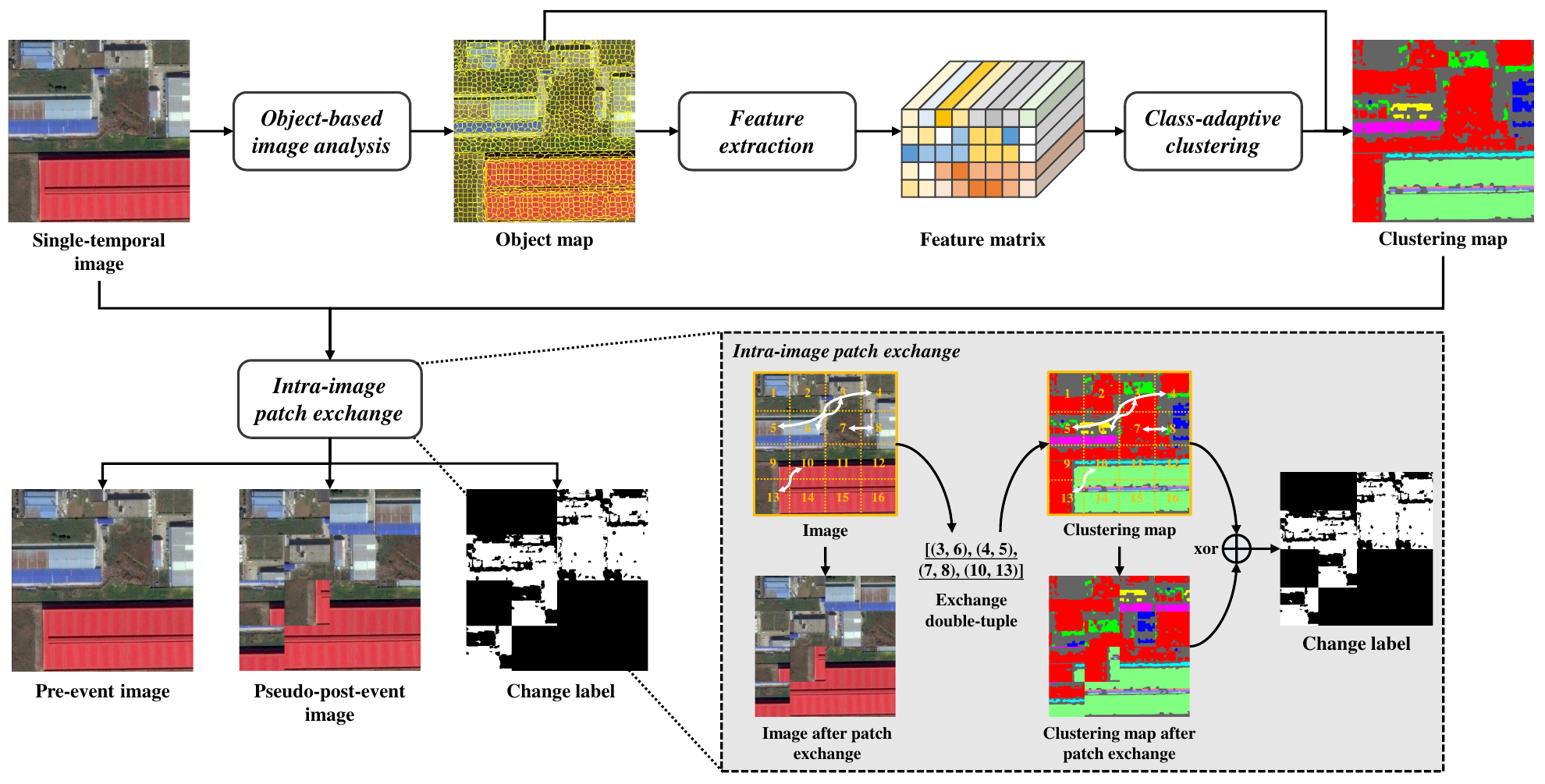}
  \caption{Workflow of the proposed intra-image patch exchange method.}
  \label{IntRA_Exc}
\end{figure*}
\par To tackle the abovementioned issues, we propose to first perform a clustering algorithm on single-temporal images in an unsupervised manner. If the clustering results are close to the actual land-cover situation, then accurate change labels can be obtained by comparing the clustering results in the locations of the two exchanged image patches. In this way, the two problems mentioned above can be effectively solved. Nevertheless, traditional clustering algorithms such as K-means require a predetermined number of clusters to be specified, whereas the number of land-cover objects varies in different images. Here, we introduce an adaptive clustering algorithm, density-based spatial clustering of applications with noise (DBSCAN) \citep{ester1996density}, to get the clustering maps of single-temporal images. As an adaptive clustering algorithm, DBSCAN can detect clusters of any arbitrary shape and size in datasets containing even noise and outliers, making it suitable for processing remote sensing images with different types of land-cover objects. Regarding the input of DBSCAN, we propose to use image objects instead of pixels as the basic analysis unit of the clustering algorithm. The advantage of doing so is that it exploits spatial information, which can avoid some noisy results while reducing the amount of data and improving computational efficiency. 

\par Figure \ref{IntRA_Exc} displays the specific workflow of our intra-image patch exchange method. Given a single-temporal image $X^{T_{1}} \in \mathbb{R}^{H \times W \times C}$, where $H$, $W$, and $C$ are the height, width, and channel of the image, respectively, the simple linear iterative clustering (SLIC) algorithm \citep{Achanta2012SLIC} is first performed on $X^{T_{1}}$ to get the image object map $\mathrm{\Omega}$ as

\begin{equation}
\left\{
    \begin{aligned}
      &\mathrm{\Omega}=\{\mathrm{\Omega}_{i} \mid i=1,2,\ldots,N_{o}\}  \\
      &\mathrm{\Omega}_{i}\cap\mathrm{\Omega}_{j} =\varnothing \ \ \mathrm{if} \ i\ \neq j  \\
      &\bigcup_{i=1}^{N_{o}}\mathrm{\Omega}_{i}=\{\left(h,w\right)\mid h=1,\dots,H; w=1,\ldots,W\} 
    \end{aligned}
\right.
\label{eq:1}
\end{equation}
where $N_{o}$ is the number of the objects. The $i$-th object in $X^{T_{1}}$ are defined as $X^{T_{1}}_{i}=\left\{X^{T_{1}}(h,w,c) \ | \ (h,w) \in \mathrm{\Omega}_{i}; c=1,...,C \right\}$. 

\par After $\mathrm{\Omega}$ is obtained, different kinds of features are extracted from the image objects as the input of the subsequent clustering algorithm, i.e., $\mathcal{X}^{T_{1}}_{i}=\mathcal{F}({X}^{T_{1}}_{i})$, where $\mathcal{F}$ is the feature extraction operator. In this paper, the mean and standard variance values in each channel are extracted as the objects' features. DBSCAN is performed on $\mathcal{X}^{T_{1}}=\left[\mathcal{X}_{1}^{T_{1}},\mathcal{X}_{2}^{T_{1}},\cdots,\mathcal{X}_{N_{o}}^{T_{1}}\right]\in \mathbb{R}^{N_{o}\times 2C}$ to get the clusutering results $\mathcal{Y}^{T_{1}}\in \mathbb{R}^{N_{o}}$. The clustering map $Y^{T_{1}} \in \mathbb{R}^{H \times W}$ can be obtained by assigning the label value of $i$-th object $\mathcal{Y}^{T_{1}}_{i}$ back to the pixels belonging to $\Omega_{i}$. 

\par Subsequently, we exchange the image patches within $X^{T_{1}}$ and $Y^{T_{1}}$ to obtain a pseudo-post-event image and associated clustering map, respectively. Specifically, given a particular scale factor $\sigma$, $X^{T_{1}}$ and $Y^{T_{1}}$ are partitioned into $\frac{HW}{\sigma^{2}}$ image patches with a size of $\sigma\times\sigma$ pixels. From left to right and from top to bottom, each image patch will be assigned an index in a sequence $S=\left\{1, 2, \cdots, \frac{HW}{\sigma^{2}}\right\}$. Next, we shuffle this sequence and then pair up adjacent indices in pairs to obtain a set of exchange tuples $\mathcal{T}=\left\{\left(s_{1}, s_{2}\right), \left(s_{3}, s_{4}\right),\cdots, \left(s_{\frac{HW}{\sigma^{2}}-1}, s_{\frac{HW}{\sigma^{2}}}\right)\right\}$, where $s_{i} \in S$ and $s_{i}\neq s_{j}$. Each tuple contains the indices of the two patches to be exchanged. According to $\mathcal{T}$, we exchange the image patches within $X^{T_{1}}$ and $Y^{T_{1}}$ to obtain the pseudo-post-event image $X^{\tilde{T}_{2}}$ and associated clustering map $Y^{\tilde{T}_{2}}$. Change labels $Y^{T_{1}\rightarrow \tilde{T}_{2}}$ are then automatically generated by comparing the clustering maps $Y^{T_{1}}$ and $Y^{\hat{T}_{2}}$. $Y^{T_{1}\rightarrow \tilde{T}_{2}}(i,j)$ is assigned as change class if $Y^{T_{1}}\left(i,j\right) \neq Y^{\tilde{T}_{2}}\left(i,j\right)$. Otherwise,  $Y^{T_{1}\rightarrow \tilde{T}_{2}}(i,j)$ is assigned as non-change class. This process can be formulated as $Y^{T_{1}\rightarrow \tilde{T}_{2}}=Y^{T_{1}} \oplus Y^{\tilde{T}_{2}}$, where $\oplus$ represents the exclusive or (xor) operation. 

\begin{figure*}[!t]
  \centering
  \includegraphics[width=6.75in]{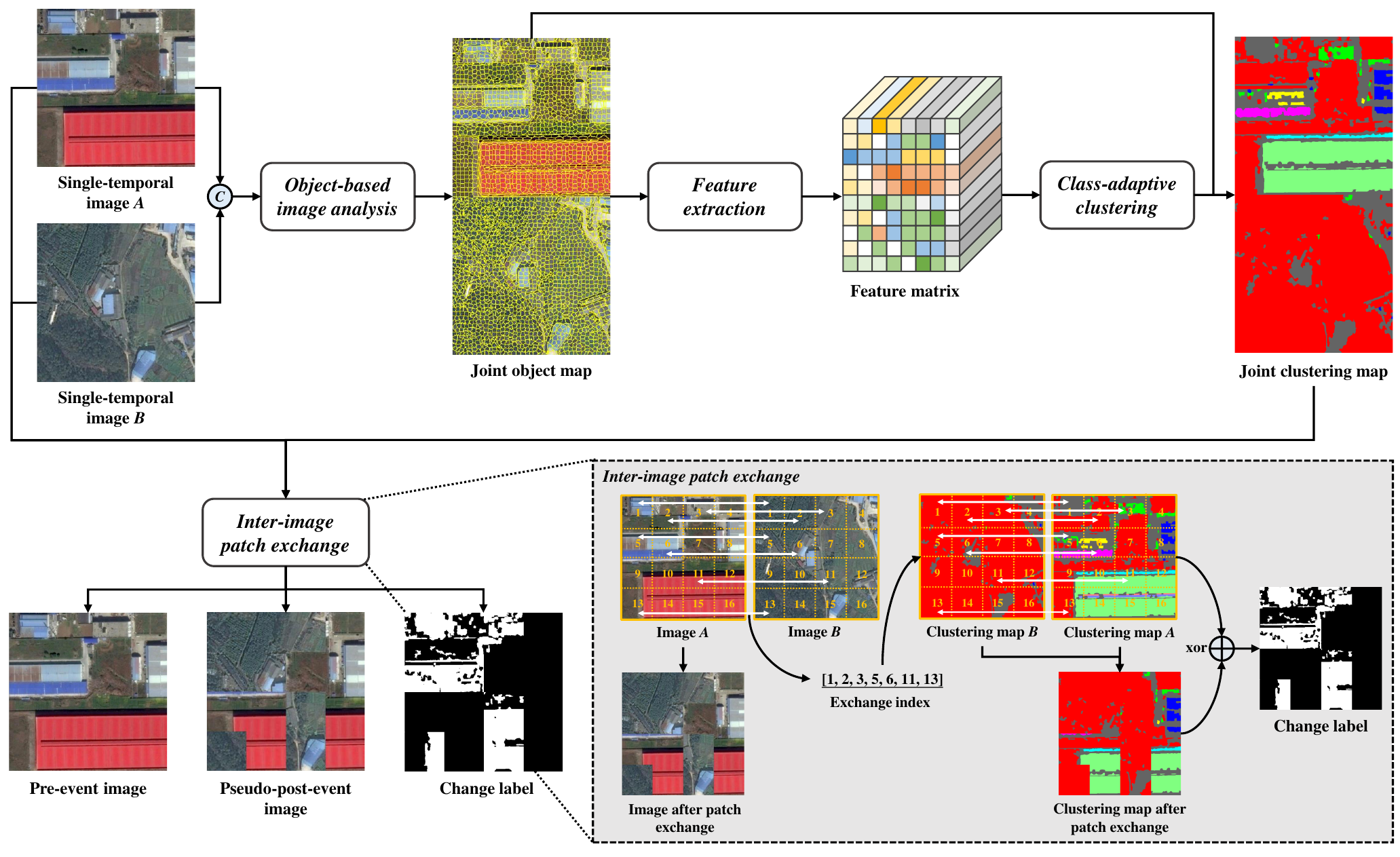}
  \caption{Workflow of the proposed inter-image patch exchange method.}
  \label{IntER_Exc}
\end{figure*}

\par The scale parameter $\sigma$ described above controls the scale of the generated land-cover changes. If $\sigma$ is large, the patches exchanged will be larger, and our method will tend to produce more continuous and larger-scale land-cover changes. Conversely, more fine-grained changes will be obtained. In order to enrich the land-cover change types and obtain different scales of land-cover changes, we propose a multi-scale sampling strategy. That is, we pre-set several different scales $\sigma_{1}$, $\sigma_{2}$,$\cdots$, $\sigma_{n}$; in each iteration of the training stage, our method randomly selects one of the multiple scales for sample generation. Alternatively, we can take a subset $\hat{\mathcal{T}}$ of $\mathcal{T}$ for exchanging only some of the image patches. This way can ensure that a fraction of the unchanged labels in the generated pseudo-bi-temporal training sample is completely accurate.

\subsubsection{Inter-image patch exchange method}
\par One limitation of the proposed intra-image patch exchange method is that the richness of the change types depends on the number of types of land-cover objects in the given images. For example, considering the image in Figure \ref{motivation}, as it only contains three major land-cover objects, i.e., farmland, trees, and buildings, we can only generate changes between these land-cover objects. From this image, it is not possible to generate land-cover changes such as ‘water to vegetation’ or ‘building to railway’. An effective solution is introducing more land-cover changes by exchanging patches with other single-temporal remote sensing images, i.e., inter-image patch exchange. Thus, we further present an inter-image patch exchange method, as shown in Figure \ref{IntER_Exc}. 

\par In order to obtain bi-temporal training samples by exchanging patches between images, a key is to ensure that the label domain of the clustering results is consistent between the two images. Given two unpaired images $X^{T_{1}}$ and $\eta X^{T_{1}}$, we adopt a joint segmentation strategy by first concatenating the two images together and then executing the SLIC algorithm to obtain a joint object map. Then, similar to the step in the intra-image patch exchange method, the features of objects in the joint object map are extracted, and the DBSCAN algorithm is performed to get the clustering maps $Y^{T_{1}}$ and $\eta Y^{T_{1}}$. The above process ensures that the label domain in the clustering maps of the two images is consistent and that the same land-cover objects on both images would have the same label values.

\par Next, we exchange the image patches between $X^{T_{1}}$ and $\eta X^{T_{1}}$, $Y^{T_{1}}$ and $\eta Y^{T_{1}}$. $X^{T_{1}}$ and $\eta X^{T_{1}}$ and their corresponding clustering maps $Y^{T_{1}}$ and $\eta Y^{T_{1}}$ are partitioned into $\frac{HW}{\sigma^{2}}$ image patches. Then, we shuffle the sequence of the patch index and get a subset of it to determine which patches will be exchanged between the two images. After exchanging process, we can generate the pseudo-post-event image $X^{\tilde{T}_{2}}$ with land-cover objects from other image and its associated clustering map $Y^{\tilde{T}_{2}}$. Finally, the change label is obtained by performing the xor operator on the clustering map of $Y^{T_{1}}$ and $Y^{\tilde{T}_{2}}$. Moreover, the inter-image patch exchange method also adopts the multi-scale sampling strategy to generate land-cover changes with different scales. 

\subsection{Simulation of different imaging conditions}
\begin{figure}[!t]
  \centering
  \includegraphics[width=3.3in]{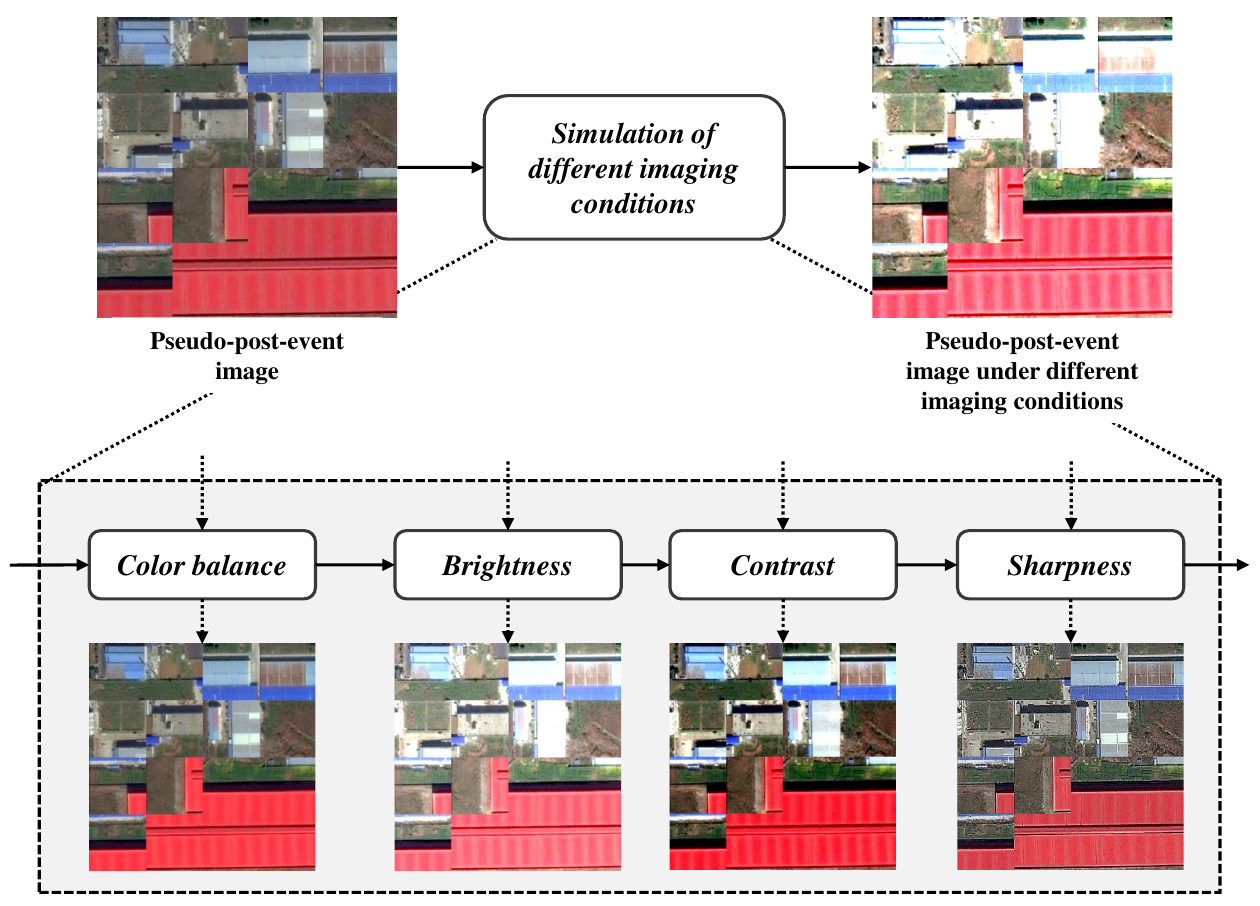}
  \caption{Workflow of simulating different imaging conditions. Here, we show an instance of a pseudo-post-event image after processing and the instances obtained for each of these sub-steps individually.}
  \label{Stimu_DiffIC}
\end{figure}

\par Through the proposed intra- and inter-image patch exchange methods, we can generate paired pseudo-bi-temporal images and corresponding change labels from single-temporal remote sensing images in a simple way without any prior information. In practical change detection scenarios, since the bi-temporal remote sensing images are acquired in different time phases, the pre-event and post-event images usually show obvious visual differences in appearance caused by different imaging conditions, like solar angles, atmospheric conditions, illumination conditions, and sensor calibration \citep{Canty2008}. However, since the pseudo-bi-temporal images in our methods are generated from single-temporal images, the pre-event image and post-event image may not show the above radiometric difference. To address this issue, we propose to simulate different imaging conditions by using commonly used image enhancement methods to introduce radiometric differences for pseudo-bi-temporal image pairs. Figure \ref{Stimu_DiffIC} shows the specific pipeline for simulating different imaging conditions and an example of a generated pseudo-post-event image processed by our pipeline. By adjusting the pseudo-post-event image in color balance, brightness, contrast, and sharpness, we could see that the adjusted image shows an obvious visual difference from the pre-event image, making our generated samples more in line with the actual situation. 

\subsection{Architecture of the deep change detector}\label{sec:3.2}
\begin{figure}[!t]
  \centering
  \includegraphics[width=3.25in]{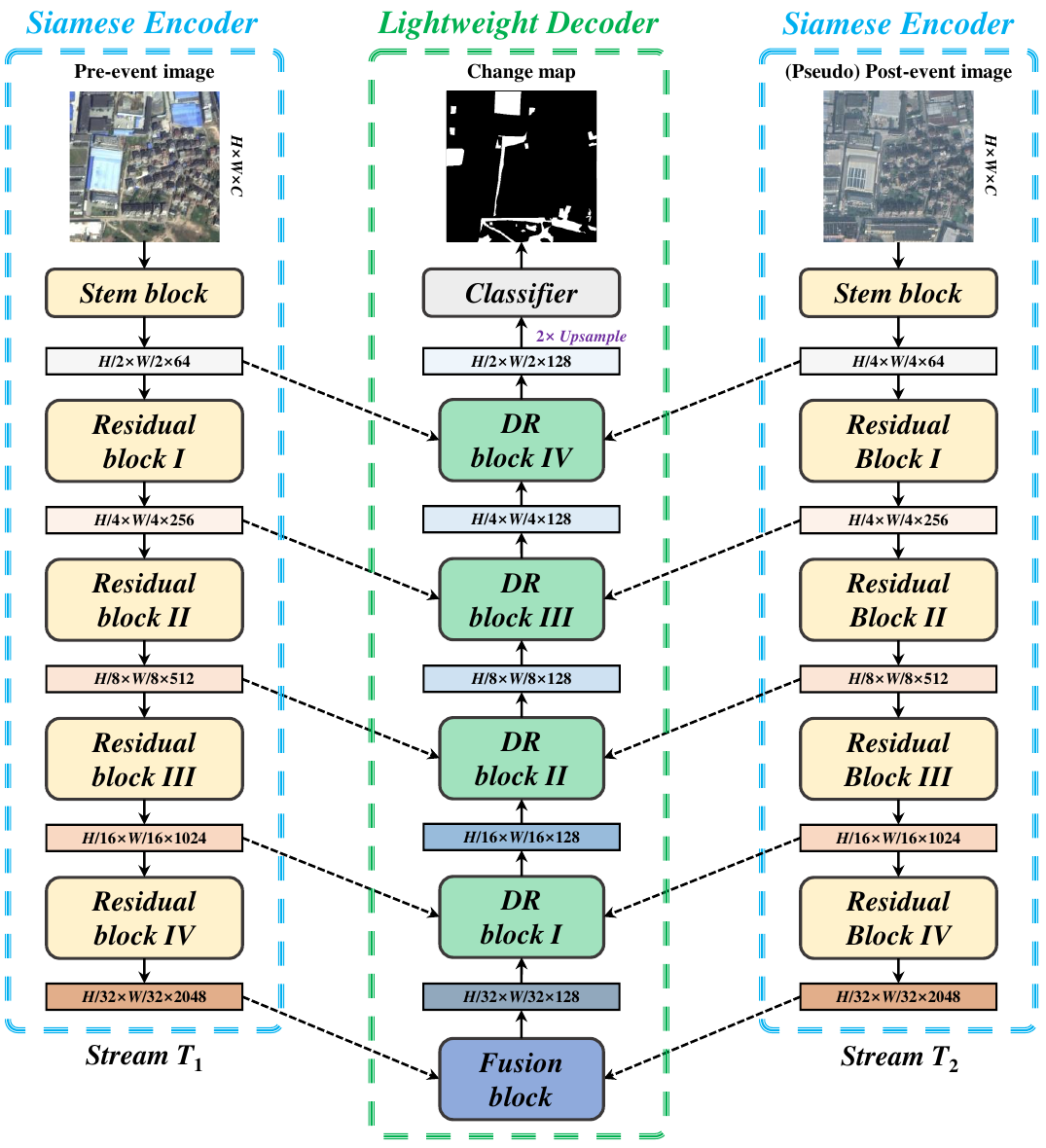}
  \caption{The structure of the proposed deep change detector.}
  \label{DFCN}
\end{figure}

\par Following the generation of pseudo-bi-temporal images and their associated change labels through intra- and inter-image patch exchange methods, a deep change detector can be trained on these samples and used to detect land-cover changes on real bi-temporal images. Compared to the lightweight models designed in most current unsupervised methods \citep{Gong2017, Liu2022, Wu2022Unsupervised}, our framework can allow us to utilize or design deeper and more powerful architectures as detectors. In particular, the fully convolutional networks (FCNs) \citep{long2015fully} have achieved decent performance in vision tasks. To this end, we propose a deep siamese FCN \citep{CayeDaudt2018, Zheng2022} as the change detector in our framework, with the network structure shown in Figure \ref{DFCN}.

\par The proposed network comprises a siamese encoder and a lightweight decoder. To fully extract hierarchy and representative semantic features from input bi-temporal remote sensing images, it is necessary for the network to have a deep encoder. However, training a deep network may pose a challenge due to the vanishing gradient problem. Thus, we employ the residual network (ResNet) \citep{He2016} as the encoder, which reformulates convolutional layers by learning residual functions of the inputs through identity mapping. The original ResNet is designed for the image classification task. We retain its stem block and four residual blocks to make it suitable for extracting features for the downstream change detection task. The stem block consists of a convolutional layer with 7$\times$7 convolutional kernels and stride 2 followed by a batch normalization (BN) layer and rectified linear unit (ReLU) activation function. The residual block comprises a max-pooling layer and several residual units. For our work, we adopt ResNet-50, which has four residual blocks with 3, 4, 6, and 3 residual units, respectively. Each unit consists of stacked 1$\times$1, 3$\times$3, and 1$\times$1 convolutional layers, where a BN layer and ReLU function follow each convolutional layer. A shortcut connection structure is employed for the input and the output of the residual unit to mitigate the vanishing gradient problem. Given that the input to the change detection task is bi-temporal image pairs, the encoder of the proposed change detector consists of two streams. To ensure comparability and reduce parameters, we design the two streams as a siamese architecture that is weight-shared and has identical structures. The feature maps from the four residual blocks in two streams are extracted for the downstream tasks.

\begin{figure}[!t]
  \centering
  \subfloat[]{
    \includegraphics[width=1.55in]{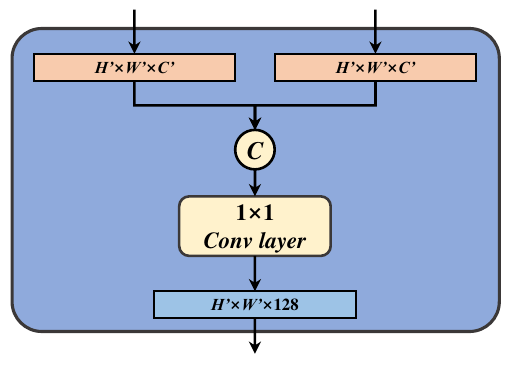}
  \label{fig_first_case}}
  \subfloat[]{
    \includegraphics[width=1.55in]{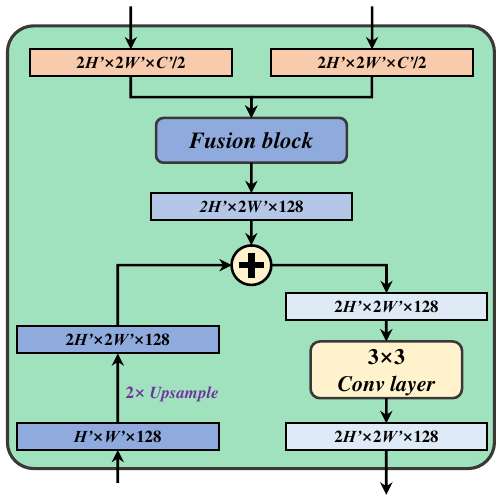}
  \label{fig_second_case}}
 
  \caption{Inner structure of (a) fusion block and (b) detail recovery block in the designed deep change detector.}
  \label{Net_block}
\end{figure}

\par After feature extraction, a lightweight detail recovery network is designed as the decoder to interpret land-cover changes from these extracted multi-level features. Firstly, the two feature maps from residual block IV of two streams are fused by a fusion block consisting of a concatenation operator and a $1\times1$ convolutional layer, as shown in Figure \ref{Net_block}-(a). The features from residual block IV contain abstract and high-level information. Some concrete and local information is required to generate changed objects with accurate boundaries. Thus, four detail recovery (DR) blocks are designed to progressively fuse the features from the remaining three residual blocks and the stem block. The structure of the DR block is shown in Figure \ref{Net_block}-(b). In the DR block, the two finer-resolution feature maps from the shallow residual block are first fused with a fusion block. The coarser-resolution feature map from the previous DR block is scaled up to twice the size to match the size of the fused finer-resolution feature map. The two feature maps are then merged with an element-wise addition operation and smoothed with a 3$\times$3 convolutional layer. Finally, the feature map with the finest resolution is generated after processed by four DR blocks. We upsample its spatial resolution by a factor of 2 and apply a 1$\times$1 convolutional layer as the classifier to predict the land-cover change map from the upsampled features.

\par Note that the main motivation of this work is trying to train effective deep change detectors utilizing unlabeled and unpaired single-temporal remote sensing images. Thus, the network presented here does not have some advanced modules or sophisticated structures. However, we also verified the generalizability of our approach to other network architectures, including the Transformer architecture, in the experiments in Section \ref{sec:4.3.3}. 

\subsection{Optimization}
\subsubsection{Network training based on temporal symmetry}
\par Finally, we optimize the change detector on the pseudo-bi-temporal image pairs generated from arbitrary unlabelled remote sensing images. Since change detection can be seen as a special semantic segmentation task, we directly utilize the cross-entropy loss to optimize the change detector as

\begin{equation}
    \mathcal{L}^{T_{1}\rightarrow \tilde{T}_{2}}_{ce} = \sum^{H}_{h=1} \sum^{W}_{w=1}\sum_{c=1}^{2}Y^{T_{1}\rightarrow \tilde{T}_{2}}(h,w,c) \mathrm{log} P^{T_{1}\rightarrow \tilde{T}_{2}}\left(h,w,c\right)
\end{equation}
where $Y^{T_{1}\rightarrow \tilde{T}_{2}}$ is the change label generated from arbitrary unlabelled single-temporal images and $P^{T_{1}\rightarrow \tilde{T}_{2}}$ is the {final} output of the deep change detector, i.e., the class probability map after the softmax activation function.

\par Since the pseudo-post-event images are obtained by exchanging image patches, there is a feature discontinuity in the pseudo-post-event image compared to the pre-event image, which does not match the situation of real bi-temporal remote sensing images. This may introduce bias to the model since this spatial discontinuity exists only in the input of stream $T_{2}$, thereby negatively affecting the performance of the change detector trained on these samples in detecting land-cover changes on real bi-temporal images. We here adopt a temporal-symmetric loss function to reduce the negative effect caused by such discontinuity on the detectors. This loss function is based on the fact that binary change detection is temporal symmetric \citep{Zheng2022}. For a bi-temporal image-pair $X^{T_{1}}$ and $X^{T_{2}}$,the predicted class probability maps $P^{T_{1}\rightarrow T_{2}}$ and $P^{T_{2}\rightarrow T_{1}}$ should be the same. Therefore, it is implemented by swapping the pre-event image and pseudo-post-event image when inputting them into the change detector, formulated as

\begin{equation}
   \mathcal{L}_{sym} = \mathcal{L}^{T_{1}\rightarrow \tilde{T}_{2}}_{ce} + \mathcal{L}^{\tilde{T}_{2}\rightarrow T_{1}}_{ce},
\end{equation} 
where $\mathcal{L}^{\tilde{T}_{2}\rightarrow T_{1}}_{ce}$ is calculated by inputting the pseudo-post-event image $X^{\tilde{T}_{2}}$ to stream $T_{1}$ and the pre-event image $X^{T_{1}}$ to stream $T_{2}$. In this way, both streams can get samples with spatial continuity, thereby reducing the negative effect caused by the spatial discontinuity problem.

\subsubsection{Self- and semi-supervised learning based on pseudo labels}\label{sec:3.4}
\par Once we have optimized the change detector on the generated samples, we can use it to detect land-cover changes on real bi-temporal images. In practical application scenarios, we can use the prediction results of the network as supervisory signals to further optimize our network, i.e., self-supervised learning \citep{Zou_2018_ECCV}. Here, we employ a self-supervised learning approach to improve the performance of our framework by using the change detector's prediction results as pseudo-labels. To assure the accuracy of pseudo-labels, we set a threshold $\tau$ to select high-confident pseudo-labels as

\begin{equation}
   \tilde{Y}^{T_{1}\rightarrow T_{2}}=\left\{
\begin{aligned}
&\operatornamewithlimits{argmax}_{c} P^{T_{1}\rightarrow T_{2}}, \ \max_{c} P^{T_{1}\rightarrow T_{2}} > \tau \\
&ignore, \ \mathrm{otherwise}
\end{aligned}\right.
\label{eq:5}
\end{equation}
where $ignore$ means that the value will not be involved in the loss calculation.

\par Moreover, another common scenario in practical applications is that there is a small fraction of labeled bi-temporal image pairs and a large number of unlabelled and unpaired remote sensing images, namely semi-supervised scenarios. Our method provides a simple way to exploit these unlabelled and unpaired images to facilitate change detection. Specifically, we propose a semi-supervised learning framework based on pseudo-labels. We take an alternating optimization approach, optimizing the change detector on real bi-temporal samples and then training the change detector on the generated pseudo-bi-temporal samples. Since the network is provided with real change supervision information, we can also use the network's predictions on the pseudo-bi-temporal images to refine the associated change labels, thereby improving the quality of the generated supervision information as

\begin{equation}
  \tilde{Y}^{T_{1}\rightarrow \tilde{T}_{2}}=\left\{
\begin{aligned}
&Y^{T_{1}\rightarrow \tilde{T}_{2}}, \ Y^{T_{1}\rightarrow \tilde{T}_{2}}=\operatornamewithlimits{argmax}_{c} P^{T_{1}\rightarrow \tilde{T}_{2}}\\
&ignore, \ \mathrm{otherwise}
\end{aligned}\right.
\label{eq:6}
\end{equation}
where pixels in change labels generated using I3PE will be used for training the change detector only if they remain the same as the predicted values of the change detector.

\section{Experiments}\label{sec:4}
\par In this section, we conduct extensive experiments to validate the effectiveness and usefulness of the I3PE framework. On the two large-scale benchmark datasets, we conduct experiments including performance comparisons with other methods, ablation studies, hyperparameter discussions, generalization validation, semi-supervised learning experiments, and efficiency comparison. On the Wuhan dataset, we additionally validate the effectiveness of our method in practical application scenarios.

\subsection{Experimental setup}
\subsubsection{Implementation details}
\par We implement our framework with Python and some of its libraries, mainly including PyTorch\footnote{https://pytorch.org/} and scikit-learn\footnote{https://scikit-learn.org/stable/}. The proposed deep change detector is implemented with PyTorch. The SLIC and DBSCAN algorithms are implemented with scikit-learn. When training the change detector on the pseudo-bi-temporal images, we utilize the SGD as the optimizer with a learning rate of 1$e^{-3}$, momentum of 0.9, and a weight decay of 5$e^{-4}$. For the subsequent self-supervised learning stage, we utilize AdamW \citep{loshchilov2017decoupled} as the optimizer with a learning rate of 1$e^{-4}$, and a weight decay of 5$e^{-4}$. For the number of objects generated by the SLIC algorithm, we set 1,000 and 2,000 on the SYSU dataset, and 4,000 and 8,000 on the SECOND dataset (the numbers before and after correspond to the intra-image and inter-image patch-exchange methods, respectively). We will discuss the critical hyperparameters related to image patch exchange methods and self-supervised learning in Section \ref{sec:4.3.2}.

\par The main goal of our framework is to train a change detector with decent performance from unpaired and unlabeled remote sensing images. Therefore, in our experiments, we mix the pre- and post-event images from the training set of the experimental datasets directly without further pairing to obtain a single-temporal image training set. Arbitrary single-temporal images are then used as input to our framework for generating pseudo-bi-temporal images and the corresponding change labels for training the deep network. After training the change detector, we test it on real bi-temporal images and reference maps from the test set. 

\par The source code of our framework will be open-sourced for replication and reference for subsequent research, thus contributing to the field of remote sensing\footnote{The source code of this work will be open-sourced in https://github.com/ChenHongruixuan/I3PE}.

\subsubsection{Evaluation metrics}
\begin{figure}[!t]
  \centering
  \includegraphics[width=3.3in]{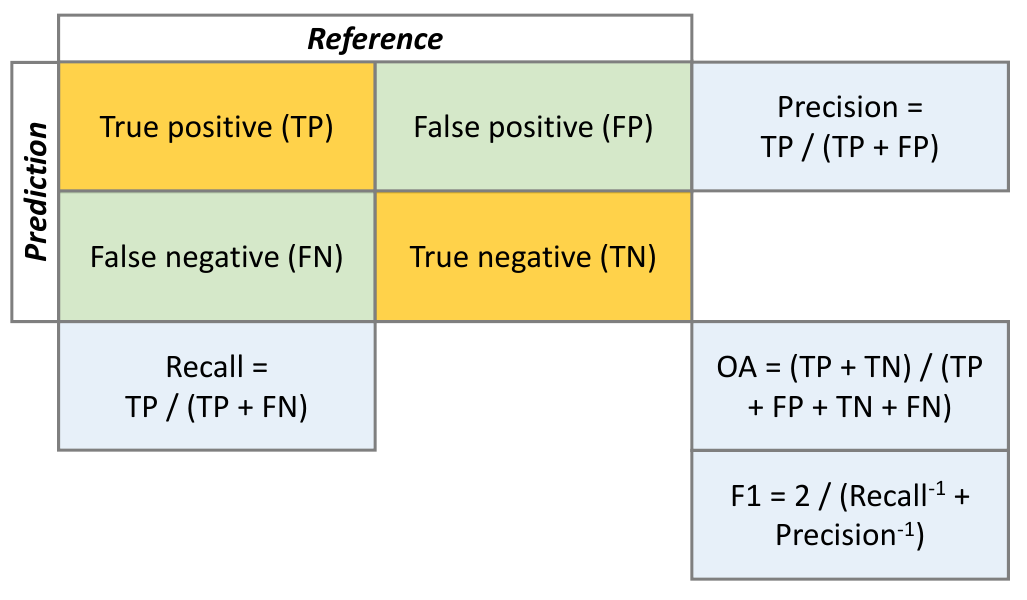}
  \caption{The confusion matrix and evaluation metrics used for accuracy assessment.}
  \label{eval_metric}
\end{figure}

\par Four evaluation metrics are used for accuracy assessment. They are recall rate, prevision rate, overall accuracy (OA), and F1 score. On the test set, we calculate the confusion matrix consisting of the numbers of the true positive (TP), true negative (TN), false positive (FP), and false negative (FN) pixels. Then, as shown in Figure \ref{eval_metric}, the evaluation metrics are calculated as follows:
\begin{enumerate}
    \item Recall rate represents the ratio of correctly detected changed pixels to all changed pixels in the test set:
    
    \begin{equation}
    Recall = \frac{TP}{TP+FN}.
    \label{eq:Rec}
    \end{equation} 
    \item Precision rate indicates the ratio of pixels that are truly changed to all pixels that are detected as changed:
    
    \begin{equation}
    Precision = \frac{TP}{TP+FP}.
    \label{eq:Pre}
    \end{equation}
    \item Overall accuracy (OA) is defined as the ratio of correctly detected pixels to all the pixels in the entire test set:

     \begin{equation}
    OA = \frac{TP+TN}{TP+TN+FP+FN}.
    \label{eq:OA}
    \end{equation}
    \item F1 score is the harmonic mean of the precision and recall rates. As the change detection task is usually a skewed class task, the percentage of change pixels is relatively low. OA does not account for such class imbalance and would lead to misinterpretations. In comparison, F1 score provides a better performance measure for change detectors. F1 score can be calculated using the following formula:
    
    \begin{equation}
    F_{1} = \frac{2}{Recall^{-1}+Precision^{-1}}.
    \label{eq:F1}
    \end{equation}
\end{enumerate}

\subsubsection{Comparison methods}
\par Here, we compare our framework with some representative unsupervised multi-temporal change detection methods to verify its effectiveness. These comparison methods are briefly introduced as follows:
\begin{enumerate}
    \item CVA \citep{Sharma2007} is the most widely adopted benchmark method in the field of unsupervised change detection. The Euclidean distance between the spectral signatures of each pixel in multi-temporal images is calculated. A threshold segmentation algorithm is executed to obtain the land-cover changes.
    \item IRMAD\footnote{http://www.imm.dtu.dk/~alan/software.html} \citep{Nielsen2007} is a transformation-based unsupervised change detection model which aims at finding the most relevant feature space for unchanged pixels in multi-temporal image pairs based on the canonical correlation analysis algorithm. An iterative reweighting scheme is designed to improve detection performance. 
    \item ISFA\footnote{http://sigma.whu.edu.cn/resource.php} \citep{Wu2014} is another effective image transformation method. By solving the slow feature analysis (SFA) problem, the method can find a feature space in which the pixel values of unchanged pixels are suppressed and the pixel values of changed pixels are highlighted.
    \item OBCD \citep{XIAO2016402} is a kind of representative unsupervised change detection method that improves detection accuracy by changing the basic unit of analysis for change detection from pixels to objects consisting of many homogeneous pixels.
    \item DCAE \citep{Bergamasco2022Unsupervised} is an unsupervised deep learning model consisting of an encoder and a decoder, both of which are composed of some convolutional layers. DCAE can extract hierarchical features for detecting land-cover changes by setting reconstructing the bi-temporal images as the optimization objective.
    \item DCVA\footnote{https://github.com/sudipansaha/dcvaVHROptical}\citep{Saha2019} is an unsupervised change detection method that utilizes a pre-trained DCNN to extract deep spatial-spectral features from bi-temporal images and then performs the CVA algorithm on the binarized features to detect land-cover changes.  
    \item DSFA\footnote{https://github.com/rulixiang/DSFANet} \citep{Du2019a} is an improved variant of the SFA approach. DSFA utilizes a dual-stream deep neural network to extract deep features from bi-temporal images and solves the SFA problem on the input bi-temporal images to optimize the parameters of the deep neural network and SFA model. 
    \item KPCA-MNet\footnote{https://github.com/ChenHongruixuan/KPCAMNet} \citep{Wu2022Unsupervised} is an unsupervised deep model that trains several KPCA convolutional layers to extract features from bi-temporal images and maps extracted features to a polar domain to detect land-cover changes.  
\end{enumerate}

\subsection{Detection performance comparison}
\subsubsection{Change detection results on SYSU dataset}

\begin{figure*}[!t]
  \centering
  \includegraphics[width=6in]{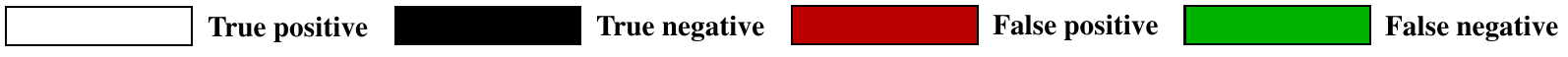}
  \\
  \subfloat[]{
    \includegraphics[height=3in]{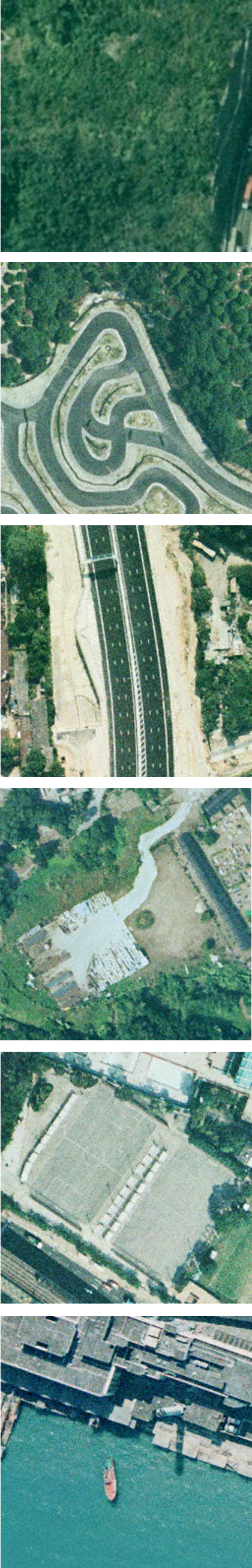}
  \label{fig_second_case}}
  \hfil
  \subfloat[]{
    \includegraphics[height=3in]{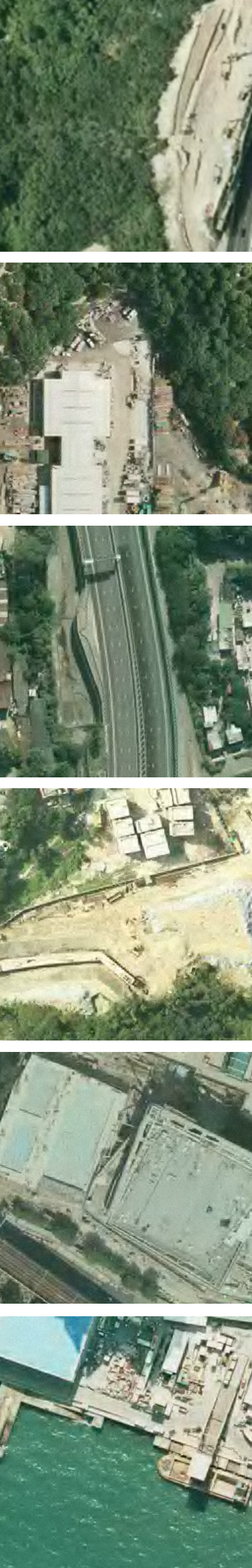}
  \label{fig_first_case}}
  \hfil
  \subfloat[]{
    \includegraphics[height=3in]{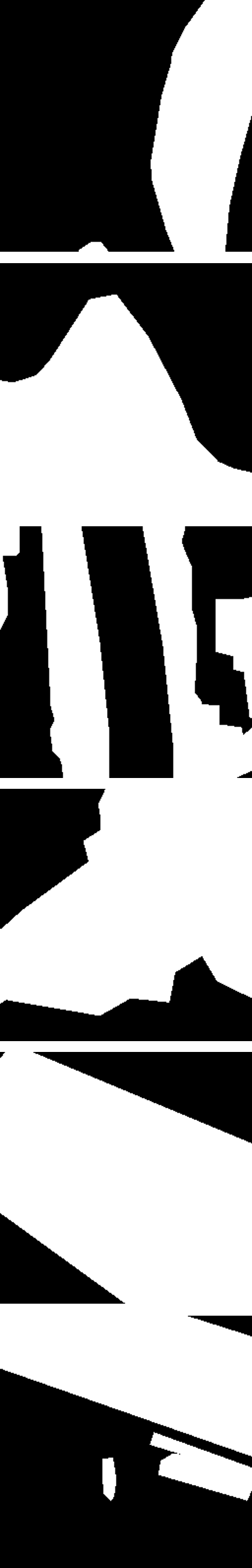}
  \label{fig_first_case}}
  \hfil
  \subfloat[]{
    \includegraphics[height=3in]{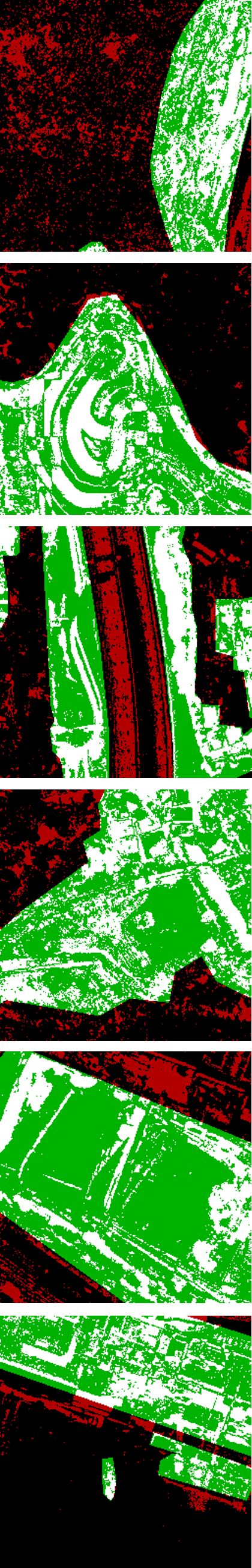}
  \label{fig_first_case}}
  \hfil
  \subfloat[]{
    \includegraphics[height=3in]{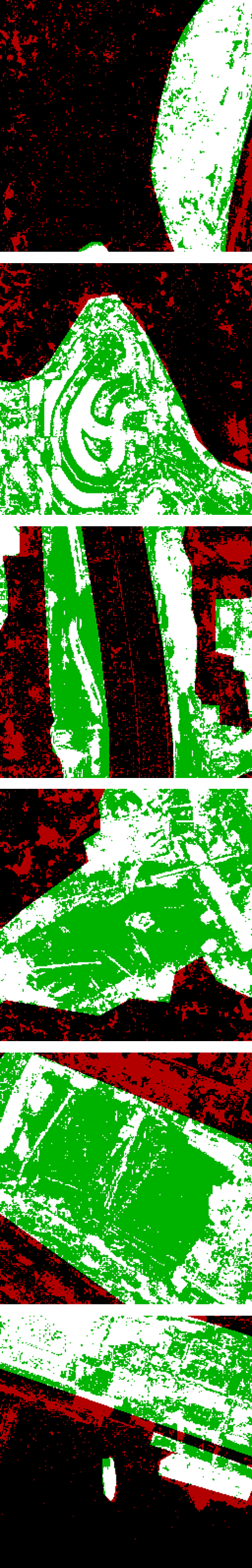}
  \label{fig_first_case}}
   \hfil
  \subfloat[]{
    \includegraphics[height=3in]{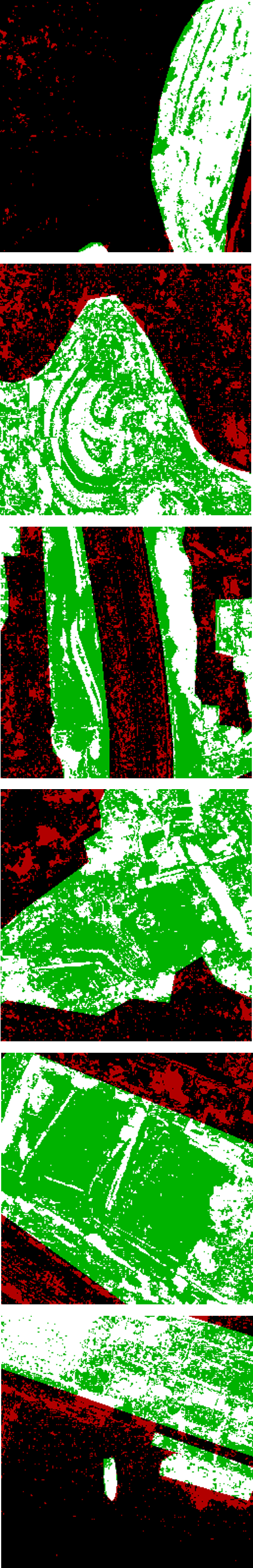}
  \label{fig_first_case}}
 \hfil
  \subfloat[]{
    \includegraphics[height=3in]{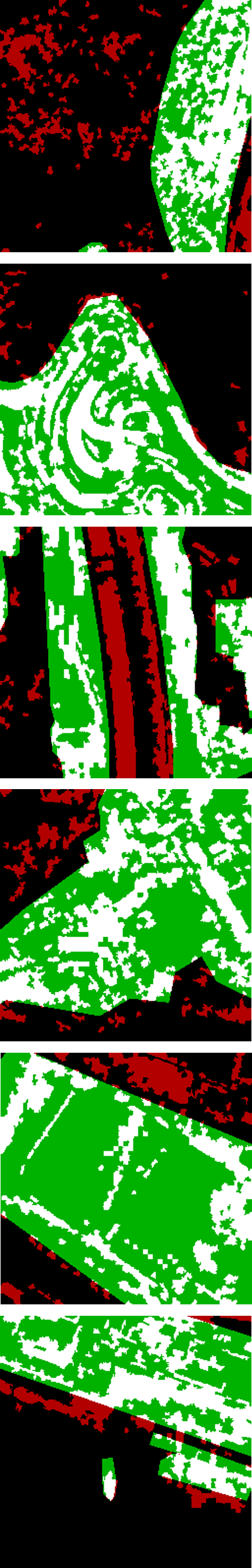}
  \label{fig_first_case}}
  \hfil
  \subfloat[]{
    \includegraphics[height=3in]{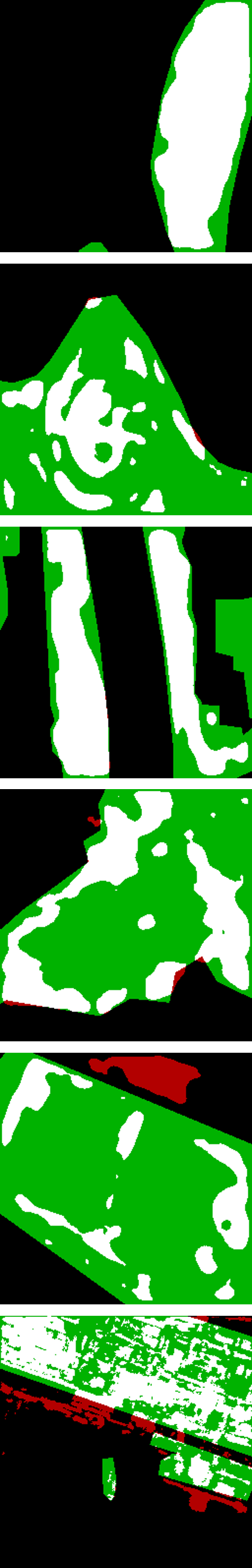}
  \label{fig_first_case}}
  \hfil
  \subfloat[]{
    \includegraphics[height=3in]{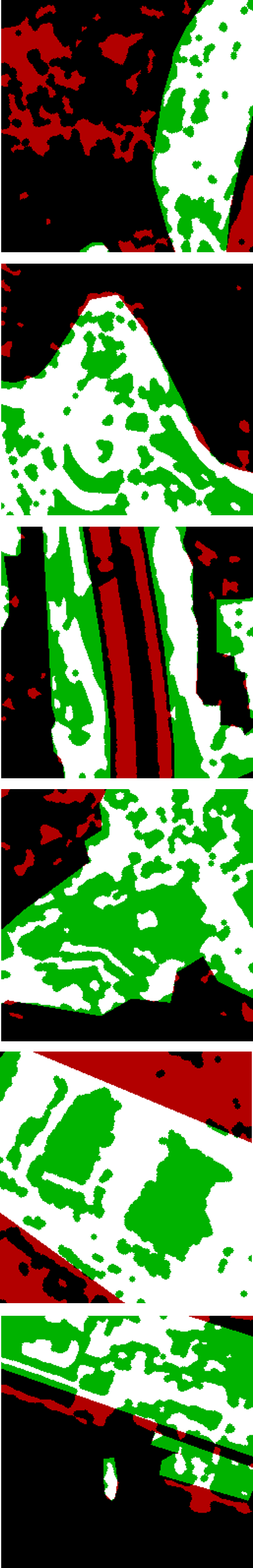}
  \label{fig_first_case}}
    \hfil
  \subfloat[]{
    \includegraphics[height=3in]{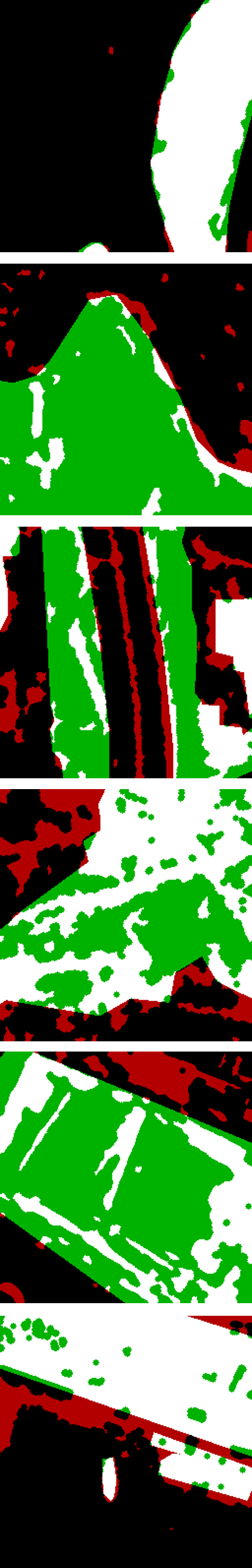}
  \label{fig_first_case}}
  \hfil
  \subfloat[]{
    \includegraphics[height=3in]{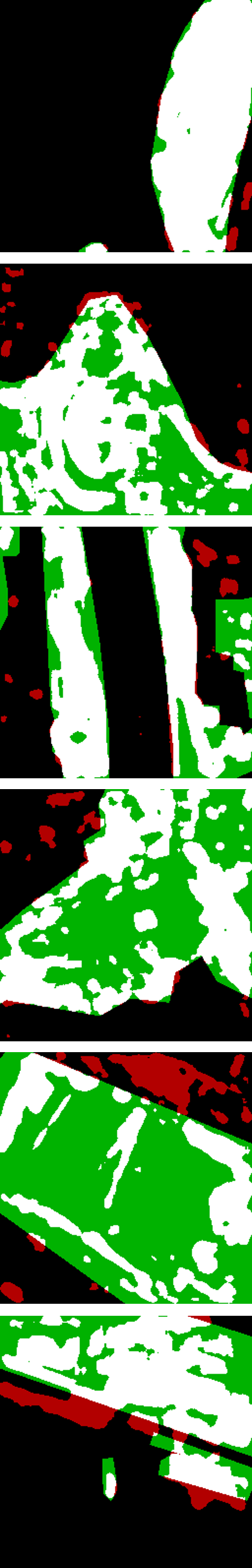}
  \label{fig_first_case}}
  \hfil
  \subfloat[]{
    \includegraphics[height=3in]{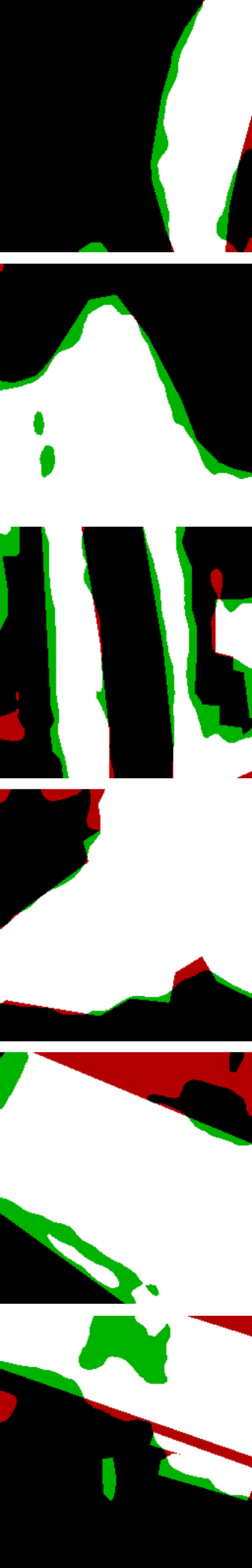}
  \label{fig_first_case}}
  \caption{Some change maps obtained by different methods on the test set of the SYSU dataset. (a) Pre-event image. (b) Post-event image. (c) Change reference map. (d) CVA. (e) IRMAD. (f) ISFA. (g) OBCD. (h) DCAE. (i) DCVA. (j) DSFA. (k) KPCA-MNet. (i) I3PE. In the obtained change maps, white represents TP; black represents TN; red represents FP; green represents FN. Zoom in for a better visual effect.}
  \label{CD_SYSU}
\end{figure*}

\par Figure \ref{CD_SYSU} shows some land-cover change maps in the test set of the SYSU dataset obtained by our framework and the eight comparison methods. Firstly, due to solely utilizing spectral information, the change maps obtained by CVA, IRMAD, and ISFA have many FP and FN pixels. OBCD reduces the number of FP pixels by utilizing object-based analysis instead of pixel-based analysis. However, it only utilizes low-level image features, resulting in missed detection of certain changed pixels. 

\par In contrast, the four deep learning-based methods demonstrate superior performance in detecting land-cover changes accurately with fewer FP pixels and more complete changed regions. Nonetheless, some change events remain challenging to detect for these methods. For instance, the fifth example shows the change event of newly constructed urban buildings. However, we can see that the new buildings in the post-event image and the impervious surface in the pre-event image show similar spectral features. This similarity poses a problem for most comparison methods, except DCVA, which leverages a deep network pre-trained on the ImageNet dataset to extract semantic features. However, the change map obtained by DCVA still has many FN pixels. In comparison, the change map yielded by our framework shows very few FP and FN pixels. This indicates that our framework can make change detectors learn information on complex land-cover changes from arbitrary unlabelled images.

\begin{table*}[width=2.0\linewidth,cols=5,pos=t]
\caption{Accuracy assessment for different unsupervised change detection approaches on the SYSU dataset. The table highlights the highest values in bold, and the second-highest results are underlined.}\label{acc_SYSU} 
\begin{tabular*}{\tblwidth}{@{} LLLLL @{} }
\toprule
Method	&	Recall	& Precision		& OA		& F1 score \\
\midrule
CVA&	\underline{0.6213}&	0.2428&	0.4539&	0.3492\\
IRMAD&	0.3851&	0.3569&	0.6914&	0.3705\\
ISFA&	0.3756&	0.3635&	0.6977&	0.3695\\
OBCD&	0.4190&	0.3912&	0.7091&	0.4046\\
DCAE&	0.3921&	\textbf{0.4984}&	\textbf{0.7636}&	0.4390\\
DCVA&	0.5109&	0.3942&	0.6995&	0.4450\\
DSFA&	0.5468&	0.3311&	0.6326&	0.4125\\
KPCA-MNet&	0.5022&	0.4047&	0.7084&	\underline{0.4482}\\
I3PE&	\textbf{0.7119}&	\underline{0.4544}&	\underline{0.7305}&	\textbf{0.5547}\\
\bottomrule
\end{tabular*}
\end{table*}

\par Table \ref{acc_SYSU} lists the overall quantitative results of our framework and comparison methods on the test set of the SYSU dataset. The benchmark unsupervised algorithm CVA obtains an F1 score of 0.3492.By converting the raw spectral features into a new feature space, IRMAD and ISFA improve the detection performance, exhibiting an improvement in F1 scores by 2.13$\%$ and 2.03$\%$, respectively, compared to CVA. By incorporating spatial contextual information, OBCD produces an F1 score of 0.4046. The deep learning-based approaches provide more accurate detection results by leveraging deep networks to extract representative spatial-spectral features. DCAE has the best value in OA and 0.4390 in the F1 score. By utilizing several KPCA convolutional layers to extract features and a 2-D polar domain to compress change information, KPCA-MNet yields the second-best F1 score. 

\par In contrast, our framework achieves the best results in both recall rate and F1 score and the second-best results in precision rate and OA. Our method shows a considerable improvement in the F1 score of 10.65$\%$ compared to KPCA-MNet, one of the most advanced unsupervised change detection algorithms that achieves the second-highest accuracy on the SYSU dataset, fully demonstrating the superiority of our method for unsupervised change detection.

\subsubsection{Change detection results on SECOND dataset}

\begin{figure*}[!t]
  \centering
  \includegraphics[width=6in]{figs/SYSU_Legend.pdf}
  
  \subfloat[]{
    \includegraphics[height=2.95in]{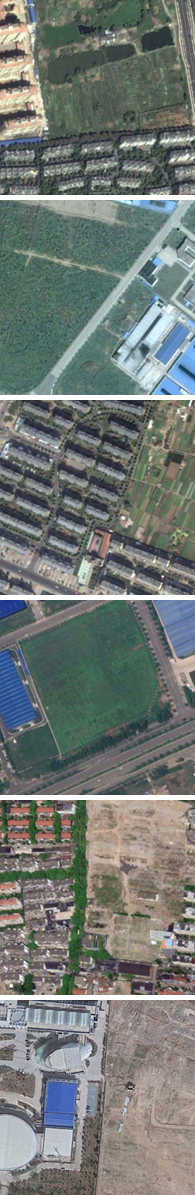}
  \label{fig_second_case}}
  \hfil
  \subfloat[]{
    \includegraphics[height=2.95in]{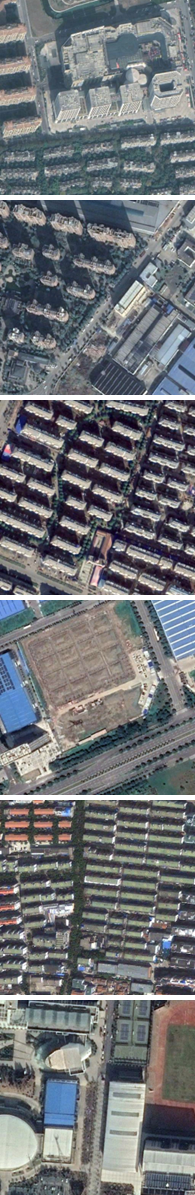}
  \label{fig_first_case}}
  \hfil
  \subfloat[]{
    \includegraphics[height=2.95in]{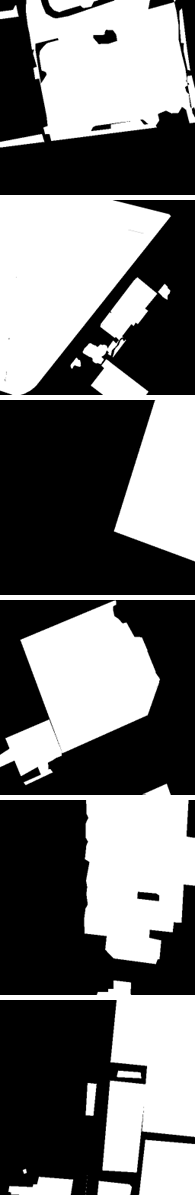}
  \label{fig_first_case}}
  \hfil
  \subfloat[]{
    \includegraphics[height=2.95in]{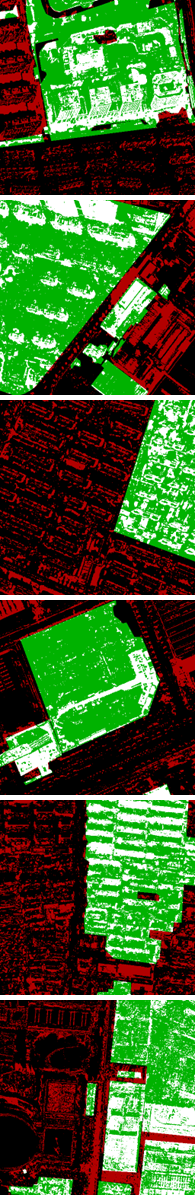}
  \label{fig_first_case}}
  \hfil
  \subfloat[]{
    \includegraphics[height=2.95in]{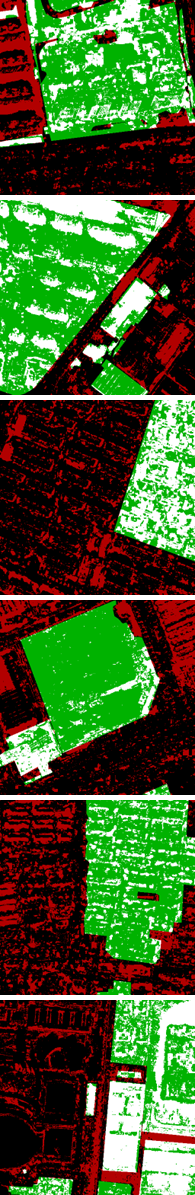}
  \label{fig_first_case}}
   \hfil
  \subfloat[]{
    \includegraphics[height=2.95in]{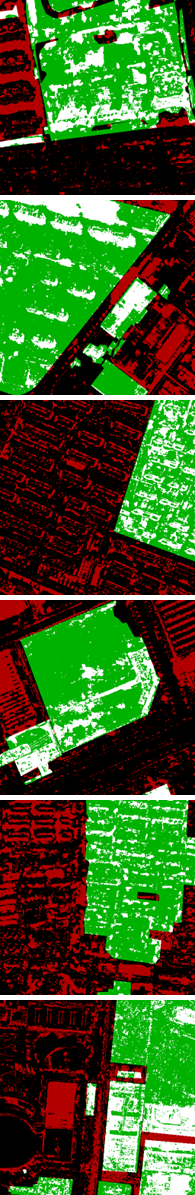}
  \label{fig_first_case}}
    \hfil
  \subfloat[]{
    \includegraphics[height=2.95in]{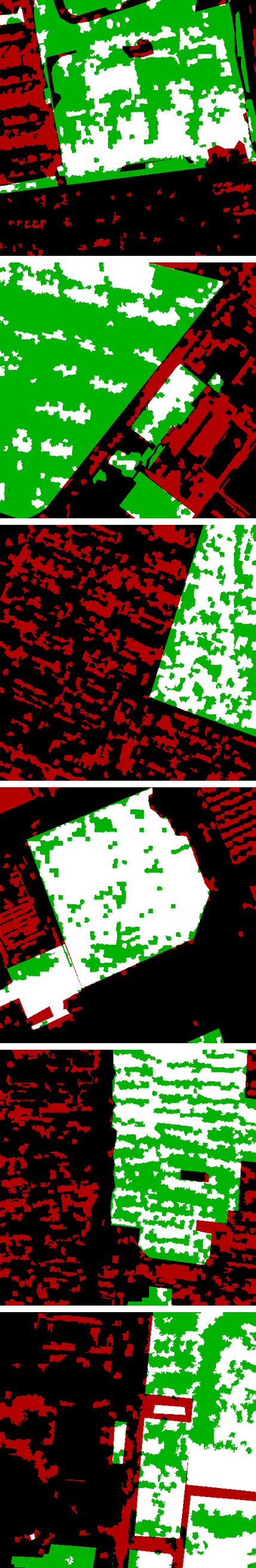}
  \label{fig_first_case}}
  \hfil
  \subfloat[]{
    \includegraphics[height=2.95in]{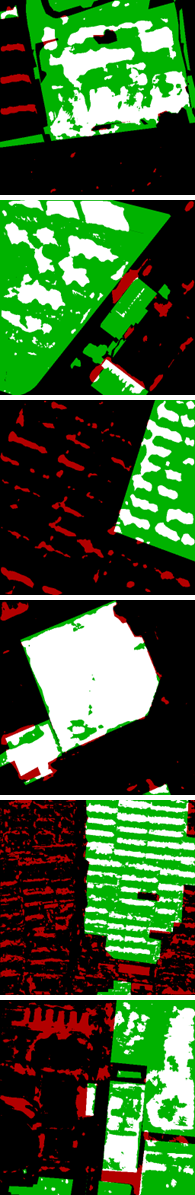}
  \label{fig_first_case}}
  \hfil
  \subfloat[]{
    \includegraphics[height=2.95in]{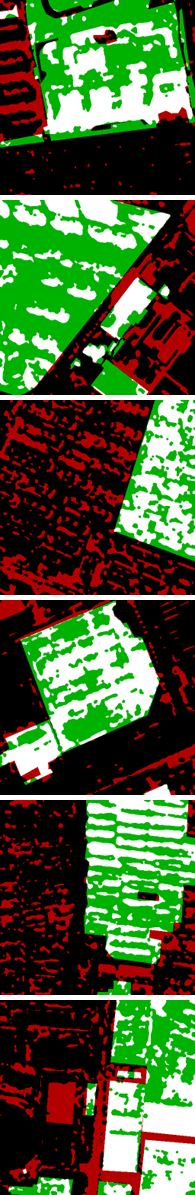}
  \label{fig_first_case}}
    \hfil
  \subfloat[]{
    \includegraphics[height=2.95in]{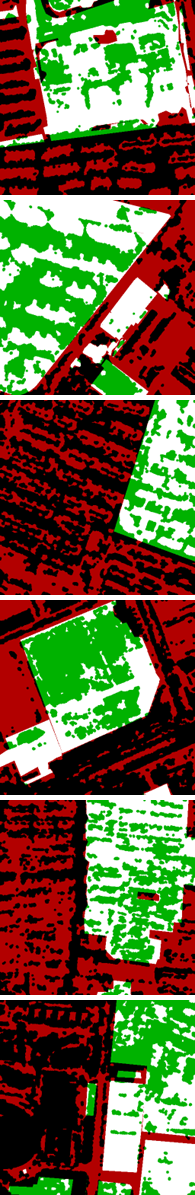}
  \label{fig_first_case}}
  \hfil
  \subfloat[]{
    \includegraphics[height=2.95in]{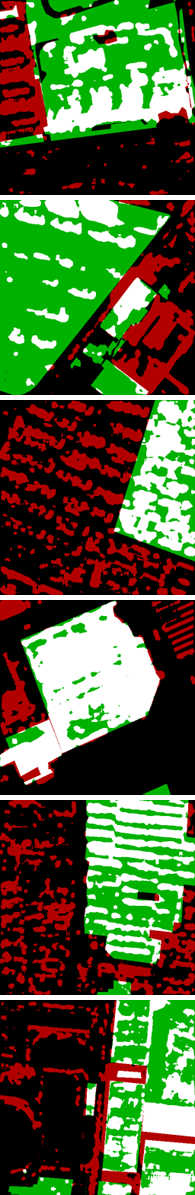}
  \label{fig_first_case}}
  \hfil
  \subfloat[]{
    \includegraphics[height=2.95in]{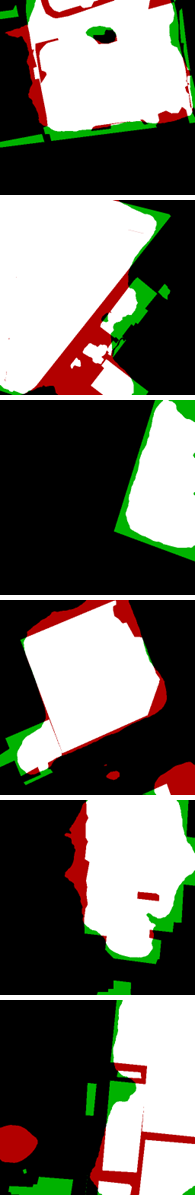}
  \label{fig_first_case}}
  \caption{Some change maps obtained by different methods on the test set of the SECOND dataset. (a) Pre-event image. (b) Post-event image. (c) Change reference map. (d) CVA. (e) IRMAD. (f) ISFA. (g) OBCD. (h) DCAE. (i) DCVA. (j) DSFA. (k) KPCA-MNet. (i) I3PE. In the obtained change maps, white represents TP; black represents TN; red represents FP; green represents FN. Zoom in for a better visual effect.}
  \label{CD_SECOND}
\end{figure*}

\par Figure \ref{CD_SECOND} shows some change maps obtained by different methods on the test set of the SECOND dataset. Compared to the SYSU dataset, the SECOND dataset covers more study areas, encompasses more complex scenarios, and includes more change events. Due to the scenes covered by the image pairs becoming complex and heterogeneous, CVA could only correctly detect a few changed areas, with numerous FP and FN pixels in the obtained change maps. The change maps obtained by IRMAD and ISFA for the six multi-temporal image pairs displayed in Figure \ref{CD_SECOND} do not seem to be more accurate than CVA. The advantages of introducing spatial contextual information are shown in more complex change scenarios. OBCD yields more accurate change maps in these six examples compared to the first three methods. Especially in the fourth example, the main change event is the vegetation to land before construction. OBCD detects a relatively complete changed area with fewer FP and FN pixels in this example. However, the low-level features are insufficient to cope with the various complex ground conditions in the SECOND dataset, so the change maps obtained by OBCD are still not accurate in the other examples. 

\par The four deep learning-based comparison methods produce visually better change maps. However, as they are not given any land-cover change information to supervise, the features they extract may not be suitable to cope with certain practical detection scenarios. In the third instance, all deep learning-based comparison methods incorrectly detect shadows cast by high-rise buildings as changes and fail to detect the emerging buildings on the right side completely. In contrast, by generating supervised information of land-cover changes via exchanging image patches, our framework can make the change detector detect the land-cover changes accurately and yield change maps with very few FP and FN pixels in the six illustrated examples.

\begin{table*}[width=2.0\linewidth,cols=5,pos=t]
\caption{Accuracy assessment for different unsupervised change detection approaches on the SECOND dataset. The table highlights the highest values in bold, and the second-highest results are underlined.}\label{acc_SECOND} 
\begin{tabular*}{\tblwidth}{@{} LLLLL @{} }
\toprule
Method	&	Recall	& Precision		& OA		& F1 score\\
\midrule
CVA&	\textbf{0.6350}&	0.1967&	0.4332&	0.3003\\
IRMAD&	0.4360&	0.2856&	0.6829&	0.3451\\
ISFA&	0.3677&	0.2982&	0.7130&	0.3293\\
OBCD&	0.4074&	0.2956&	0.7005&	0.3426\\
DCAE&	0.3142&	\underline{0.3566}&	\textbf{0.7600}&	0.3340\\
DCVA&	0.4872&	0.2958&	0.6795&	\underline{0.3681}\\
DSFA&	0.5194& 	0.2419&	0.5961&	0.3301\\
KPCA-MNet&	0.4852&	0.2951&	0.6793&	0.3670\\
I3PE&	\underline{0.5525}&	\textbf{0.3628}&	\underline{0.7283}&	\textbf{0.4380}\\
\bottomrule
\end{tabular*}
\end{table*}

\par The quantitative results of our framework and comparison methods are reported in Table \ref{acc_SECOND}. Since it is more challenging to detect land-cover changes on the SECOND dataset, the accuracy of all methods is reduced on the SECOND dataset compared to that on the SYSU dataset. The F1 score for the benchmark method CVA is 0.3003. Due to using two projection matrices, IRMAD can find a better feature space to highlight the change information than ISFA. As a result, IRMAD achieves a 1.58$\%$ improvement in F1 score over ISFA on the SECOND dataset. In addition, the improvement of the deep learning-based methods over the benchmark method CVA is not as pronounced as on the SYSU dataset. The improvement in F1 scores for the four deep learning-based methods ranged from 2.98$\%$ to 6.78$\%$. DCVA received the second highest F1 score of 0.3681 because it uses a deep CNN pre-trained on the ImageNet, which is suitable for processing remote sensing images with complex scenes. 

\par Finally, our I3PE framework achieved the highest F1 score of 0.4380, an improvement of 13.77$\%$ compared to the benchmark method CVA and 6.99$\%$ compared to the SOTA method DCVA. The comparisons on both datasets demonstrate the superiority of our proposed framework for detecting land-cover changes in different scenarios and the validity of our motivation to train an effective change detector from unlabelled and unpaired remote sensing images by exchanging image patches.

\subsection{Discussion}
\par In the last subsection, we compared our method to some representative and SOTA unsupervised change detection models on two large-scale datasets. The superiority of our approach in unsupervised change detection is demonstrated. In this subsection, we delve further into the various parts of our framework.

\subsubsection{Ablation study}
\begin{table*}[width=2.0\linewidth,cols=8,pos=t]
\caption{Ablation experimental results of the proposed framework on the two datasets. Here, IntraIPE means intra-image patch exchange method, InterIPE means inter-image patch exchange method, SDIC is the simulation of different imaging conditions, and SSL indicates self-supervised learning}\label{ablation_study} 
\begin{tabular*}{\tblwidth}{@{} LLLLLLLL @{} }
\toprule
\multicolumn{4}{c}{Step}	&	\multicolumn{2}{c}{SYSU}	& \multicolumn{2}{c}{SECOND}\\
\midrule
IntraIPE & InterIPE & SDIC & SSL  & OA  & F1  & OA  & F1 \\
\midrule
$\checkmark$ & & & & 0.7007& 0.4731&	0.5944&	0.3925 \\
 & $\checkmark$ & & & 0.7024&	0.4884&	0.7200&	0.4053 \\
$\checkmark$ & $\checkmark$& & & 0.7096& 0.4962&	0.7037&0.4179  \\
$\checkmark$ &$\checkmark$  &$\checkmark$  & & 0.7277	&0.5024&	0.7136& 0.4213	 \\
$\checkmark$ &$\checkmark$ & $\checkmark$& $\checkmark$& 0.7305	&0.5547&	0.7283&	0.4380 \\

\bottomrule
\end{tabular*}
\end{table*}
\par Our framework contains these four key parts, i.e., intra-image patch exchange method, inter-image patch exchange method, simulation of different imaging conditions, and self-supervised learning based on pseudo-labels. To verify the effectiveness of each part, we carry out the ablation study on the two benchmark datasets and report the contribution of each part to the final detection performance in Table \ref{ablation_study}. 

\begin{figure}[!t]
  \centering
  \subfloat[]{
    \includegraphics[width=1.55in]{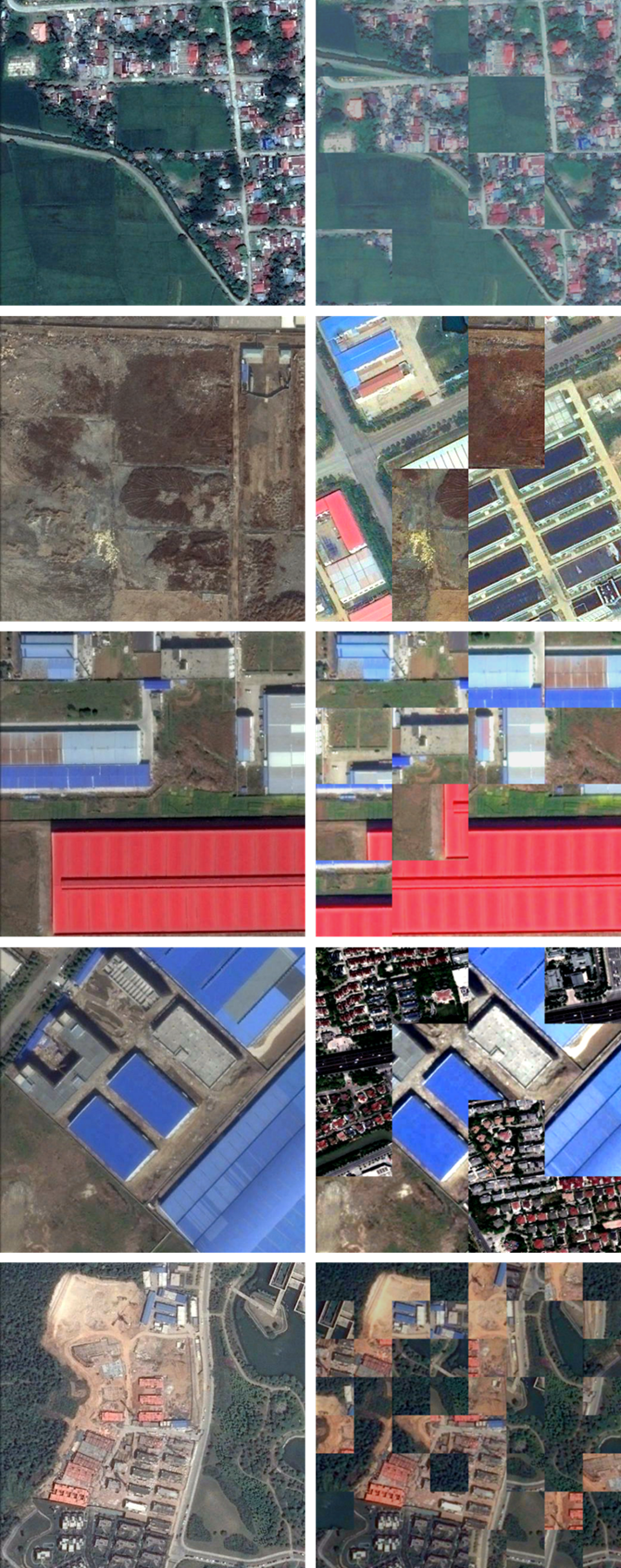}
  \label{fig_second_case}}
  \hfil
  \subfloat[]{
    \includegraphics[width=1.55in]{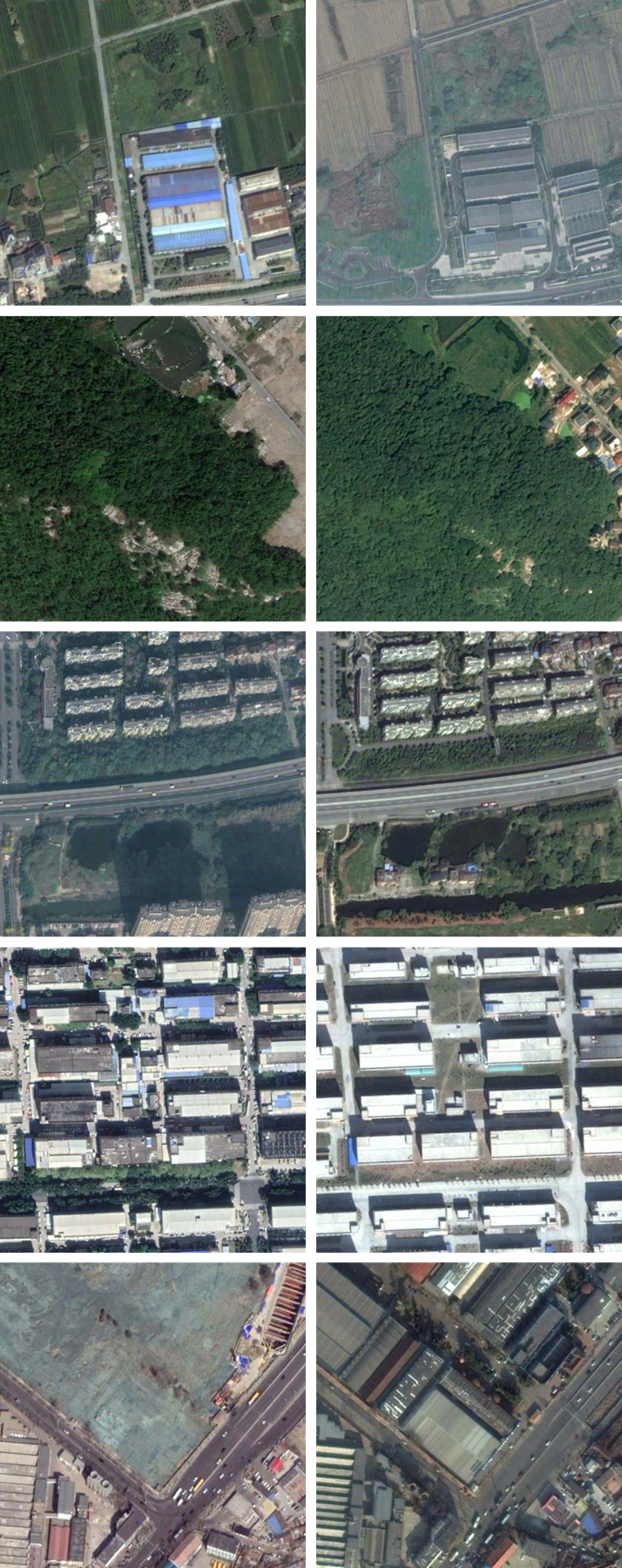}
  \label{fig_first_case}}
  \caption{Comparison of the (a) pseudo-bi-temporal image-pairs adjusted by our different imaging conditions simulation method and (b) real bi-temporal image pairs in the SECOND dataset. }
  \label{compare_pesudo_real}
\end{figure}

\par Firstly, only utilizing the intra-image patch exchange method to generate bi-temporal image pairs to train the change detector can obtain 0.4731 and 0.3925 F1 scores on the SYSU and SECOND datasets, respectively. These values are better than that of unsupervised SOTA models such as DCVA and KPCA-MNet. Compared to the intra-image patch exchange method, the inter-image patch exchange method can generate a wider variety of change events in the generated pseudo-bi-temporal samples. As a result, the change detector trained using samples generated by the inter-image patch exchange method can achieve better accuracy, with F1 values of 0.4884 and 0.4053 on the two datasets, respectively. By combining the two methods, the performance of the change detector can be further improved, with F1 scores of 0.4962 and 0.4179 on the SYSU and SECOND datasets, respectively. These results suggest that we can indeed train an effective change detector on unpaired and unlabelled remote sensing images by the simple idea of exchanging image patches to produce different kinds of land-cover changes.

\par Then, adjusting the pseudo-bi-temporal images in color balance, brightness, contrast, and sharpness to simulate different imaging conditions can improve the F1 score of the change detector to 0.5024 and 0.4213 on the two datasets, respectively. Figure \ref{compare_pesudo_real} compares some pseudo-bi-temporal images processed by our simulation method for different imaging conditions to real bi-temporal images on the SECOND dataset. It can be observed that there is a radiometric difference between pre-event images and post-event images in real bi-temporal image pairs due to differences in solar altitude angle, sensor attitude, and atmospheric conditions at the time of imaging. This difference is effectively modeled by our method. Visually, the adjusted pseudo-bi-temporal image does close to the actual real image in appearance.

\par Finally, self-supervised learning based on pseudo-labels can further improve the performance of the change detector. The final F1 values of our framework on the two datasets are 0.5547 and 0.4380, respectively. Note that the whole process of self-supervised learning is automatic, and no additional human supervision information is required. Therefore, the whole framework still remains unsupervised. In addition, we can see that the improvement of self-supervised learning on the SYSU dataset is much more significant than that on the SECOND dataset. This is because the SYSU dataset has relatively simple scenes and relatively few change events compared to the SECOND dataset. The pseudo-labels obtained by the change detector can be used as more accurate and effective supervision information.

\subsubsection{Hyperparameter analysis}\label{sec:4.3.2}
\par After the ablation study, we further analyze some hyperparameters in our framework that have a significant impact on the final change detection performance, including the size of exchanged image patches $\sigma$, the ratio of the number of exchanged image patches to the total number of image patches $r$, and the threshold value $\tau$ to filter low-confident pseudo labels in self-supervised learning.

\begin{figure*}[!t]
  \centering
  \subfloat[]{
    \includegraphics[width=3.3in]{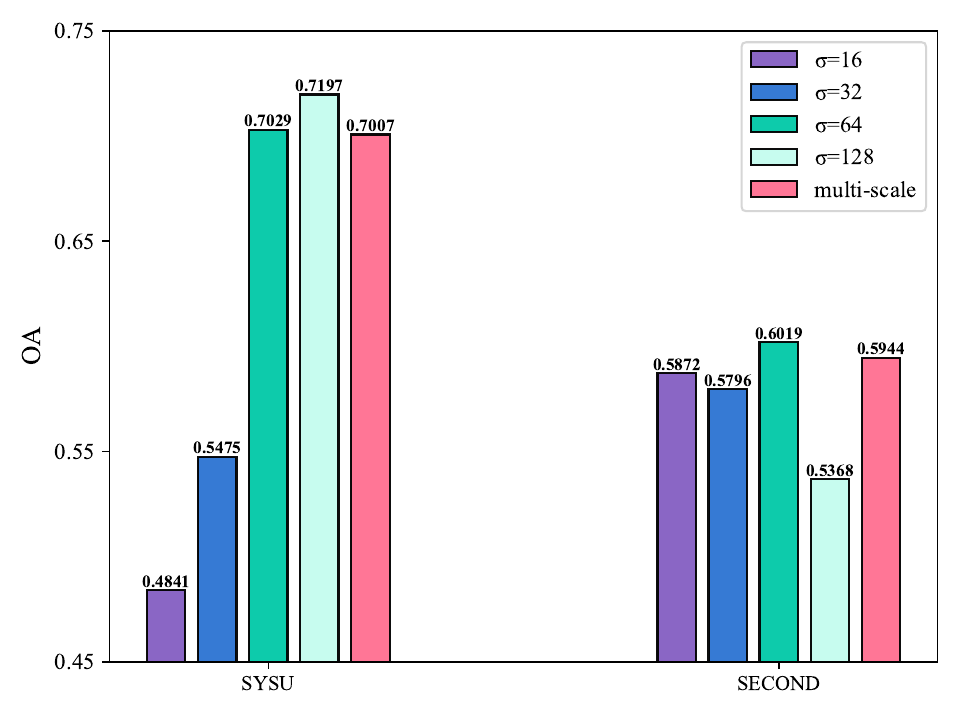}
  \label{fig_second_case}}
  \hfil
  \subfloat[]{
    \includegraphics[width=3.3in]{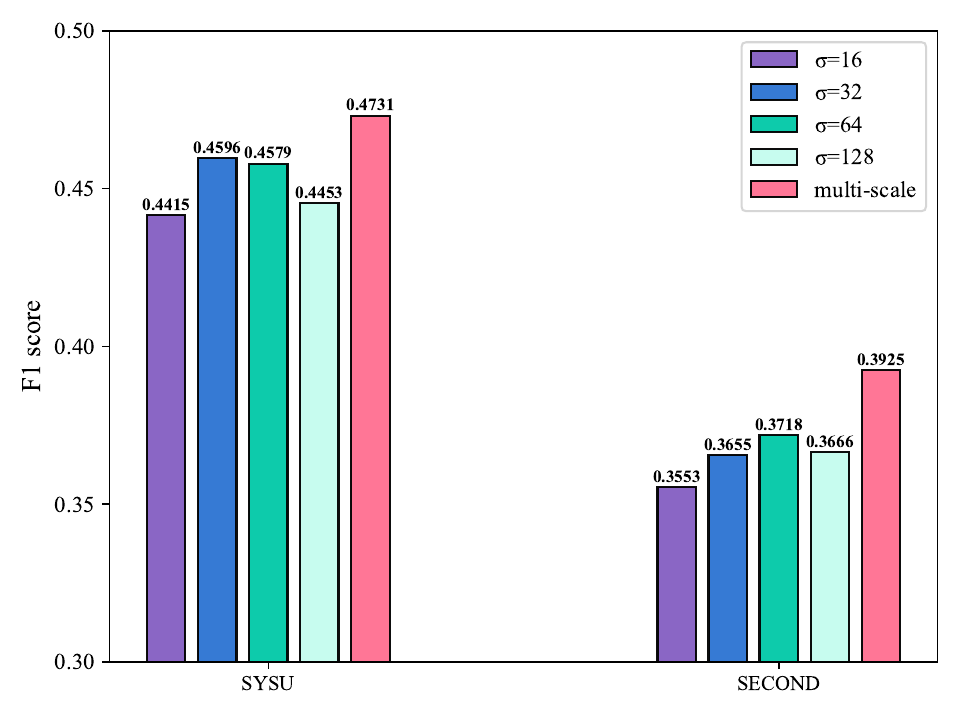}
  \label{fig_first_case}}
  \caption{The accuracy of the change detector trained on the samples generated by intra-image patch exchange under different values of scale parameter $\sigma$ on the two datasets. (a) Overall accuracy. (b) F1 score.}
  \label{discuss_sigma}
\end{figure*}

\par \textbf{1) The scale factor $\sigma$} is a very important hyperparameter in our framework, which controls the scale of land-cover changes in the generated samples. Figure \ref{discuss_sigma} shows the accuracy of the change detector trained on the samples generated by our intra-image patch exchange method under different values of $\sigma$. Considering that the image sizes in the two datasets are 256 and 512, respectively, the sampling value of $\sigma$ is set to 16, 32, 64, and 128, respectively, for ease of integer division. On the SYSU dataset, when $\sigma=16$, the resulting land-cover change is too fine-grained. Thus, only an OA of 0.4841 and an F1 value of 0.4415 are available. As  $\sigma$ increased, OA and F1 also increased. The optimal OA and F1 are obtained at $\sigma = 32$ and $\sigma = 64$, respectively. On the SECOND dataset, the trend of OA and F1 values with $\sigma$ values is slightly different from the SYSU dataset due to the difference in covered scenarios and change events. Optimal OA and F1 values are obtained at $\sigma = 64$. By using the proposed multi-scale sampling strategy, better F1 values than single-scale can be achieved, with 1.34$\%$ and 2.07$\%$ improvement in F1 score on the two datasets, respectively. As the SECOND dataset is richer in terms of land-cover change categories and scales, the performance of the trained change detectors is more significantly improved by the multi-scale sampling strategy on this dataset.

\begin{figure*}[!t]
  \centering
  \subfloat[]{
    \includegraphics[width=3.3in]{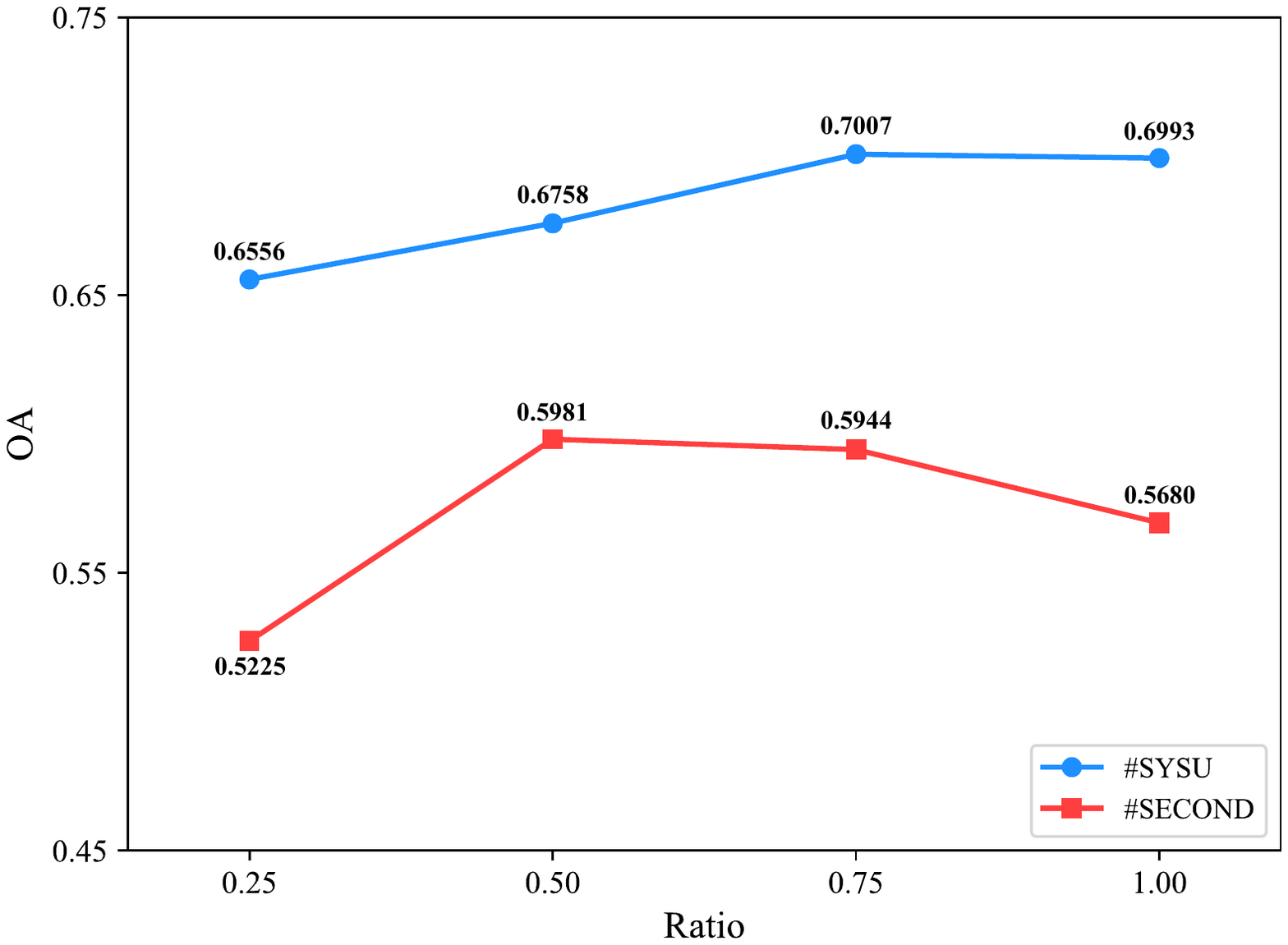}
  \label{fig_second_case}}
  \hfil
  \subfloat[]{
    \includegraphics[width=3.3in]{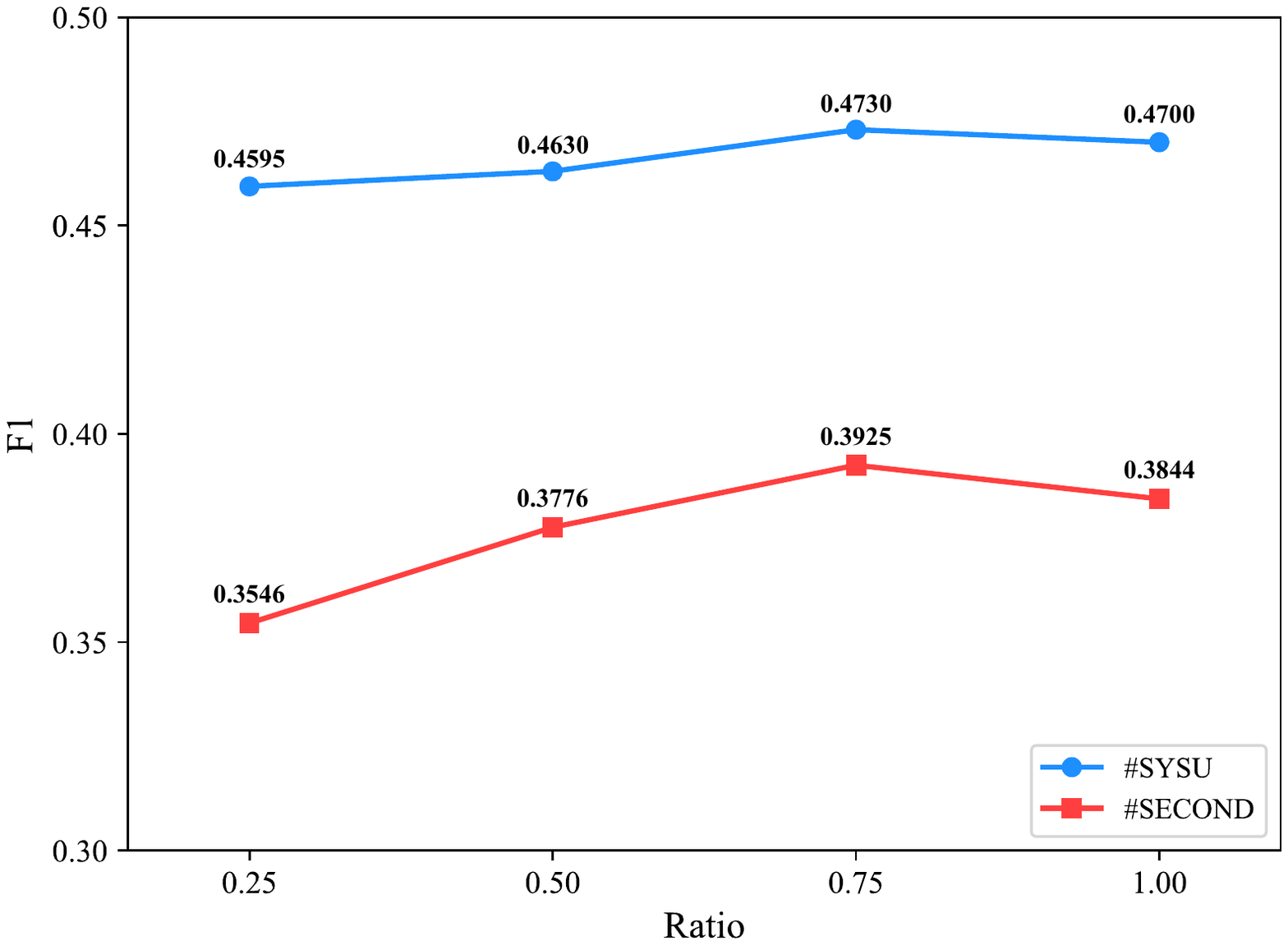}
  \label{fig_first_case}}
  \caption{The accuracy of the change detector trained on the samples generated by the intra-image patch exchange method under different values of exchange ratio $r_{intra}$ on the two datasets. (a) Overall accuracy. (b) F1 score.}
  \label{discuss_r_intra}
\end{figure*}

\begin{figure*}[!t]
  \centering
  \subfloat[]{
    \includegraphics[width=3.3in]{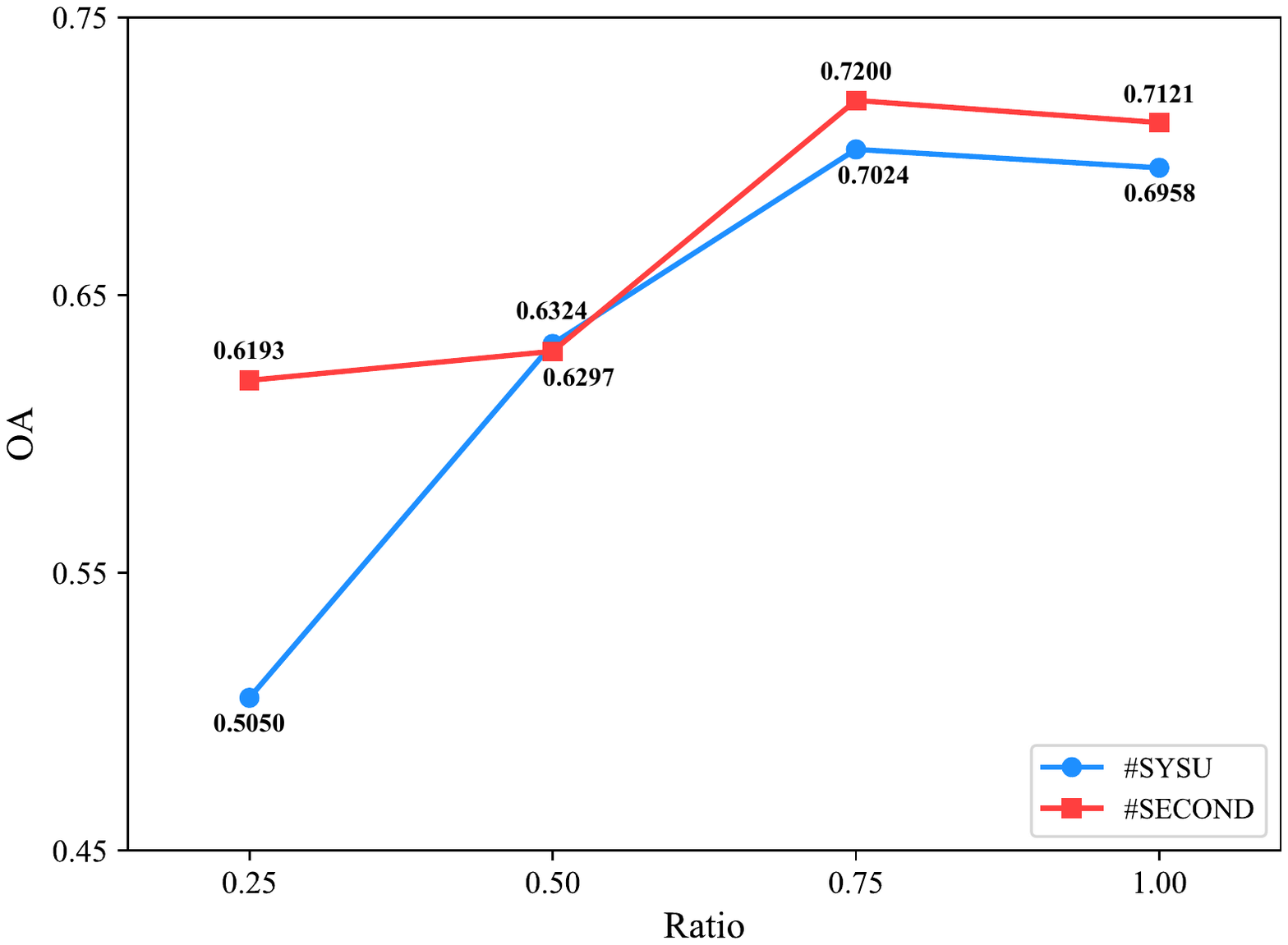}
  \label{fig_second_case}}
  \hfil
  \subfloat[]{
    \includegraphics[width=3.3in]{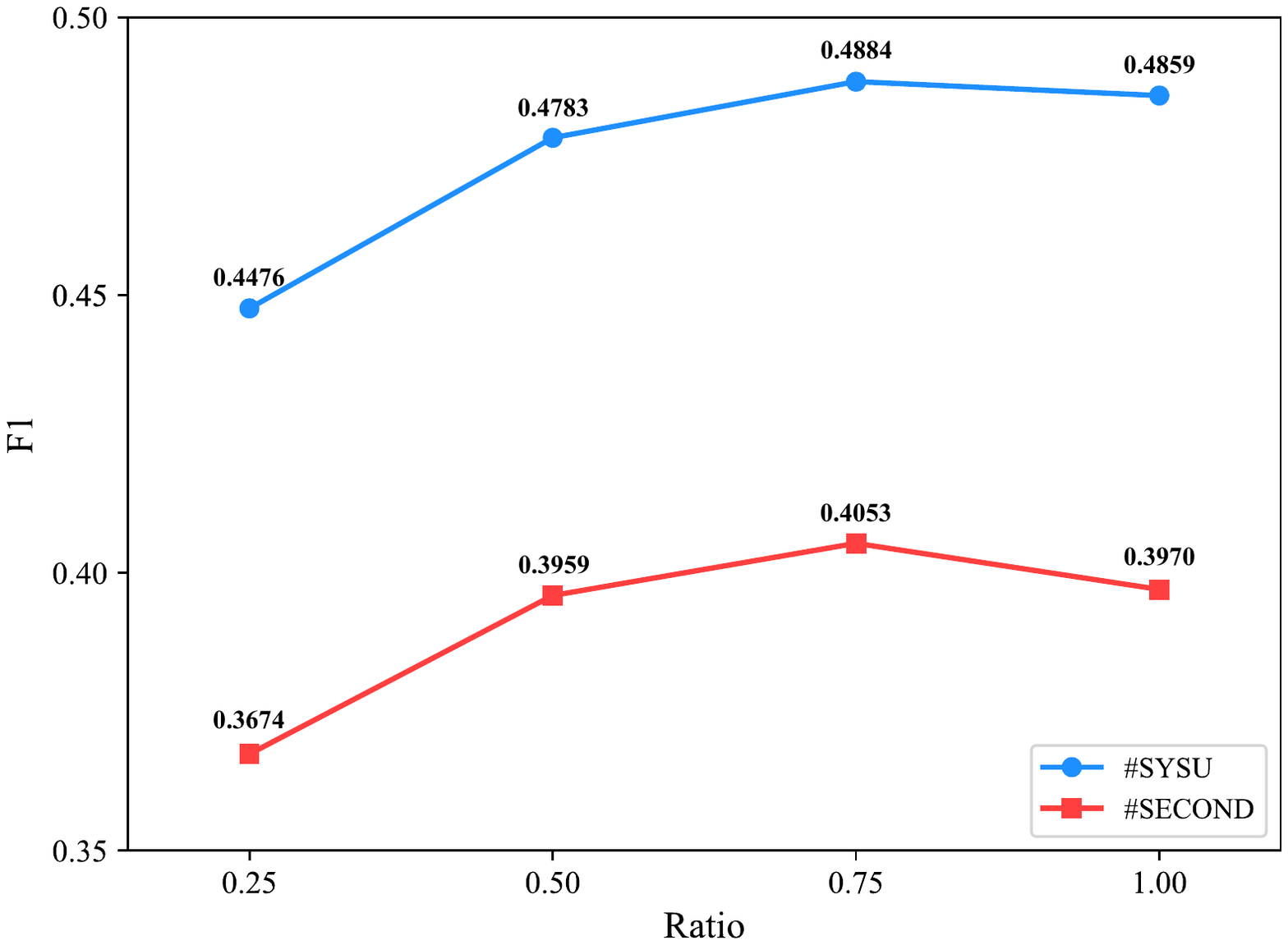}
  \label{fig_first_case}}
  \caption{The accuracy of the change detector trained on the samples generated by the inter-image patch exchange method under different values of exchange ratio $r_{inter}$ on the two datasets. (a) Overall accuracy. (b) F1 score.}
  \label{discuss_r_inter}
\end{figure*}

\par \textbf{2) The exchange ratio} in intra-image patch exchange $r_{intra}$ and inter-image patch exchange $r_{inter}$ are two other important hyperparameters that influence the detection performance. The larger $r_{intra}$ and $r_{inter}$, the greater the number of changed pixels and the richer the type of land-cover changes produced. Figure \ref{discuss_r_intra} and Figure \ref{discuss_r_inter} show the relationship between change detection performance and $r_{intra}$ and $r_{inter}$, respectively. As $r_{intra}$ and $r_{inter}$ increase, the performance of the change detectors trained using samples obtained from both intra- and inter-image patch exchange methods increases. The highest F1 values are achieved on both datasets when $r_{intra}=r_{inter}=0.75$. When $r_{intra}$ and $r_{inter}$ are further increased to 1, i.e., all image patches are exchanged, no labels can provide accurate information about unchanged pixels. Hence, the performance of the trained change detector instead undergoes a decrease. In addition, when $r_{inter}=1$, our inter-image patch exchange method directly compares the two clustering maps to generate change labels. Thus, the ChangeStar framework proposed in \citep{zheng2021change} can be treated as a special case of the inter-image patch exchange method in our I3PE framework in unsupervised scenarios.

\begin{table}[width=1.0\linewidth,cols=3,pos=t]
\caption{Comparison in F1 score of self-supervised learning with different threshold $\tau$ to generate pseudo labels}\label{com_stcd} 
\begin{tabular*}{\tblwidth}{@{} LLL @{} }
\toprule
$\tau$	&	SYSU		& SECOND \\
\midrule
0.7 & 0.5258 &  0.4274 \\
0.8 & 0.5316 &  0.4293 \\
0.9 & 0.5439  & 0.4357 \\
0.95 &  \textbf{0.5547} & \textbf{0.4380}  \\
0.99 & 0.5519 & 0.4324\\ 
\bottomrule
\end{tabular*}
\end{table}

\par \textbf{3) The threshold value $\tau$} is an important hyperparameter for self-supervised learning. If $\tau$ is too small, the generated pseudo-labels contain too many noisy labels, thereby damaging the detection performance. However, if $\tau$ is too large, the available land-cover change information for self-supervised learning would be too less. In Table \ref{com_stcd}, we report the F1 score achieved by the change detector under different threshold values. It can be seen that the F1 values obtained by our method increase as $\tau$ increases, with the best performance of the change detector when the $\tau$ value is 0.95; as $\tau$ increases further to 0.99, a decrease in the F1 values obtained by the detector occurs on both datasets. Therefore, setting $\tau=0.95$ is optimal for the two datasets.

\subsubsection{Performance of different change detectors}\label{sec:4.3.3}
\begin{table}[width=1.0\linewidth,cols=3,pos=t]
\caption{Comparison in F1 score obtained by the proposed change detector with different encoders.}\label{diff_backbone} 
\begin{tabular*}{\tblwidth}{@{} LLL @{} }
\toprule
Encoder	&	SYSU		& SECOND \\
\midrule
ResNet-18 & 0.5507 &  0.4316 \\
ResNet-34 & 0.5533  & 0.4342\\
ResNet-50 &  0.5547 & 0.4380  \\
ResNet-101 & 0.5458 & 0.4417 \\
\midrule
SENet-50 & 0.5493  &  0.4402 \\
EfficientNet-B3 &  0.5553 &  0.4431 \\
MixFormer-B2 & 0.5396 &  0.4472 \\
\bottomrule
\end{tabular*}
\end{table}
\par In the above experiments, our framework presents a deep change detector based on the FCN architecture to detect land-cover changes. Actually, the proposed I3PE is a general framework. Thus, we can employ other advanced deep network architectures as change detectors for better detection performance. Here, to verify this point briefly, we replace our change detector's encoder with other ResNet variants and three off-the-shelf representative networks, i.e., SENet \citep{Hu2018Squeeze}, EfficientNet \citep{Tan2019EfficientNet}, and MixFormer \citep{Xie2021SegFormer}. SENet is a deep CNN architecture with a channel attention mechanism. EfficientNet is a lightweight CNN architecture. MixFormer is a Transformer architecture. We report their F1 scores on the two datasets in Table \ref{diff_backbone}.

\par On the SYSU dataset, ResNet-50 has better F1 values than ResNet-18 and ResNet-34 due to the deeper network architecture of ResNet-50, which allows for more representative semantic information to be extracted. However, the performance of ResNet-101 is inferior to that of ResNet-50 and even ResNet-18. It can also be observed that the performance of SENet-50 and MixFormer-B2 is also inferior to that of ResNet-18. This may be due to the fact that it is easier to fit noisy labels as the network's feature extraction capability increases. The best F1 value of 0.5553 is achieved by using the lightweight network EfficientNet-B3 as the encoder for the change detector.

\par On the SECOND dataset, we can see that more sophisticated and advanced detectors give better detection performance due to the greater difficulty of change detection. As can be seen, the F1 value of ResNet series increases as the depth of the network increases. The SENet-50 has a boost in F1 values by introducing a channel attention mechanism into the ResNet-50 architecture. In comparison to the CNN architecture, the Transformer architecture, MixFormer-B2, achieves the highest accuracy on the SECOND dataset, with an F1 value of 0.4472. 

\subsubsection{Comparison with PCC method}
\begin{table}[width=1.0\linewidth,cols=3,pos=t]
\caption{Comparison in F1 score with the PCC approach based on DBSCAN and OBIA.}\label{com_pcc} 
\begin{tabular*}{\tblwidth}{@{} LLL @{} }
\toprule
Method	&	SYSU		& SECOND \\
\midrule
PCC & 0.3631& 0.3496 \\
I3PE & 0.5547 & 0.4380\\
\bottomrule
\end{tabular*}
\end{table}

\par To further validate the effectiveness of our motivation to exchange patches of unpaired remote sensing images to enable deep networks to learn information on land-cover changes, we compare here with the post-classification comparison (PCC) method. The post-classification comparison method is a prevalent and typical paradigm for change detection. Its core idea is to classify the multi-temporal images and then compare the classification results to obtain land-cover change results. Here, similar to the scheme presented in our inter-image patch exchange framework, we execute the SLIC and DBSCAN algorithms on the stacked bi-temporal images to get the classification results with unified categories and then compare them to get the land-cover change maps. 

\par Table \ref{com_pcc} presents the F1 values obtained by the proposed I3PE framework and the PCC approach on both datasets. We can see that the F1 values of I3PE are significantly better than those of PCC. This is because the PCC method suffers from the problem of cumulative classification errors \citep{Singh1989}, and its detection accuracy is heavily dependent on classification accuracy. On the other hand, unsupervised clustering methods often have difficulty obtaining very accurate classification results. In contrast, although our method uses OBIA and adaptive clustering, it does allow the deep network to learn the distribution of land-cover changes by exchanging intra- and inter-image patches, which results in better detection results. 

\subsubsection{Computational efficiency}\label{sec:ce}
\begin{table*}[width=1.87\linewidth,cols=11,pos=t]
\caption{Computational time (in hour) of the eight comparison models and the proposed I3PE on the two datasets.} \label{computational_efficiency}
    \begin{tabular*}{\tblwidth}{@{} l  | l l l l l l l  l | l l @{}}
\toprule
\multirow{2}{*}{Datasets} &	\multirow{2}{*}{CVA}	&\multirow{2}{*}{IRMAD}&	\multirow{2}{*}{ISFA}&	\multirow{2}{*}{OBCD}&	\multirow{2}{*}{DCAE}&	\multirow{2}{*}{DCVA}&	\multirow{2}{*}{DSFA}&	\multirow{2}{*}{KPCA-MNet}&	\multicolumn{2}{c}{I3PE} \\
\cline{10-11}
 & & & & & & &  & &Training  & Inference  \\
\midrule
SYSU & 0.124 &  0.570 & 0.319 & 2.082& 0.341&  5.444& 16.389& 6.673&  3.542 & 0.027 \\
SECOND &0.472	 &1.343   &0.472 & 5.926&0.479 &3.632 &10.805 &9.484 & 5.796 & 0.018\\
\bottomrule
\end{tabular*}
\end{table*}

\par The computational overhead of the eight comparison methods and our I3PE framework on the two datasets are listed in Table \ref{computational_efficiency}. Note that CVA, IRMAD, ISFA, OBCD, and KPCA-MNet run on the CPU, while DCAE, DCVA, and I3PE run on the CPU and GPU. 

\par The benchmark method CVA takes 0.124 and 0.472 hours on the two datasets, respectively. IRMAD and ISFA are more time-consuming than CVA due to the need to solve the associated optimization problem and the inclusion of an iterative process. The three methods, DCVA, DSFA, and KPCA-MNet, are all very time-consuming. This is because these unsupervised deep learning-based methods generally focus on relatively small study regions and are only tested on a few image pairs. In order to achieve better detection performance, they have several optimizations and operations on each pair of images. For example, DSFA needs to perform a pre-detection method and separate optimization of network parameters and SFA transformation matrix for each image pair; KPCA-MNet needs to solve the KPCA problem on each image pair.

\par In comparison, our method requires a bit long time in the training stage, although we can obtain the object maps of remote sensing images and clustering maps required by the intra-image patch exchange framework in advance. This is because the inter-image exchange method jointly clusters two images in real-time during the training stage. This part of the algorithm runs on the CPU and is therefore time-consuming. However, after the training stage is complete, the change detector can make inferences very quickly. The time taken to complete the inference on the two test sets is 0.027 hours and 0.018 hours, respectively. The average time required to detect land changes from a pair of bi-temporal images of size 512$\times$512 is only 0.04 seconds.

\subsubsection{Semi-supervised learning scenarios}
\begin{table*}[width=2.0\linewidth,cols=4,pos=t]
\caption{Performance comparison of the change detector trained with and without I3PE on the SYSU dataset in semisupervised learning case. Here, GT means ground truth (GT) annotations}\label{sscd_SYSU} 
\begin{tabular*}{\tblwidth}{@{} LLLL @{} }
\toprule
Supervision type	&	Method	& OA		& F1 score\\
\midrule
Unsupervised &	I3PE &	0.7305&	0.5547 \\
\midrule
\multirow{6}{*}{Semisupervised} & 1$\%$ GT + I3PE &	0.8057	 &0.5877 \\
& 5$\%$ GT + I3PE &	0.8187	&0.6357 \\
& 10$\%$ GT + I3PE&	0.8198	&0.6664 \\
\cline{2-4} 
& 1$\%$ GT&	0.7820	&0.5440\\
& 5$\%$ GT&	0.8198	&0.6095\\
& 10$\%$ GT&	0.8388	&0.6516\\
\midrule
Supervised&	Oracle&	0.8638&	0.7207 \\
\bottomrule
\end{tabular*}
\end{table*}

\begin{table*}[width=2.0\linewidth,cols=4,pos=t]
\caption{Performance comparison of change detectors trained with and without I3PE on the SECOND dataset in semisupervised learning case. Here, GT means ground truth (GT) annotations}\label{sscd_SECOND} 
\begin{tabular*}{\tblwidth}{@{} LLLL @{} }
\toprule
Supervision type	&	Method	& OA		& F1 score\\
\midrule
Unsupervised &	I3PE &	0.7283&	0.4380 \\
\midrule
\multirow{6}{*}{Semisupervised} & 5$\%$ GT + I3PE&	0.7426	 &0.4689 \\
& 10$\%$ GT + I3PE & 	0.8035	& 0.4842\\
& 20$\%$ GT + I3PE &	0.8086	& 0.5053\\
\cline{2-4}
& 5$\%$ GT&	0.7379	&0.4428\\
& 10$\%$ GT&	0.8150	&0.4710\\
& 20$\%$ GT&	0.8245	&0.4979\\
\midrule
Supervised&	Oracle&	0.8301	&	0.5389 \\
\bottomrule
\end{tabular*}
\end{table*}

\par A common scenario in real-world task and production environments is that we have a large number of unlabeled single-temporal images and a small number of multi-temporal images with annotation information. For this scenario, we present the corresponding semi-supervised learning framework in Section \ref{sec:3.4} that exploits the unpaired and unlabelled images through our image patch exchange approach to improve the performance of the change detector. Table \ref{sscd_SYSU} and \ref{sscd_SECOND} compare the accuracy obtained by change detectors trained on a small number of labeled bi-temporal images with and without the aid of our I3PE method. 

\par We can see that in the case of sparse annotation information, applying our method to provide additional change information can bring a relatively significant performance improvement for the change detector. On the SYSU dataset, with only 1$\%$ and 5$\%$ of the annotated samples in the training set used to train the detector, the utilization of I3PE as an additional training aid can result in a 4.37$\%$ and 2.63$\%$ improvement in the F1 score. On the SECOND dataset, with 5$\%$ of annotated samples, I3PE boosts the F1 score of the change detector by 2.61$\%$. As the number of labeled bi-temporal images increases, the change detector receives abundant land-cover change information. Thus, the performance improvement of our method for the change detector is not as pronounced. This result aligns with the intuition because the land-cover changes created by exchanging image patches are certainly less accurate, rich, and consistent with the actual land-cover change distribution than the real labeled samples in the training set. However, the apparent performance improvement of our method for detectors in the presence of sparsely annotated samples and its ability to be seamlessly embedded in the training process of deep networks make our approach well-suited to a practical production environment. 

\subsection{Application at a real study site}
\par The highlight of I3PE is that we lift the restriction on training change detectors that require pairwise bi-temporal images with annotated information. We can use a large number of unpaired and unlabelled images, which are easier to collect in practice, to train the change detector. In addition to the experiments on two large-scale datasets that provide benchmark results, we have further carried out experiments here to detect land-cover changes of an actual study area using the I3PE framework. Specifically, we blended 10$\%$ of the SYSU training set and 20$\%$ of the SECOND dataset with the Wuhan dataset as unpaired and unlabelled images for training the change detector. The specific change detector still uses the architecture proposed in section \ref{sec:3.2}, with ResNet-50 as the encoder. We also adopt the benchmark unsupervised model CVA, image transformation method ISFA, and SOTA deep learning-based method DCVA as comparison methods.  

\begin{figure*}[!t]
  \centering
  \includegraphics[width=6.3in]{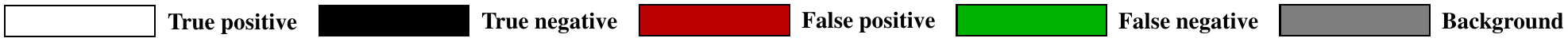}
  \\
  \subfloat[]{
    \includegraphics[width=3.1in]{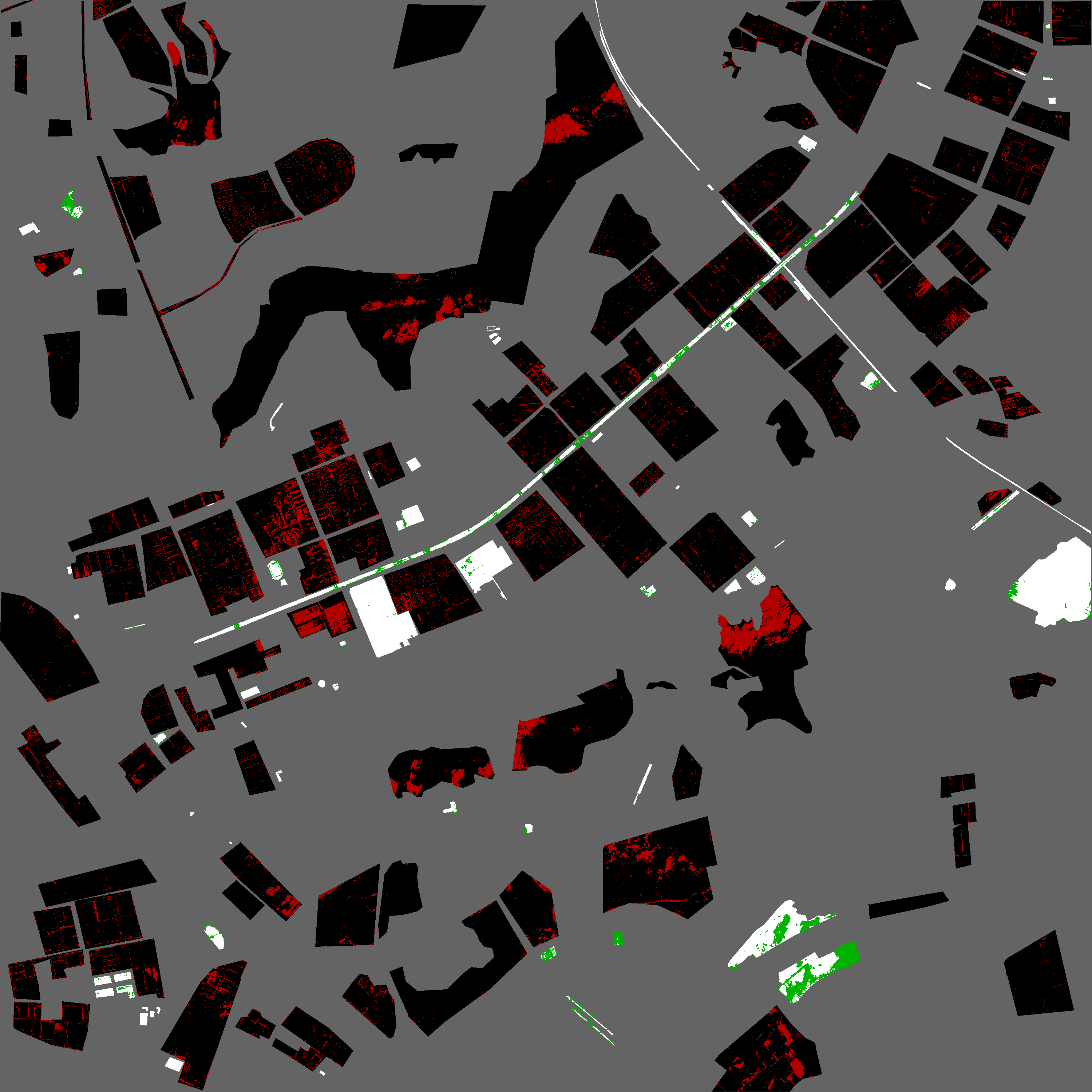}
  \label{fig_second_case}}
  \hfil
  \subfloat[]{
    \includegraphics[width=3.1in]{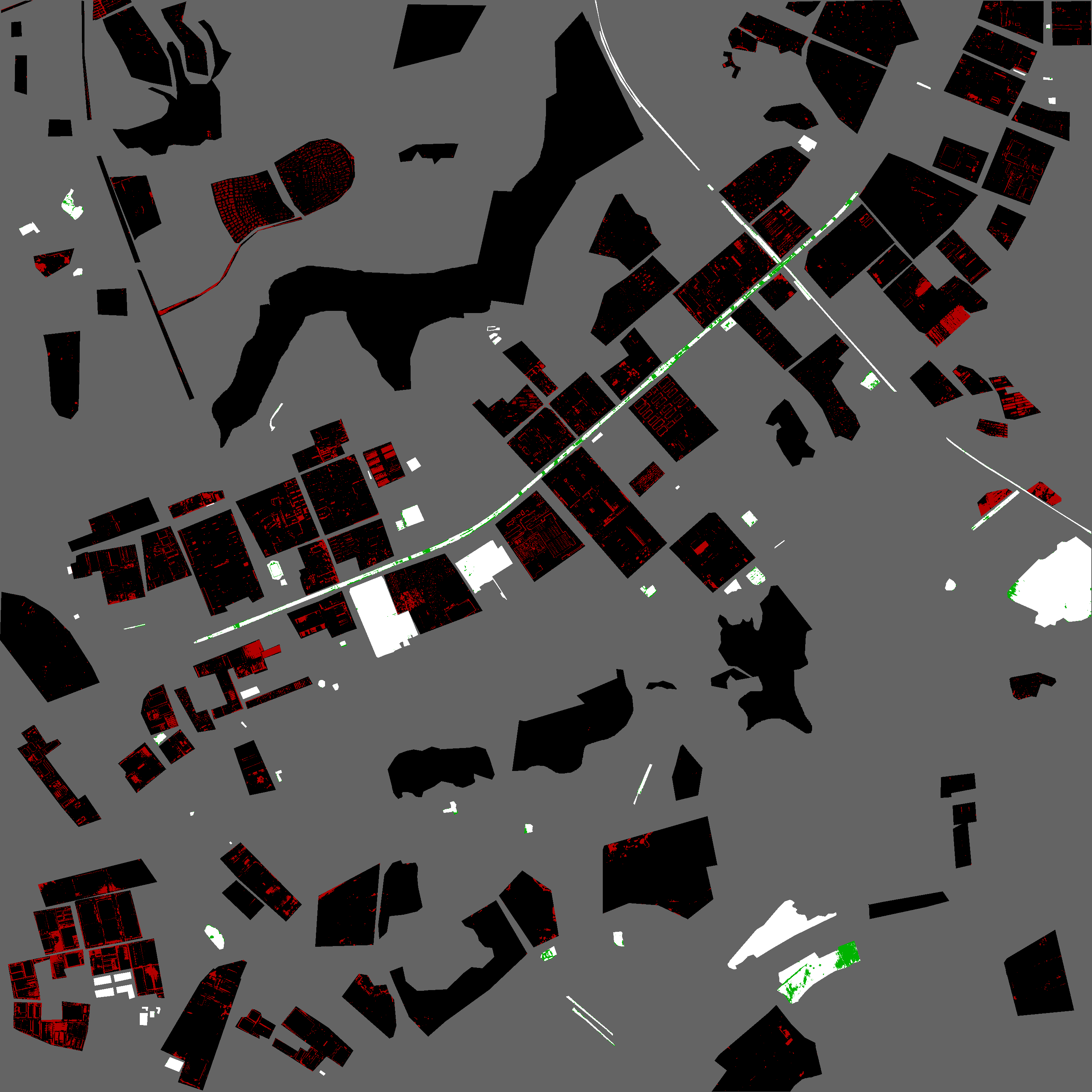}
  \label{fig_second_case}}
  
  \subfloat[]{
    \includegraphics[width=3.1in]{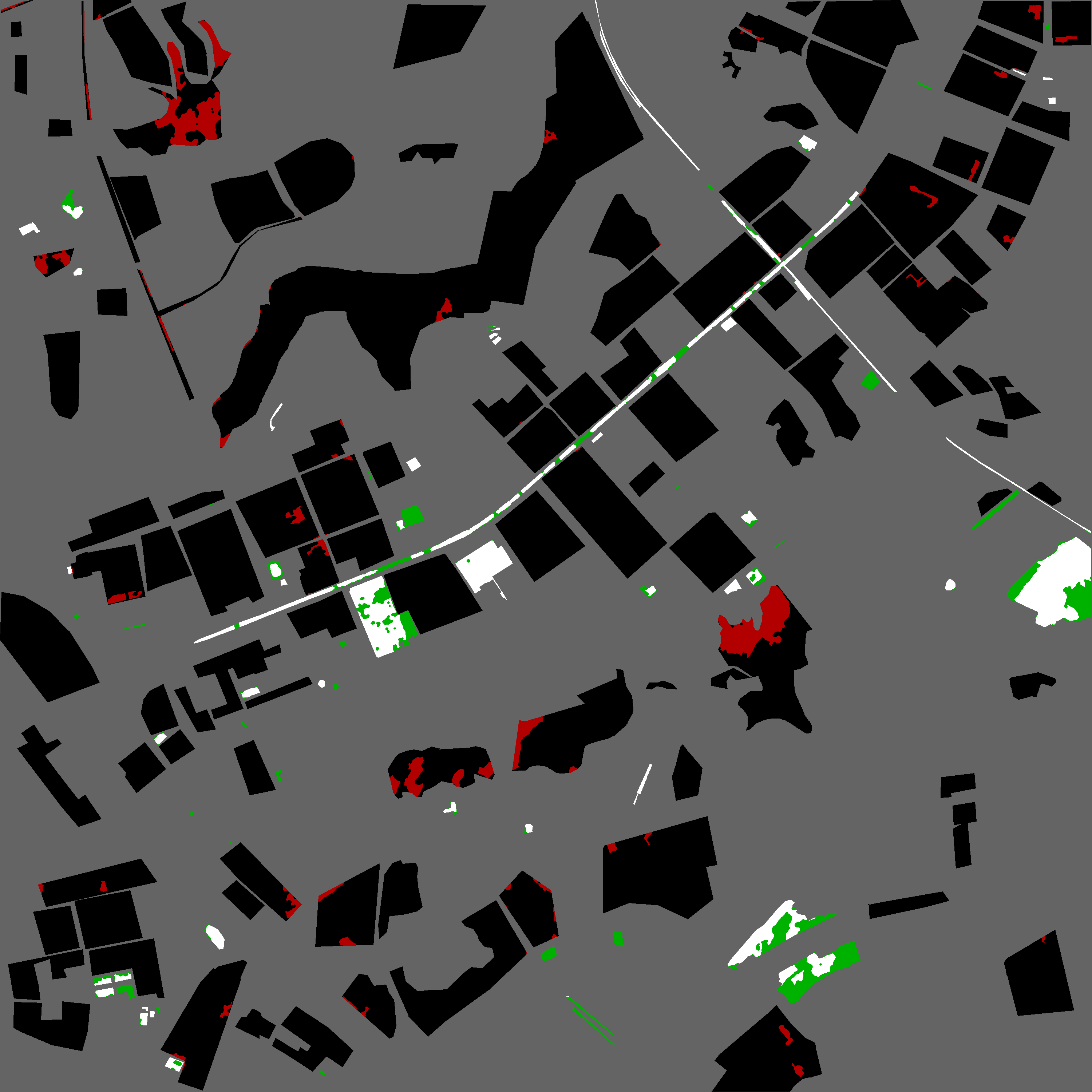}
  \label{fig_first_case}}
  \hfil
  \subfloat[]{
    \includegraphics[width=3.1in]{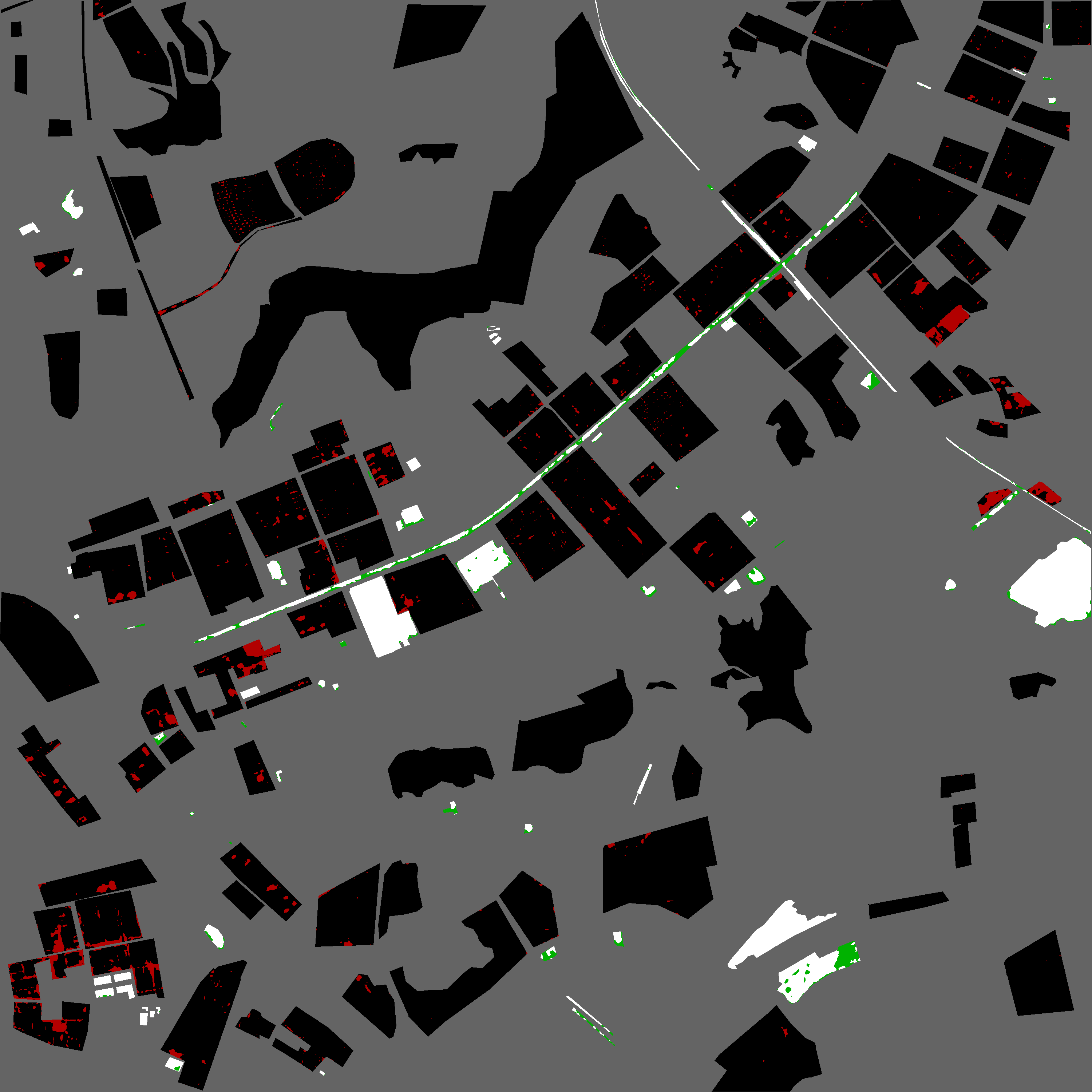}
  \label{fig_first_case}}
  \caption{Change maps obtained by (a) CVA, (b) ISFA, (c) DCVA, and (d) I3PE on the Wuhan dataset. In the obtained change maps, white represents TP; black represents TN; red represents FP; green represents FN; gray is background. Zoom in for a better visual effect.}
  \label{WH_CD}
\end{figure*}

\begin{table*}[width=2.0\linewidth,cols=6,pos=t]
\caption{Accuracy assessment for different unsupervised change detection approaches on the Wuhan dataset. The table highlights the highest values in bold.}\label{acc_WH} 
\begin{tabular*}{\tblwidth}{@{}  LLLLLL @{} }
\toprule
Method	&Recall  & Precision &	OA		& F1 score & Inference time (s)\\
\midrule
CVA&	0.8412 & 0.4681 & 0.9178 & 0.6015 & 21.9\\
ISFA&  \textbf{0.9105} & 0.5274 & 0.9333 & 0.6679 & 42.7\\
DCVA&	0.6773 & 0.6269 & 0.9465 &	0.6511 & 320.0\\
I3PE&	0.8547 & \textbf{0.7161}&\textbf{0.9643}& \textbf{0.7793}& 6.7 \\
\bottomrule
\end{tabular*}
\end{table*}

\par The specific land-cover change maps obtained by our framework and comparison models are shown in Figure \ref{WH_CD}. Table \ref{acc_WH} reports the specific quantitative results. In the study area covered by this Wuhan dataset, I3PE achieves the highest accuracy compared to traditional and deep learning-based methods. As can be seen from the obtained change maps, CVA and ISFA can detect most of the changed areas in the study area, but there are many unchanged pixels that are falsely detected, i.e., more red FP pixels. The main types of these FP pixels are pixel shifts caused by alignment errors, shadows, and differences in radiation from one region to another caused by larger study areas. DCVA can reduce these FP pixels to some extent, but there are many changed pixels that are ignored, i.e., more green FN pixels. In contrast, our I3PE framework is able to effectively use multi-source unpaired and unlabelled images from which land-cover changes are learned and thus enable us to analyze land-cover changes in the study area accurately. In addition, since our method directly uses an FCN to infer the change map on the GPU, it is more efficient than the methods CVA and ISFA, which run on the CPU, and the DCVA method, which requires many additional operations to be taken.

\section{Limitations and future study}\label{sec:5}
\par The experiments in the previous section amply demonstrate the effectiveness of our I3PE framework, which can train change detectors from unpaired and unlabelled remote sensing images with significantly better accuracy than the existing unsupervised SOTA models. It can also be used as a means of data augmentation to improve the performance of the change detectors in the case of sparsely labeled data. However, there are still some shortcomings in the existing framework, which we discuss in this section to inspire subsequent research. 

\begin{figure*}[!t]
\centering
  \includegraphics[width=6in]{figs/SYSU_Legend.pdf}
  \centering
  \subfloat[]{
    \includegraphics[width=3.1in]{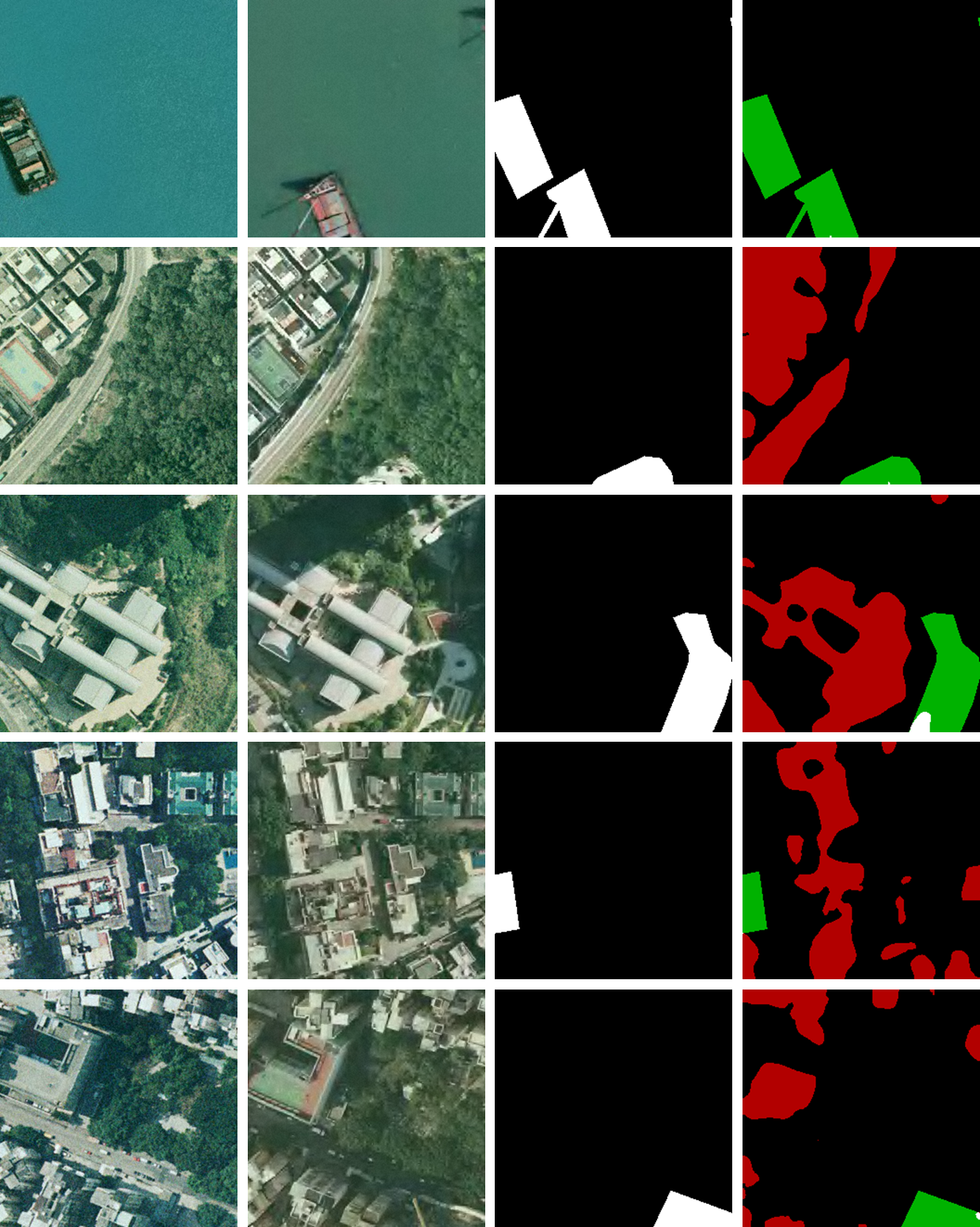}
  \label{fig_second_case}}
  \hfil
  \subfloat[]{
    \includegraphics[width=3.1in]{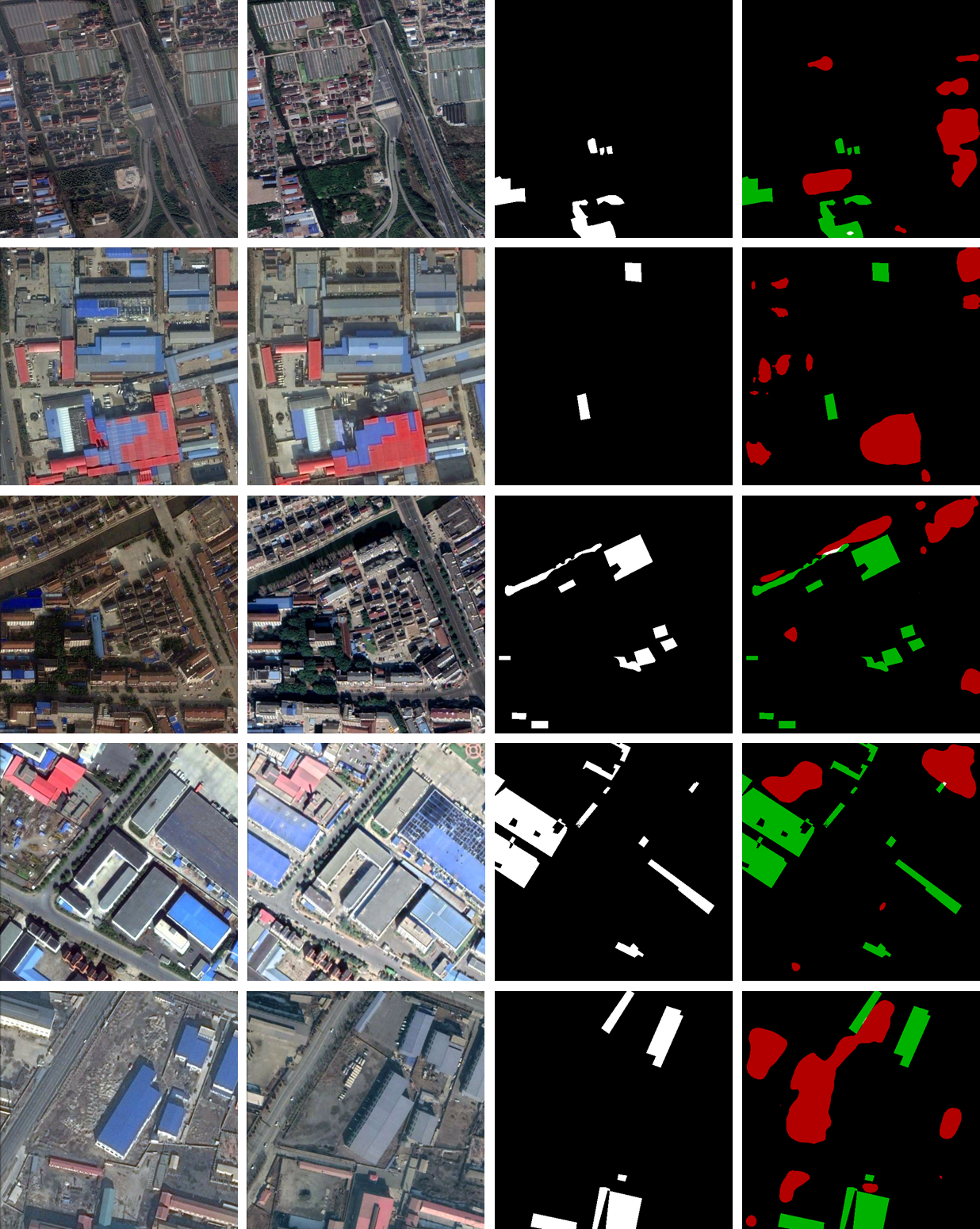}
  \label{fig_first_case}}
  \caption{Some bitemporal images in (a) SYSU dataset and (b) SECOND dataset for which our framework fails to detect land-cover changes. In change maps, white represents TP; black represents TN; red represents FP; green represents FN.}
  \label{failed_cd}
\end{figure*}

\par Firstly, the ultimate accuracy of our framework depends heavily on the accuracy of the clustering algorithm and the number of types of land-cover changes generated through intra- and inter-image patch exchange. Therefore, if the accuracy of the clustering algorithm is too low or sufficient labels are not generated for certain types of changes through exchanging image patches, the trained change detectors may not be able to detect the corresponding changes accurately. Figure \ref{failed_cd} shows some bi-temporal image pairs in the two test sets where our framework fails to detect land-cover changes. Therefore, we will consider adding other image features, such as texture information and spatial statistical properties, to improve the performance of the clustering algorithms. Another point about the clustering algorithm is that we only empirically set hyperparameters for the entire large dataset. However, it is clear that the hyperparameters should be set differently for an image with a simple scene and a complex scene containing many kinds of land-cover objects. Therefore, we will consider adaptively adjusting the hyperparameters of the clustering algorithm according to the complexity of the image scene and the richness of the features within the image.

\par Then, spatial discontinuity is inevitably introduced due to the proposed image patch exchange schemes. Change labels have square patterns as the exchange process is performed randomly on the square patch level. This means truncation and incompleteness can occur for many large-scale and continuous land-cover features, even though we design a multi-scale sampling strategy. The samples generated in this way do not adequately reflect their actual distribution. This may result in the change detector not being able to thoroughly learn their distribution patterns from our generated samples, thus limiting the performance of the detectors to some extent. Inspired by this issue, we would like to explore more elegant ways to generate samples closer to real land-cover change patterns in future studies, such as performing an exchange process on the object/instance level. Moreover, improving our framework to deal with pseudo-changes caused by seasons (e.g., vegetation) and change detection in tilted viewpoints is worth studying.

\par The whole I3PE framework can also be seen as a special weakly supervised learning process; that is, the change detector needs to learn the true distribution of land-cover change from the noisy labels generated by our patch exchange methods. In this paper, as our major motivation is primarily whether we can develop a simple but effective method to make deep networks learn land-cover changes leveraging unpaired and unlabelled images, we are directly allowing the deep network to learn from noisy labels without employing theories and techniques related to weakly supervised learning to improve the performance of the network. In the future, we can investigate how to make the network able to learn robustly from these generated noisy labels by studying and developing related theories and techniques \citep{Han2018Co} in the change detection scenarios.

\par Regarding our approach to simulate different imaging conditions, while it enables the generated pseudo-bi-temporal images to appear close to the actual radiation difference, the hyperparameters of these image enhancement methods have only been adjusted empirically and the whole pipeline does not consider the exact distribution of the data in the dataset. In the future, more accurate statistical models or even some generative methods such as generative adversarial networks \citep{Goodfellow2020Generative} could be considered to fit real data distribution, thereby better simulating the actual radiation differences.

\par Finally, as we mentioned in section \ref{sec:ce}, the inter-image patch exchange method in our framework requires clustering the stacked images in real-time while the change detector is being trained. This process is time-consuming, especially when the volume of data and the scale of remote sensing images are large. Therefore, we will consider how to accelerate the clustering algorithm, including multi-threading and implementing the corresponding adaptive clustering algorithm on the GPU. In addition, we currently set a fixed number of objects in the segmentation method for the whole dataset. Actually, for images only containing simple scenes (e.g., only water bodies/vegetation), we can reduce the number of objects obtained by the segmentation algorithm, thus further improving the efficiency of the clustering algorithm.

\section{Conclusion}\label{sec:6}
\par This paper proposes an unsupervised single-temporal change detection framework called I3PE that can train deep learning-based change detectors from more readily available unlabelled and unpaired single-temporal images. The I3PE framework is easily implemented based on the simple idea of generating land-cover changes by exchanging image patches within the image and between images. Specifically, we propose intra- and inter-image patch exchange methods based on the OBIA method and adaptive clustering algorithm, which can generate corresponding pseudo-bi-temporal image pairs and change labels from single-temporal images. In order to make the generated image pairs more realistic, we propose a simulation method to fit the different imaging conditions in real imaging situations. Finally, we introduce a self-supervised learning method based on pseudo-labels that can further improve the performance of change detectors in both unsupervised and semi-supervised settings. 

\par Experimental results on two large benchmark datasets, SYSU and SECOND, show that our framework can outperform some representative traditional and deep learning-based unsupervised approaches, with F1 value improvements of 10.65$\%$ and 6.99$\%$ to SOTA approaches. The ablation study and hyperparameters discussions have demonstrated the effectiveness of the various components of the I3PE framework. In addition, our I3PE method can be seamlessly embedded in the training process of deep change detectors to leverage unlabeled single-temporal images. Experiments in the semi-supervised setting show that I3PE can be used as an additional auxiliary training method to boost the F1 value of the change detector by 4.37$\%$ and 2.61$\%$ in the presence of sparse annotated data on the SYSU and SECOND datasets, respectively. Finally, we have further validated the usability and effectiveness of the I3PE method for the practical land-cover change analysis task on a specific study site. We believe that I3PE could become a simple and effective benchmark method for land-cover change detection and has the potential to be widely applied in real applications.

\section*{Acknowledgements}
\par This work was supported in part by
the JSPS, KAKENHI under Grant Number 22H03609, JST, FOREST under Grant Number JPMJFR206S, Microsoft Research Asia, and the Graduate School of Frontier Sciences, The University of Tokyo, through the Challenging New Area Doctoral Research Grant (Project No. C2303).

\printcredits

\bibliographystyle{cas-model2-names}

\bibliography{cas-refs}

\begin{thebibliography}{71}
\expandafter\ifx\csname natexlab\endcsname\relax\def\natexlab#1{#1}\fi
\providecommand{\url}[1]{\texttt{#1}}
\providecommand{\href}[2]{#2}
\providecommand{\path}[1]{#1}
\providecommand{\DOIprefix}{doi:}
\providecommand{\ArXivprefix}{arXiv:}
\providecommand{\URLprefix}{URL: }
\providecommand{\Pubmedprefix}{pmid:}
\providecommand{\doi}[1]{\href{http://dx.doi.org/#1}{\path{#1}}}
\providecommand{\Pubmed}[1]{\href{pmid:#1}{\path{#1}}}
\providecommand{\bibinfo}[2]{#2}
\ifx\xfnm\relax \def\xfnm[#1]{\unskip,\space#1}\fi
\bibitem[{Achanta et~al.(2012)Achanta, Shaji, Smith, Lucchi, Fua and
  Süsstrunk}]{Achanta2012SLIC}
\bibinfo{author}{Achanta, R.}, \bibinfo{author}{Shaji, A.},
  \bibinfo{author}{Smith, K.}, \bibinfo{author}{Lucchi, A.},
  \bibinfo{author}{Fua, P.}, \bibinfo{author}{Süsstrunk, S.},
  \bibinfo{year}{2012}.
\newblock \bibinfo{title}{Slic superpixels compared to state-of-the-art
  superpixel methods}.
\newblock \bibinfo{journal}{IEEE Trans. Pattern Anal. Mach. Intell.}
  \bibinfo{volume}{34}, \bibinfo{pages}{2274--2282}.
\bibitem[{Bandara and Patel(2022)}]{Bandara2022Transformer}
\bibinfo{author}{Bandara, W.G.C.}, \bibinfo{author}{Patel, V.M.},
  \bibinfo{year}{2022}.
\newblock \bibinfo{title}{A transformer-based siamese network for change
  detection}, in: \bibinfo{booktitle}{IGARSS 2022 - 2022 IEEE International
  Geoscience and Remote Sensing Symposium}, pp. \bibinfo{pages}{207--210}.
\bibitem[{Bergamasco et~al.(2022)Bergamasco, Saha, Bovolo and
  Bruzzone}]{Bergamasco2022Unsupervised}
\bibinfo{author}{Bergamasco, L.}, \bibinfo{author}{Saha, S.},
  \bibinfo{author}{Bovolo, F.}, \bibinfo{author}{Bruzzone, L.},
  \bibinfo{year}{2022}.
\newblock \bibinfo{title}{Unsupervised change detection using
  convolutional-autoencoder multiresolution features}.
\newblock \bibinfo{journal}{IEEE Trans. Geosci. Remote Sens.}
  \bibinfo{volume}{60}, \bibinfo{pages}{1--19}.
\bibitem[{Bovolo and Bruzzone(2007)}]{Bovolo2007a}
\bibinfo{author}{Bovolo, F.}, \bibinfo{author}{Bruzzone, L.},
  \bibinfo{year}{2007}.
\newblock \bibinfo{title}{{A Theoretical Framework for Unsupervised Change
  Detection Based on Change Vector Analysis in the Polar Domain}}.
\newblock \bibinfo{journal}{IEEE Trans. Geosci. Remote Sens.}
  \bibinfo{volume}{45}, \bibinfo{pages}{218--236}.
\bibitem[{Bovolo et~al.(2008)Bovolo, Bruzzone and Marconcini}]{Bovolo2008}
\bibinfo{author}{Bovolo, F.}, \bibinfo{author}{Bruzzone, L.},
  \bibinfo{author}{Marconcini, M.}, \bibinfo{year}{2008}.
\newblock \bibinfo{title}{{A novel approach to unsupervised change detection
  based on a semisupervised SVM and a similarity measure}}.
\newblock \bibinfo{journal}{IEEE Trans. Geosci. Remote Sens.}
  \bibinfo{volume}{46}, \bibinfo{pages}{2070--2082}.
\bibitem[{Bruzzone and {Diego Fern{\`{a}}ndez Prieto}(2000)}]{Sharma2007}
\bibinfo{author}{Bruzzone, L.}, \bibinfo{author}{{Diego Fern{\`{a}}ndez
  Prieto}}, \bibinfo{year}{2000}.
\newblock \bibinfo{title}{{Automatic Analysis of the Difference Image for
  Unsupervised Change Detection}}.
\newblock \bibinfo{journal}{IEEE Trans. Geosci. Remote Sens.}
  \bibinfo{volume}{38}, \bibinfo{pages}{1171--1182}.
\bibitem[{Canty and Nielsen(2008)}]{Canty2008}
\bibinfo{author}{Canty, M.J.}, \bibinfo{author}{Nielsen, A.A.},
  \bibinfo{year}{2008}.
\newblock \bibinfo{title}{{Automatic radiometric normalization of multitemporal
  satellite imagery with the iteratively re-weighted MAD transformation}}.
\newblock \bibinfo{journal}{Remote Sens. Environ.} \bibinfo{volume}{112},
  \bibinfo{pages}{1025--1036}.
\bibitem[{Cao and Huang(2023)}]{CAO2023full}
\bibinfo{author}{Cao, Y.}, \bibinfo{author}{Huang, X.}, \bibinfo{year}{2023}.
\newblock \bibinfo{title}{A full-level fused cross-task transfer learning
  method for building change detection using noise-robust pretrained networks
  on crowdsourced labels}.
\newblock \bibinfo{journal}{Remote Sens. Environ.} \bibinfo{volume}{284},
  \bibinfo{pages}{113371}.
\bibitem[{{Caye Daudt} et~al.(2018){Caye Daudt}, {Le Saux} and
  Boulch}]{CayeDaudt2018}
\bibinfo{author}{{Caye Daudt}, R.}, \bibinfo{author}{{Le Saux}, B.},
  \bibinfo{author}{Boulch, A.}, \bibinfo{year}{2018}.
\newblock \bibinfo{title}{{Fully convolutional siamese networks for change
  detection}}, in: \bibinfo{booktitle}{Proceedings - International Conference
  on Image Processing, ICIP}, pp. \bibinfo{pages}{4063--4067}.
\bibitem[{Celik(2009)}]{Celik2009}
\bibinfo{author}{Celik, T.}, \bibinfo{year}{2009}.
\newblock \bibinfo{title}{{Unsupervised change detection in satellite images
  using principal component analysis and K-means clustering}}.
\newblock \bibinfo{journal}{IEEE Geosci. Remote Sens. Lett.}
  \bibinfo{volume}{6}, \bibinfo{pages}{772--776}.
\bibitem[{Chen et~al.(2022a)Chen, Nemni, Vallecorsa, Li, Wu and
  Bromley}]{chen2022dual}
\bibinfo{author}{Chen, H.}, \bibinfo{author}{Nemni, E.},
  \bibinfo{author}{Vallecorsa, S.}, \bibinfo{author}{Li, X.},
  \bibinfo{author}{Wu, C.}, \bibinfo{author}{Bromley, L.},
  \bibinfo{year}{2022}a.
\newblock \bibinfo{title}{Dual-tasks siamese transformer framework for building
  damage assessment}, in: \bibinfo{booktitle}{International Geoscience and
  Remote Sensing Symposium (IGARSS)}, pp. \bibinfo{pages}{1600--1603}.
\bibitem[{Chen et~al.(2022b)Chen, Qi and Shi}]{Chen2022Remote}
\bibinfo{author}{Chen, H.}, \bibinfo{author}{Qi, Z.}, \bibinfo{author}{Shi,
  Z.}, \bibinfo{year}{2022}b.
\newblock \bibinfo{title}{Remote sensing image change detection with
  transformers}.
\newblock \bibinfo{journal}{IEEE Trans. Geosci. Remote Sens.}
  \bibinfo{volume}{60}, \bibinfo{pages}{1--14}.
\bibitem[{Chen et~al.(2020)Chen, Wu, Du, Zhang and Wang}]{Chen2019a}
\bibinfo{author}{Chen, H.}, \bibinfo{author}{Wu, C.}, \bibinfo{author}{Du, B.},
  \bibinfo{author}{Zhang, L.}, \bibinfo{author}{Wang, L.},
  \bibinfo{year}{2020}.
\newblock \bibinfo{title}{{Change Detection in Multisource VHR Images via Deep
  Siamese Convolutional Multiple-Layers Recurrent Neural Network}}.
\newblock \bibinfo{journal}{IEEE Trans. Geosci. Remote Sens.}
  \bibinfo{volume}{58}, \bibinfo{pages}{2848--2864}.
\bibitem[{Chen et~al.(2023)Chen, Yokoya and Chini}]{chen2023fourier}
\bibinfo{author}{Chen, H.}, \bibinfo{author}{Yokoya, N.},
  \bibinfo{author}{Chini, M.}, \bibinfo{year}{2023}.
\newblock \bibinfo{title}{Fourier domain structural relationship analysis for
  unsupervised multimodal change detection}.
\newblock \bibinfo{journal}{ISPRS Journal of Photogrammetry and Remote Sensing}
  \bibinfo{volume}{198}, \bibinfo{pages}{99--114}.
\bibitem[{Chen et~al.(2022c)Chen, Yokoya, Wu and Du}]{chen2022unsupervised}
\bibinfo{author}{Chen, H.}, \bibinfo{author}{Yokoya, N.}, \bibinfo{author}{Wu,
  C.}, \bibinfo{author}{Du, B.}, \bibinfo{year}{2022}c.
\newblock \bibinfo{title}{{Unsupervised Multimodal Change Detection Based on
  Structural Relationship Graph Representation Learning}}.
\newblock \bibinfo{journal}{IEEE Trans. Geosci. Remote Sens.} ,
  \bibinfo{pages}{1--18}.
\bibitem[{Chen et~al.(2022d)Chen, Zhang, Hong, Chen, Yang and Li}]{CHEN2022101}
\bibinfo{author}{Chen, P.}, \bibinfo{author}{Zhang, B.}, \bibinfo{author}{Hong,
  D.}, \bibinfo{author}{Chen, Z.}, \bibinfo{author}{Yang, X.},
  \bibinfo{author}{Li, B.}, \bibinfo{year}{2022}d.
\newblock \bibinfo{title}{Fccdn: Feature constraint network for vhr image
  change detection}.
\newblock \bibinfo{journal}{ISPRS J. Photogramm. Remote Sens.}
  \bibinfo{volume}{187}, \bibinfo{pages}{101--119}.
\bibitem[{Coppin et~al.(2004)Coppin, Jonckheere, Nackaerts, Muys and
  Lambin}]{Coppin2004}
\bibinfo{author}{Coppin, P.}, \bibinfo{author}{Jonckheere, I.},
  \bibinfo{author}{Nackaerts, K.}, \bibinfo{author}{Muys, B.},
  \bibinfo{author}{Lambin, E.}, \bibinfo{year}{2004}.
\newblock \bibinfo{title}{{Digital change detection methods in ecosystem
  monitoring: A review}}.
\newblock \bibinfo{journal}{Int. J. Remote Sens.} \bibinfo{volume}{25},
  \bibinfo{pages}{1565--1596}.
\bibitem[{Deng et~al.(2008)Deng, Wang, Deng, Qi, Wang, Deng and Pca}]{Deng2008}
\bibinfo{author}{Deng, J.S.}, \bibinfo{author}{Wang, K.},
  \bibinfo{author}{Deng, Y.H.}, \bibinfo{author}{Qi, G.J.},
  \bibinfo{author}{Wang, K.}, \bibinfo{author}{Deng, Y.H.},
  \bibinfo{author}{Pca, G.J.Q.}, \bibinfo{year}{2008}.
\newblock \bibinfo{title}{{PCA-based land-use change detection and analysis
  using multitemporal and multisensor satellite data}}.
\newblock \bibinfo{journal}{Int. J. Remote Sens.} \bibinfo{volume}{1161}.
\bibitem[{Dosovitskiy et~al.(2020)Dosovitskiy, Beyer, Kolesnikov, Weissenborn,
  Zhai, Unterthiner, Dehghani, Minderer, Heigold, Gelly
  et~al.}]{dosovitskiy2020image}
\bibinfo{author}{Dosovitskiy, A.}, \bibinfo{author}{Beyer, L.},
  \bibinfo{author}{Kolesnikov, A.}, \bibinfo{author}{Weissenborn, D.},
  \bibinfo{author}{Zhai, X.}, \bibinfo{author}{Unterthiner, T.},
  \bibinfo{author}{Dehghani, M.}, \bibinfo{author}{Minderer, M.},
  \bibinfo{author}{Heigold, G.}, \bibinfo{author}{Gelly, S.}, et~al.,
  \bibinfo{year}{2020}.
\newblock \bibinfo{title}{An image is worth 16x16 words: Transformers for image
  recognition at scale}.
\newblock \bibinfo{journal}{arXiv preprint arXiv:2010.11929} .
\bibitem[{Du et~al.(2019)Du, Ru, Wu and Zhang}]{Du2019a}
\bibinfo{author}{Du, B.}, \bibinfo{author}{Ru, L.}, \bibinfo{author}{Wu, C.},
  \bibinfo{author}{Zhang, L.}, \bibinfo{year}{2019}.
\newblock \bibinfo{title}{{Unsupervised Deep Slow Feature Analysis for Change
  Detection in Multi-Temporal Remote Sensing Images}}.
\newblock \bibinfo{journal}{IEEE Trans. Geosci. Remote Sens.}
  \bibinfo{volume}{57}, \bibinfo{pages}{9976--9992}.
\bibitem[{Du et~al.(2020)Du, Wang, Chen, Liu, Lin and Meng}]{DU2020278}
\bibinfo{author}{Du, P.}, \bibinfo{author}{Wang, X.}, \bibinfo{author}{Chen,
  D.}, \bibinfo{author}{Liu, S.}, \bibinfo{author}{Lin, C.},
  \bibinfo{author}{Meng, Y.}, \bibinfo{year}{2020}.
\newblock \bibinfo{title}{An improved change detection approach using
  tri-temporal logic-verified change vector analysis}.
\newblock \bibinfo{journal}{ISPRS J. Photogramm. Remote Sens.}
  \bibinfo{volume}{161}, \bibinfo{pages}{278--293}.
\bibitem[{Ester et~al.(1996)Ester, Kriegel, Sander, Xu
  et~al.}]{ester1996density}
\bibinfo{author}{Ester, M.}, \bibinfo{author}{Kriegel, H.P.},
  \bibinfo{author}{Sander, J.}, \bibinfo{author}{Xu, X.}, et~al.,
  \bibinfo{year}{1996}.
\newblock \bibinfo{title}{A density-based algorithm for discovering clusters in
  large spatial databases with noise.}, in: \bibinfo{booktitle}{kdd}, pp.
  \bibinfo{pages}{226--231}.
\bibitem[{Gil-Yepes et~al.(2016)Gil-Yepes, Ruiz, Recio, Balaguer-Beser and
  Hermosilla}]{Gil-Yepes2016}
\bibinfo{author}{Gil-Yepes, J.L.}, \bibinfo{author}{Ruiz, L.A.},
  \bibinfo{author}{Recio, J.A.}, \bibinfo{author}{Balaguer-Beser, {\'{A}}.},
  \bibinfo{author}{Hermosilla, T.}, \bibinfo{year}{2016}.
\newblock \bibinfo{title}{{Description and validation of a new set of
  object-based temporal geostatistical features for land-use/land-cover change
  detection}}.
\newblock \bibinfo{journal}{ISPRS J. Photogramm. Remote Sens.}
  \bibinfo{volume}{121}, \bibinfo{pages}{77--91}.
\bibitem[{Gong et~al.(2017a)Gong, Niu, Zhang and Li}]{Gong2017}
\bibinfo{author}{Gong, M.}, \bibinfo{author}{Niu, X.}, \bibinfo{author}{Zhang,
  P.}, \bibinfo{author}{Li, Z.}, \bibinfo{year}{2017}a.
\newblock \bibinfo{title}{{Generative Adversarial Networks for Change Detection
  in Multispectral Imagery}}.
\newblock \bibinfo{journal}{IEEE Geosci. Remote Sens. Lett.}
  \bibinfo{volume}{14}, \bibinfo{pages}{2310--2314}.
\bibitem[{Gong et~al.(2017b)Gong, Zhan, Zhang and Miao}]{Gong2017Superpixel}
\bibinfo{author}{Gong, M.}, \bibinfo{author}{Zhan, T.}, \bibinfo{author}{Zhang,
  P.}, \bibinfo{author}{Miao, Q.}, \bibinfo{year}{2017}b.
\newblock \bibinfo{title}{Superpixel-based difference representation learning
  for change detection in multispectral remote sensing images}.
\newblock \bibinfo{journal}{IEEE Trans. Geosci. Remote Sens.}
  \bibinfo{volume}{55}, \bibinfo{pages}{2658--2673}.
\bibitem[{Goodfellow et~al.(2020)Goodfellow, Pouget-Abadie, Mirza, Xu,
  Warde-Farley, Ozair, Courville and Bengio}]{Goodfellow2020Generative}
\bibinfo{author}{Goodfellow, I.}, \bibinfo{author}{Pouget-Abadie, J.},
  \bibinfo{author}{Mirza, M.}, \bibinfo{author}{Xu, B.},
  \bibinfo{author}{Warde-Farley, D.}, \bibinfo{author}{Ozair, S.},
  \bibinfo{author}{Courville, A.}, \bibinfo{author}{Bengio, Y.},
  \bibinfo{year}{2020}.
\newblock \bibinfo{title}{Generative adversarial networks}.
\newblock \bibinfo{journal}{Commun. ACM} \bibinfo{volume}{63},
  \bibinfo{pages}{139–144}.
\bibitem[{Guo et~al.(2021)Guo, Shi, Marinoni, Du and Zhang}]{guo2021deep}
\bibinfo{author}{Guo, H.}, \bibinfo{author}{Shi, Q.},
  \bibinfo{author}{Marinoni, A.}, \bibinfo{author}{Du, B.},
  \bibinfo{author}{Zhang, L.}, \bibinfo{year}{2021}.
\newblock \bibinfo{title}{Deep building footprint update network: A
  semi-supervised method for updating existing building footprint from
  bi-temporal remote sensing images}.
\newblock \bibinfo{journal}{Remote Sens. Environ.} \bibinfo{volume}{264},
  \bibinfo{pages}{112589}.
\bibitem[{Han et~al.(2018)Han, Yao, Yu, Niu, Xu, Hu, Tsang and
  Sugiyama}]{Han2018Co}
\bibinfo{author}{Han, B.}, \bibinfo{author}{Yao, Q.}, \bibinfo{author}{Yu, X.},
  \bibinfo{author}{Niu, G.}, \bibinfo{author}{Xu, M.}, \bibinfo{author}{Hu,
  W.}, \bibinfo{author}{Tsang, I.}, \bibinfo{author}{Sugiyama, M.},
  \bibinfo{year}{2018}.
\newblock \bibinfo{title}{Co-teaching: Robust training of deep neural networks
  with extremely noisy labels}, in: \bibinfo{booktitle}{Advances in Neural
  Information Processing Systems}.
\bibitem[{He et~al.(2016)He, Zhang, Ren and Sun}]{He2016}
\bibinfo{author}{He, K.}, \bibinfo{author}{Zhang, X.}, \bibinfo{author}{Ren,
  S.}, \bibinfo{author}{Sun, J.}, \bibinfo{year}{2016}.
\newblock \bibinfo{title}{{Deep residual learning for image recognition}}.
\newblock \bibinfo{journal}{Proceedings of the IEEE Computer Society Conference
  on Computer Vision and Pattern Recognition} , \bibinfo{pages}{770--778}.
\bibitem[{{Hoberg} et~al.(2015){Hoberg}, {Rottensteiner}, {Feitosa} and
  {Heipke}}]{Hoberg2015}
\bibinfo{author}{{Hoberg}, T.}, \bibinfo{author}{{Rottensteiner}, F.},
  \bibinfo{author}{{Feitosa}, R.Q.}, \bibinfo{author}{{Heipke}, C.},
  \bibinfo{year}{2015}.
\newblock \bibinfo{title}{Conditional random fields for multitemporal and
  multiscale classification of optical satellite imagery}.
\newblock \bibinfo{journal}{IEEE Trans. Geosci. Remote Sens.}
  \bibinfo{volume}{53}, \bibinfo{pages}{659--673}.
\bibitem[{Hou et~al.(2021)Hou, Bai, Li, Shang and Shen}]{HOU2021103}
\bibinfo{author}{Hou, X.}, \bibinfo{author}{Bai, Y.}, \bibinfo{author}{Li, Y.},
  \bibinfo{author}{Shang, C.}, \bibinfo{author}{Shen, Q.},
  \bibinfo{year}{2021}.
\newblock \bibinfo{title}{High-resolution triplet network with dynamic
  multiscale feature for change detection on satellite images}.
\newblock \bibinfo{journal}{ISPRS J. Photogramm. Remote Sens.}
  \bibinfo{volume}{177}, \bibinfo{pages}{103--115}.
\bibitem[{Hu et~al.(2018)Hu, Shen and Sun}]{Hu2018Squeeze}
\bibinfo{author}{Hu, J.}, \bibinfo{author}{Shen, L.}, \bibinfo{author}{Sun,
  G.}, \bibinfo{year}{2018}.
\newblock \bibinfo{title}{Squeeze-and-excitation networks}, in:
  \bibinfo{booktitle}{2018 IEEE/CVF Conference on Computer Vision and Pattern
  Recognition}, pp. \bibinfo{pages}{7132--7141}.
\bibitem[{Hussain et~al.(2013)Hussain, Chen, Cheng, Wei and
  Stanley}]{Hussain2013}
\bibinfo{author}{Hussain, M.}, \bibinfo{author}{Chen, D.},
  \bibinfo{author}{Cheng, A.}, \bibinfo{author}{Wei, H.},
  \bibinfo{author}{Stanley, D.}, \bibinfo{year}{2013}.
\newblock \bibinfo{title}{{Change detection from remotely sensed images: From
  pixel-based to object-based approaches}}.
\newblock \bibinfo{journal}{ISPRS J. Photogramm. Remote Sens.}
  \bibinfo{volume}{80}, \bibinfo{pages}{91--106}.
\bibitem[{Kasetkasem and Varshney(2002)}]{Kasetkasem2002}
\bibinfo{author}{Kasetkasem, T.}, \bibinfo{author}{Varshney, P.},
  \bibinfo{year}{2002}.
\newblock \bibinfo{title}{An image change detection algorithm based on markov
  random field models}.
\newblock \bibinfo{journal}{IEEE Trans. Geosci. Remote Sens.}
  \bibinfo{volume}{40}, \bibinfo{pages}{1815--1823}.
\bibitem[{Kipf and Welling(2016)}]{kipf2016semi}
\bibinfo{author}{Kipf, T.N.}, \bibinfo{author}{Welling, M.},
  \bibinfo{year}{2016}.
\newblock \bibinfo{title}{Semi-supervised classification with graph
  convolutional networks}.
\newblock \bibinfo{journal}{arXiv preprint arXiv:1609.02907} .
\bibitem[{Liu et~al.(2020)Liu, Gong, Qin and Tan}]{Liu2020Bipartite}
\bibinfo{author}{Liu, J.}, \bibinfo{author}{Gong, M.}, \bibinfo{author}{Qin,
  A.K.}, \bibinfo{author}{Tan, K.C.}, \bibinfo{year}{2020}.
\newblock \bibinfo{title}{Bipartite differential neural network for
  unsupervised image change detection}.
\newblock \bibinfo{journal}{IEEE Trans. Neural Netw. Learn. Syst.}
  \bibinfo{volume}{31}, \bibinfo{pages}{876--890}.
\bibitem[{Liu et~al.(2018)Liu, Gong, Qin and Zhang}]{Liu2018}
\bibinfo{author}{Liu, J.}, \bibinfo{author}{Gong, M.}, \bibinfo{author}{Qin,
  K.}, \bibinfo{author}{Zhang, P.}, \bibinfo{year}{2018}.
\newblock \bibinfo{title}{{A Deep Convolutional Coupling Network for Change
  Detection Based on Heterogeneous Optical and Radar Images}}.
\newblock \bibinfo{journal}{IEEE Trans. Neural Netw. Learn. Syst.}
  \bibinfo{volume}{29}, \bibinfo{pages}{545--559}.
\bibitem[{Liu et~al.(2022)Liu, Zhang, Liu and Xiao}]{Liu2022}
\bibinfo{author}{Liu, J.}, \bibinfo{author}{Zhang, W.}, \bibinfo{author}{Liu,
  F.}, \bibinfo{author}{Xiao, L.}, \bibinfo{year}{2022}.
\newblock \bibinfo{title}{A probabilistic model based on bipartite
  convolutional neural network for unsupervised change detection}.
\newblock \bibinfo{journal}{IEEE Trans. Geosci. Remote Sens.}
  \bibinfo{volume}{60}, \bibinfo{pages}{1--14}.
\bibitem[{Liu et~al.(2021)Liu, Yang and Lunga}]{Liu2021}
\bibinfo{author}{Liu, T.}, \bibinfo{author}{Yang, L.}, \bibinfo{author}{Lunga,
  D.}, \bibinfo{year}{2021}.
\newblock \bibinfo{title}{{Change detection using deep learning approach with
  object-based image analysis}}.
\newblock \bibinfo{journal}{Remote Sens. Environ.} \bibinfo{volume}{256},
  \bibinfo{pages}{112308}.
\bibitem[{Long et~al.(2015)Long, Shelhamer and Darrell}]{long2015fully}
\bibinfo{author}{Long, J.}, \bibinfo{author}{Shelhamer, E.},
  \bibinfo{author}{Darrell, T.}, \bibinfo{year}{2015}.
\newblock \bibinfo{title}{Fully convolutional networks for semantic
  segmentation}, in: \bibinfo{booktitle}{Proceedings of the IEEE conference on
  computer vision and pattern recognition}, pp. \bibinfo{pages}{3431--3440}.
\bibitem[{Loshchilov and Hutter(2017)}]{loshchilov2017decoupled}
\bibinfo{author}{Loshchilov, I.}, \bibinfo{author}{Hutter, F.},
  \bibinfo{year}{2017}.
\newblock \bibinfo{title}{Decoupled weight decay regularization}.
\newblock \bibinfo{journal}{arXiv preprint arXiv:1711.05101} .
\bibitem[{Luppino et~al.(2022)Luppino, Kampffmeyer, Bianchi, Moser, Serpico,
  Jenssen and Anfinsen}]{Luppino2022Deep}
\bibinfo{author}{Luppino, L.T.}, \bibinfo{author}{Kampffmeyer, M.},
  \bibinfo{author}{Bianchi, F.M.}, \bibinfo{author}{Moser, G.},
  \bibinfo{author}{Serpico, S.B.}, \bibinfo{author}{Jenssen, R.},
  \bibinfo{author}{Anfinsen, S.N.}, \bibinfo{year}{2022}.
\newblock \bibinfo{title}{{Deep Image Translation with an Affinity-Based Change
  Prior for Unsupervised Multimodal Change Detection}}.
\newblock \bibinfo{journal}{IEEE Trans. Geosci. Remote Sens.}
  \bibinfo{volume}{60}.
\bibitem[{Mou et~al.(2019)Mou, Bruzzone and Zhu}]{Mou2019}
\bibinfo{author}{Mou, L.}, \bibinfo{author}{Bruzzone, L.},
  \bibinfo{author}{Zhu, X.X.}, \bibinfo{year}{2019}.
\newblock \bibinfo{title}{{Learning spectral-spatialoral features via a
  recurrent convolutional neural network for change detection in multispectral
  imagery}}.
\newblock \bibinfo{journal}{IEEE Trans. Geosci. Remote Sens.}
  \bibinfo{volume}{57}, \bibinfo{pages}{924--935}.
\bibitem[{Nielsen(2007)}]{Nielsen2007}
\bibinfo{author}{Nielsen, A.A.}, \bibinfo{year}{2007}.
\newblock \bibinfo{title}{{The regularized iteratively reweighted MAD method
  for change detection in multi- and hyperspectral data}}.
\newblock \bibinfo{journal}{IEEE Trans. Image Process.} \bibinfo{volume}{16},
  \bibinfo{pages}{463--478}.
\bibitem[{Nielsen et~al.(1998)Nielsen, Conradsen and Simpson}]{Nielsen1997}
\bibinfo{author}{Nielsen, A.A.}, \bibinfo{author}{Conradsen, K.},
  \bibinfo{author}{Simpson, J.J.}, \bibinfo{year}{1998}.
\newblock \bibinfo{title}{{Multivariate alteration detection (MAD) and MAF
  Postprocessing in multispectral, bitemporal image data: New approaches to
  change detection studies}}.
\newblock \bibinfo{journal}{Remote Sens. Environ.} \bibinfo{volume}{64},
  \bibinfo{pages}{1--19}.
\bibitem[{Peng et~al.(2019)Peng, Zhang and Guan}]{Peng2019End}
\bibinfo{author}{Peng, D.}, \bibinfo{author}{Zhang, Y.}, \bibinfo{author}{Guan,
  H.}, \bibinfo{year}{2019}.
\newblock \bibinfo{title}{End-to-end change detection for high resolution
  satellite images using improved unet++}.
\newblock \bibinfo{journal}{Remote Sensing} \bibinfo{volume}{11}.
\bibitem[{Saha et~al.(2019)Saha, Bovolo and Bruzzone}]{Saha2019}
\bibinfo{author}{Saha, S.}, \bibinfo{author}{Bovolo, F.},
  \bibinfo{author}{Bruzzone, L.}, \bibinfo{year}{2019}.
\newblock \bibinfo{title}{{Unsupervised deep change vector analysis for
  multiple-change detection in VHR Images}}.
\newblock \bibinfo{journal}{IEEE Trans. Geosci. Remote Sens.}
  \bibinfo{volume}{57}, \bibinfo{pages}{3677--3693}.
\bibitem[{Shi et~al.(2022)Shi, Liu, Li, Liu, Wang and Zhang}]{Shi2022Deeply}
\bibinfo{author}{Shi, Q.}, \bibinfo{author}{Liu, M.}, \bibinfo{author}{Li, S.},
  \bibinfo{author}{Liu, X.}, \bibinfo{author}{Wang, F.},
  \bibinfo{author}{Zhang, L.}, \bibinfo{year}{2022}.
\newblock \bibinfo{title}{A deeply supervised attention metric-based network
  and an open aerial image dataset for remote sensing change detection}.
\newblock \bibinfo{journal}{IEEE Trans. Geosci. Remote Sens.}
  \bibinfo{volume}{60}, \bibinfo{pages}{1--16}.
\bibitem[{Shi et~al.(2020)Shi, Zhang, Zhang, Chen and Zhan}]{Shi2020Change}
\bibinfo{author}{Shi, W.}, \bibinfo{author}{Zhang, M.}, \bibinfo{author}{Zhang,
  R.}, \bibinfo{author}{Chen, S.}, \bibinfo{author}{Zhan, Z.},
  \bibinfo{year}{2020}.
\newblock \bibinfo{title}{Change detection based on artificial intelligence:
  State-of-the-art and challenges}.
\newblock \bibinfo{journal}{Remote Sensing} \bibinfo{volume}{12}.
\bibitem[{Singh(1989)}]{Singh1989}
\bibinfo{author}{Singh, A.}, \bibinfo{year}{1989}.
\newblock \bibinfo{title}{{Review Articlel: Digital change detection techniques
  using remotely-sensed data}}.
\newblock \bibinfo{journal}{Int. J. Remote Sens.} \bibinfo{volume}{10},
  \bibinfo{pages}{989--1003}.
\bibitem[{Tan and Le(2019)}]{Tan2019EfficientNet}
\bibinfo{author}{Tan, M.}, \bibinfo{author}{Le, Q.}, \bibinfo{year}{2019}.
\newblock \bibinfo{title}{Efficientnet: Rethinking model scaling for
  convolutional neural networks}, in: \bibinfo{editor}{Chaudhuri, K.},
  \bibinfo{editor}{Salakhutdinov, R.} (Eds.), \bibinfo{booktitle}{Proceedings
  of the 36th International Conference on Machine Learning},
  \bibinfo{publisher}{PMLR}. pp. \bibinfo{pages}{6105--6114}.
\bibitem[{Tang et~al.(2022)Tang, Zhang, Mou, Liu, Zhang, Zhu and
  Jiao}]{Tang2022}
\bibinfo{author}{Tang, X.}, \bibinfo{author}{Zhang, H.}, \bibinfo{author}{Mou,
  L.}, \bibinfo{author}{Liu, F.}, \bibinfo{author}{Zhang, X.},
  \bibinfo{author}{Zhu, X.X.}, \bibinfo{author}{Jiao, L.},
  \bibinfo{year}{2022}.
\newblock \bibinfo{title}{{An Unsupervised Remote Sensing Change Detection
  Method Based on Multiscale Graph Convolutional Network and Metric Learning}}.
\newblock \bibinfo{journal}{IEEE Trans. Geosci. Remote Sens.}
  \bibinfo{volume}{60}.
\bibitem[{Tewkesbury et~al.(2015)Tewkesbury, Comber, Tate, Lamb and
  Fisher}]{Tewkesbury2015}
\bibinfo{author}{Tewkesbury, A.P.}, \bibinfo{author}{Comber, A.J.},
  \bibinfo{author}{Tate, N.J.}, \bibinfo{author}{Lamb, A.},
  \bibinfo{author}{Fisher, P.F.}, \bibinfo{year}{2015}.
\newblock \bibinfo{title}{{A critical synthesis of remotely sensed optical
  image change detection techniques}}.
\newblock \bibinfo{journal}{Remote Sens. Environ.} \bibinfo{volume}{160},
  \bibinfo{pages}{1--14}.
\bibitem[{Tian et~al.(2022)Tian, Zhong, Zheng, Ma, Tan and Zhang}]{TIAN2022164}
\bibinfo{author}{Tian, S.}, \bibinfo{author}{Zhong, Y.},
  \bibinfo{author}{Zheng, Z.}, \bibinfo{author}{Ma, A.}, \bibinfo{author}{Tan,
  X.}, \bibinfo{author}{Zhang, L.}, \bibinfo{year}{2022}.
\newblock \bibinfo{title}{Large-scale deep learning based binary and semantic
  change detection in ultra high resolution remote sensing imagery: From
  benchmark datasets to urban application}.
\newblock \bibinfo{journal}{ISPRS J. Photogramm. Remote Sens.}
  \bibinfo{volume}{193}, \bibinfo{pages}{164--186}.
\bibitem[{Wu et~al.(2022)Wu, Chen, Du and Zhang}]{Wu2022Unsupervised}
\bibinfo{author}{Wu, C.}, \bibinfo{author}{Chen, H.}, \bibinfo{author}{Du, B.},
  \bibinfo{author}{Zhang, L.}, \bibinfo{year}{2022}.
\newblock \bibinfo{title}{Unsupervised change detection in multitemporal vhr
  images based on deep kernel pca convolutional mapping network}.
\newblock \bibinfo{journal}{IEEE Trans. Cybern} \bibinfo{volume}{52},
  \bibinfo{pages}{12084--12098}.
\bibitem[{Wu et~al.(2014)Wu, Du and Zhang}]{Wu2014}
\bibinfo{author}{Wu, C.}, \bibinfo{author}{Du, B.}, \bibinfo{author}{Zhang,
  L.}, \bibinfo{year}{2014}.
\newblock \bibinfo{title}{{Slow feature analysis for change detection in
  multispectral imagery}}.
\newblock \bibinfo{journal}{IEEE Trans. Geosci. Remote Sens.}
  \bibinfo{volume}{52}, \bibinfo{pages}{2858--2874}.
\bibitem[{Wu et~al.(2021)Wu, Li, Qin, Ni, Zhang, Fu and Sun}]{Wu2021multiscale}
\bibinfo{author}{Wu, J.}, \bibinfo{author}{Li, B.}, \bibinfo{author}{Qin, Y.},
  \bibinfo{author}{Ni, W.}, \bibinfo{author}{Zhang, H.}, \bibinfo{author}{Fu,
  R.}, \bibinfo{author}{Sun, Y.}, \bibinfo{year}{2021}.
\newblock \bibinfo{title}{{A multiscale graph convolutional network for change
  detection in homogeneous and heterogeneous remote sensing images}}.
\newblock \bibinfo{journal}{Int. J. Appl. Earth Obs. Geoinf.}
  \bibinfo{volume}{105}, \bibinfo{pages}{102615}.
\bibitem[{Xian et~al.(2009)Xian, Homer and Fry}]{Xian2009}
\bibinfo{author}{Xian, G.}, \bibinfo{author}{Homer, C.}, \bibinfo{author}{Fry,
  J.}, \bibinfo{year}{2009}.
\newblock \bibinfo{title}{{Updating the 2001 National Land Cover Database land
  cover classification to 2006 by using Landsat imagery change detection
  methods}}.
\newblock \bibinfo{journal}{Remote Sens. Environ.} \bibinfo{volume}{113},
  \bibinfo{pages}{1133--1147}.
\bibitem[{Xiao et~al.(2016)Xiao, Zhang, Wang, Yuan, Feng and
  Kelly}]{XIAO2016402}
\bibinfo{author}{Xiao, P.}, \bibinfo{author}{Zhang, X.}, \bibinfo{author}{Wang,
  D.}, \bibinfo{author}{Yuan, M.}, \bibinfo{author}{Feng, X.},
  \bibinfo{author}{Kelly, M.}, \bibinfo{year}{2016}.
\newblock \bibinfo{title}{Change detection of built-up land: A framework of
  combining pixel-based detection and object-based recognition}.
\newblock \bibinfo{journal}{ISPRS J. Photogramm. Remote Sens.}
  \bibinfo{volume}{119}, \bibinfo{pages}{402--414}.
\bibitem[{Xie et~al.(2021)Xie, Wang, Yu, Anandkumar, Alvarez and
  Luo}]{Xie2021SegFormer}
\bibinfo{author}{Xie, E.}, \bibinfo{author}{Wang, W.}, \bibinfo{author}{Yu,
  Z.}, \bibinfo{author}{Anandkumar, A.}, \bibinfo{author}{Alvarez, J.M.},
  \bibinfo{author}{Luo, P.}, \bibinfo{year}{2021}.
\newblock \bibinfo{title}{Segformer: Simple and efficient design for semantic
  segmentation with transformers}, in: \bibinfo{editor}{Ranzato, M.},
  \bibinfo{editor}{Beygelzimer, A.}, \bibinfo{editor}{Dauphin, Y.},
  \bibinfo{editor}{Liang, P.}, \bibinfo{editor}{Vaughan, J.W.} (Eds.),
  \bibinfo{booktitle}{Advances in Neural Information Processing Systems},
  \bibinfo{publisher}{Curran Associates, Inc.}. pp.
  \bibinfo{pages}{12077--12090}.
\bibitem[{Yang et~al.(2022)Yang, Xia, Liu, Du, Yang, Pelillo and
  Zhang}]{Yang2022Asymmetric}
\bibinfo{author}{Yang, K.}, \bibinfo{author}{Xia, G.S.}, \bibinfo{author}{Liu,
  Z.}, \bibinfo{author}{Du, B.}, \bibinfo{author}{Yang, W.},
  \bibinfo{author}{Pelillo, M.}, \bibinfo{author}{Zhang, L.},
  \bibinfo{year}{2022}.
\newblock \bibinfo{title}{Asymmetric siamese networks for semantic change
  detection in aerial images}.
\newblock \bibinfo{journal}{IEEE Trans. Geosci. Remote Sens.}
  \bibinfo{volume}{60}, \bibinfo{pages}{1--18}.
\bibitem[{Zhan et~al.(2017)Zhan, Fu, Yan, Sun, Wang and Qiu}]{Zhan2017}
\bibinfo{author}{Zhan, Y.}, \bibinfo{author}{Fu, K.}, \bibinfo{author}{Yan,
  M.}, \bibinfo{author}{Sun, X.}, \bibinfo{author}{Wang, H.},
  \bibinfo{author}{Qiu, X.}, \bibinfo{year}{2017}.
\newblock \bibinfo{title}{{Change Detection Based on Deep Siamese Convolutional
  Network for Optical Aerial Images}}.
\newblock \bibinfo{journal}{IEEE Geosci. Remote Sens. Lett.}
  \bibinfo{volume}{14}, \bibinfo{pages}{1845--1849}.
\bibitem[{Zhang et~al.(2020)Zhang, Yue, Tapete, Jiang, Shangguan, Huang and
  Liu}]{Zhang2020}
\bibinfo{author}{Zhang, C.}, \bibinfo{author}{Yue, P.},
  \bibinfo{author}{Tapete, D.}, \bibinfo{author}{Jiang, L.},
  \bibinfo{author}{Shangguan, B.}, \bibinfo{author}{Huang, L.},
  \bibinfo{author}{Liu, G.}, \bibinfo{year}{2020}.
\newblock \bibinfo{title}{{A deeply supervised image fusion network for change
  detection in high resolution bi-temporal remote sensing images}}.
\newblock \bibinfo{journal}{ISPRS J. Photogramm. Remote Sens.}
  \bibinfo{volume}{166}, \bibinfo{pages}{183--200}.
\bibitem[{Zhang et~al.(2016a)Zhang, Gong, Zhang, Su, Shi, Member, Zhang, Su and
  Shi}]{Zhang2016c}
\bibinfo{author}{Zhang, H.}, \bibinfo{author}{Gong, M.},
  \bibinfo{author}{Zhang, P.}, \bibinfo{author}{Su, L.}, \bibinfo{author}{Shi,
  J.}, \bibinfo{author}{Member, S.}, \bibinfo{author}{Zhang, P.},
  \bibinfo{author}{Su, L.}, \bibinfo{author}{Shi, J.}, \bibinfo{year}{2016}a.
\newblock \bibinfo{title}{{Feature-Level Change Detection Using Deep
  Representation and Feature Change Analysis for Multispectral Imagery}}.
\newblock \bibinfo{journal}{IEEE Geosci. Remote Sens. Lett.}
  \bibinfo{volume}{13}, \bibinfo{pages}{1666--1670}.
\bibitem[{Zhang et~al.(2016b)Zhang, Gong, Su, Liu and Li}]{Zhang2016}
\bibinfo{author}{Zhang, P.}, \bibinfo{author}{Gong, M.}, \bibinfo{author}{Su,
  L.}, \bibinfo{author}{Liu, J.}, \bibinfo{author}{Li, Z.},
  \bibinfo{year}{2016}b.
\newblock \bibinfo{title}{{Change detection based on deep feature
  representation and mapping transformation for multi-spatial-resolution remote
  sensing images}}.
\newblock \bibinfo{journal}{ISPRS J. Photogramm. Remote Sens.}
  \bibinfo{volume}{116}, \bibinfo{pages}{24--41}.
\bibitem[{Zheng et~al.(2021a)Zheng, Ma, Zhang and Zhong}]{zheng2021change}
\bibinfo{author}{Zheng, Z.}, \bibinfo{author}{Ma, A.}, \bibinfo{author}{Zhang,
  L.}, \bibinfo{author}{Zhong, Y.}, \bibinfo{year}{2021}a.
\newblock \bibinfo{title}{Change is everywhere: Single-temporal supervised
  object change detection in remote sensing imagery}, in:
  \bibinfo{booktitle}{Proceedings of the IEEE/CVF international conference on
  computer vision}, pp. \bibinfo{pages}{15193--15202}.
\bibitem[{Zheng et~al.(2022)Zheng, Zhong, Tian, Ma and Zhang}]{Zheng2022}
\bibinfo{author}{Zheng, Z.}, \bibinfo{author}{Zhong, Y.},
  \bibinfo{author}{Tian, S.}, \bibinfo{author}{Ma, A.}, \bibinfo{author}{Zhang,
  L.}, \bibinfo{year}{2022}.
\newblock \bibinfo{title}{{ChangeMask: Deep multi-task
  encoder-transformer-decoder architecture for semantic change detection}}.
\newblock \bibinfo{journal}{ISPRS J. Photogramm. Remote Sens.}
  \bibinfo{volume}{183}, \bibinfo{pages}{228--239}.
\bibitem[{Zheng et~al.(2021b)Zheng, Zhong, Wang, Ma and
  Zhang}]{ZHENG2021Building}
\bibinfo{author}{Zheng, Z.}, \bibinfo{author}{Zhong, Y.},
  \bibinfo{author}{Wang, J.}, \bibinfo{author}{Ma, A.}, \bibinfo{author}{Zhang,
  L.}, \bibinfo{year}{2021}b.
\newblock \bibinfo{title}{Building damage assessment for rapid disaster
  response with a deep object-based semantic change detection framework: From
  natural disasters to man-made disasters}.
\newblock \bibinfo{journal}{Remote Sens. Environ.} \bibinfo{volume}{265},
  \bibinfo{pages}{112636}.
\bibitem[{Zhou et~al.(2018)Zhou, Rahman~Siddiquee, Tajbakhsh and
  Liang}]{zhou2018unet}
\bibinfo{author}{Zhou, Z.}, \bibinfo{author}{Rahman~Siddiquee, M.M.},
  \bibinfo{author}{Tajbakhsh, N.}, \bibinfo{author}{Liang, J.},
  \bibinfo{year}{2018}.
\newblock \bibinfo{title}{Unet++: A nested u-net architecture for medical image
  segmentation}, in: \bibinfo{booktitle}{Deep Learning in Medical Image
  Analysis and Multimodal Learning for Clinical Decision Support: 4th
  International Workshop, DLMIA 2018, and 8th International Workshop, ML-CDS
  2018, Held in Conjunction with MICCAI 2018, Granada, Spain, September 20,
  2018, Proceedings 4}, \bibinfo{organization}{Springer}. pp.
  \bibinfo{pages}{3--11}.
\bibitem[{Zhu(2017)}]{Zhu2017b}
\bibinfo{author}{Zhu, Z.}, \bibinfo{year}{2017}.
\newblock \bibinfo{title}{{Change detection using landsat time series: A review
  of frequencies, preprocessing , algorithms, and applications}}.
\newblock \bibinfo{journal}{ISPRS J. Photogramm. Remote Sens.}
  \bibinfo{volume}{130}, \bibinfo{pages}{370--384}.
\bibitem[{Zou et~al.(2018)Zou, Yu, Kumar and Wang}]{Zou_2018_ECCV}
\bibinfo{author}{Zou, Y.}, \bibinfo{author}{Yu, Z.}, \bibinfo{author}{Kumar,
  B.V.}, \bibinfo{author}{Wang, J.}, \bibinfo{year}{2018}.
\newblock \bibinfo{title}{Unsupervised domain adaptation for semantic
  segmentation via class-balanced self-training}, in:
  \bibinfo{booktitle}{Proceedings of the European Conference on Computer Vision
  (ECCV)}.

\end{thebibliography}





\end{document}